\documentclass{article} % For LaTeX2e
\usepackage{iclr2026_conference,times}

\usepackage[utf8]{inputenc} % allow utf-8 input
\usepackage[T1]{fontenc}    % use 8-bit T1 fonts
\usepackage{hyperref}       % hyperlinks
\usepackage{amsmath}
\usepackage{cleveref}       % enhanced cross-referencing

\usepackage{multirow}
\usepackage{url}            % simple URL typesetting
\usepackage{fontawesome5}   % for \faGithub icon
\usepackage{enumitem}       % customize list environments
\usepackage{graphicx}       % for including graphics
\usepackage{amsfonts}       % blackboard math symbols
\usepackage{nicefrac}       % compact symbols for 1/2, etc.
\usepackage{microtype}      % microtypography
\usepackage{tcolorbox}
\usepackage{wrapfig}

\usepackage{amsmath,amsfonts,bm}

\def\eqref#1{equation~\ref{#1}}
\def\1{\bm{1}}

\DeclareMathAlphabet{\mathsfit}{\encodingdefault}{\sfdefault}{m}{sl}
\SetMathAlphabet{\mathsfit}{bold}{\encodingdefault}{\sfdefault}{bx}{n}

\usepackage{hyperref}
\usepackage{url}
\usepackage{pifont} % for \ding{51} (\cmark) and \ding{55} (\xmark) symbols
\usepackage{tabularx}
\usepackage{makecell}
\usepackage{caption}
\usepackage{booktabs}       % professional-quality tables
\usepackage[table]{xcolor}

\usepackage{inconsolata}
\usepackage{listings}

\title{Vision-Language-Action Instruction Tuning:\\
From Understanding to Manipulation}

\iclrfinalcopy % Uncomment for camera-ready version, but NOT for submission.

\author{%
    \textbf{Shuai Yang}$^{2,3}$\thanks{Equal contributions.} \quad
    \textbf{Hao Li}$^{1,3*}$ \quad
    \textbf{Bin Wang}$^{2,3}$ \quad
    \textbf{Yilun Chen}$^{3}$ \quad
    \textbf{Yang Tian}$^{3}$ \quad
    \textbf{Tai Wang}$^{3}$ \\
    \textbf{Hanqing Wang}$^{3}$ \quad
    \textbf{Feng Zhao}$^{1}$ \quad
    \textbf{Yiyi Liao}$^{2}$ \quad
    \textbf{Jiangmiao Pang}$^{3}$\\
    $^1$University of Science and Technology of China, 
    $^2$Zhejiang University,\\
    $^3$Shanghai Artificial Intelligence Laboratory
}

\begin{document}

\maketitle
\let\oldaddcontentsline\addcontentsline
\renewcommand{\addcontentsline}[3]{}
\begin{abstract}

To operate effectively in the real world, robots should integrate multimodal reasoning with precise action generation. However, existing vision-language-action (VLA) models often sacrifice one for the other, narrow their abilities to task-specific manipulation data, and suffer catastrophic forgetting of pre-trained vision-language capabilities. To bridge this gap, we introduce \textbf{InstructVLA}, an end-to-end VLA model that preserves the flexible reasoning of large vision-language models (VLMs) while delivering leading manipulation performance with the help of embodied reasoning. InstructVLA introduces a novel training paradigm, \textit{Vision-Language-Action Instruction Tuning (VLA-IT)}, which employs multimodal training with mixture-of-experts adaptation to jointly optimize embodied reasoning and action generation on both standard VLM corpora and a curated 650K-sample VLA-IT dataset. On in-domain SimplerEnv tasks, InstructVLA achieves 33\% improvement over SpatialVLA. To evaluate generalization, we introduce SimplerEnv-Instruct, an 80-task benchmark requiring closed-loop control and high-level instruction understanding, where it outperforms a fine-tuned OpenVLA by 96\% and an action expert aided by GPT-4o by 29\%. Additionally, InstructVLA surpasses baseline VLMs on multimodal tasks and exhibits inference-time scaling by leveraging textual reasoning to boost manipulation performance in both simulated and real-world settings. These results demonstrate InstructVLA's potential for bridging intuitive and steerable human-robot interaction with efficient policy learning. \href{https://yangs03.github.io/InstructVLA_Home/}{\textcolor{blue}{Project website.}} 
\end{abstract}

\section{Introduction}

Large-scale pretraining has produced versatile foundation models in computer vision (CV)~\citep{oquab2023dinov2,clip} and natural language processing (NLP)~\citep{qwen,touvron2023llama}. Building on this progress, recent Vision-Language-Action (VLA) models~\citep{pi_0,openvla} adapt large vision-language models (VLMs)~\citep{karamcheti2024prismatic,beyer2024paligemma} and finetune them on embodied datasets to achieve generalizable manipulation. {While the integration of multimodal reasoning has led to significant advances in VLMs~\citep{cot,liu2024deepseek}, such reasoning remains largely unexplored in VLA settings}. Fully leveraging VLMs for reasoning-guided manipulation beyond VLA initialization remains an open challenge. Current attempts to incorporate the reasoning capabilities of VLMs into action learning face three main obstacles: (1) task interference, catastrophic forgetting~\citep{french1999catastrophic} of multimodal ability during action training; (2) data scarcity, particularly the limited availability of manipulation datasets with rich multimodal supervision; and (3) methodological gaps, specifically the lack of effective mechanisms and training paradigm to translate multimodal reasoning into action generation. These limitations lead to a fundamental question for VLA-based manipulation:

{

\emph{How can we acquire manipulation skills without eroding the VLM’s multimodal reasoning, and how can such reasoning, in turn, enhance manipulation?}

}

To address these challenges and utilize VLMs more effectively, prior work has primarily adopted two strategies. The first aims to retain general multimodal capabilities while learning manipulation skills through unified auto-regressive modeling. Models such as RT-2~\citep{RT-2} and Magma~\citep{magma} follow this approach by co-training on vision-language and manipulation data. Yet, this paradigm often overlooks complex embodied reasoning, and our ablations reveal that the general VLM corpus exhibits a domain gap in embodied scenarios. The second strategy tightly integrates embodied reasoning into manipulation datasets to transfer VLM capabilities. Methods such as ECoT~\citep{ecot} and Emma-X~\citep{sun2024emma} embed chain-of-thought (CoT) reasoning into manipulation datasets. While promising, these methods rely on action-pretrained architectures~\citep{openvla} and structured reasoning formats (e.g., subtasks, grounding), which limit expressiveness, suffer from catastrophic forgetting, and fail to demonstrate general multimodal capabilities-even with additional finetuning. Consequently, the extent to which VLM capabilities can be effectively translated into action generation in embodied contexts remains largely unexplored.

\begin{figure*}[t]
    \centering
    \includegraphics[width=0.9\linewidth]{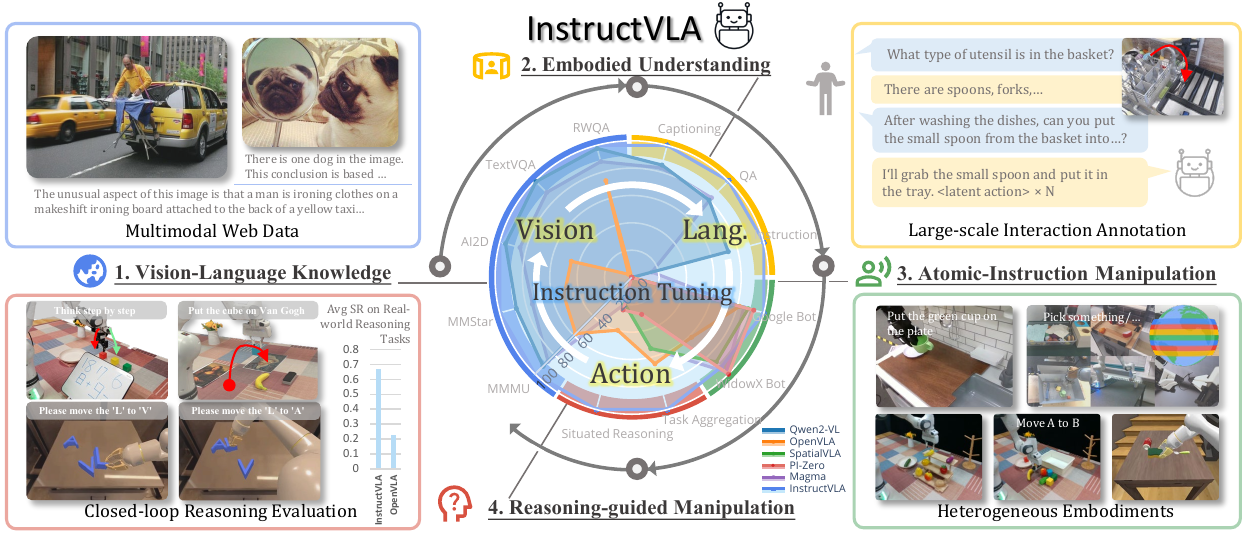}
    \caption{\textbf{Method overview.} InstructVLA integrates vision-language understanding with precise robotic control to achieve reasoning-guided manipulation. Its core training strategy, \textbf{Vision-Language-Action Instruction Tuning}, enhances manipulation by unifying general multimodal knowledge, embodied reasoning, and atomic instruction-based manipulation into a coherent chain of thought.}
    \vspace{-0.4cm}
    \label{fig:teaser}
\end{figure*}

Building on these observations, we propose \textbf{InstructVLA}, a generalist VLA model that extends pretrained VLMs for accurate action generation while preserving strong multimodal understanding. Building on this unified modeling, we conduct extensive experiments to investigate how multimodal capabilities contribute to manipulation. Motivated by these insights, we design the \textbf{Vision-Language-Action Intruction Tuning} (VLA-IT) paradigm specifically tailored to bridge vision-language knowledge with action generation, treating language-conditioned action generation as an integral component of instruction following, as illustrated in~\Cref{fig:teaser}. To support this paradigm, we curate the \textbf{Vision-Language-Action Instruction Tuning dataset}, consisting of 650K human-robot interactions annotated with diverse instructions, scene captions, and question-answer pairs grounded in high-quality manipulation tasks~\citep{Bridge_data, RT-1}. The training follows two stages: (1) \textit{Action Pretraining}, which trains a VLM-driven action expert using latent action queries distilled from language-based motion descriptions, providing a learnable interface to the VLM while decoupling low-level control learning from the VLM backbone to preserve its multimodal reasoning capabilities; (2) \textit{Vision-Language-Action Instruction Tuning}, which unifies language and latent action generation through a trainable mixture-of-experts(MoE) adaptation framework. This framework is jointly trained on multimodal datasets~\citep{bunny}, manipulation datasets, and the curated VLA-IT corpus, enabling the automatic switch between textual reasoning and action generation, thereby effectively leveraging vision-language understanding and reasoning for action generation.

To validate the performance of InstructVLA, we introduce the \textbf{SimplerEnv-Instruct benchmark}, a manually designed evaluation suite featuring 80 zero-shot manipulation tasks. It encompasses both closed-loop manipulation tasks and high-level instruction reasoning, involving either situated understanding or decomposition into actionable subtasks. With its thinking ability during manipulation, InstructVLA outperforms the fine-tuned OpenVLA baseline by 96\% and achieves a 29\% improvement over an action expert model assisted by GPT-4o on SimplerEnv-Instruct, demonstrating its effectiveness in instruction following and task decomposition. Furthermore, InstructVLA surpasses similarly sized VLMs in multimodal performance and shows a 33\% improvement over SpatialVLA in closed-loop manipulation~\citep{simpleenv}. Our contributions can be summarized as follows:
\begin{itemize}[leftmargin=0.1in]

\item \textbf{Model.} We propose \textbf{InstructVLA}, a VLA architecture and training pipeline that \textbf{supports studying language capability in VLAs} by efficiently preserving pretrained vision-language knowledge from VLMs while integrating manipulation as a component of instruction following.

\item \textbf{Dataset \& Benchmark.} We design a \textbf{practical data and evaluation pipeline} for vision-language-action instruction following, supported by 650K tailored VLA-IT annotations and a manually curated benchmark suite, enabling evaluation of VLAs' instruction generalization capabilities.

\item \textbf{Validation.} InstructVLA achieves leading performance across robotic manipulation tasks, multimodal benchmarks, and real-world deployments, enabling intuitive and controllable manipulation.
\end{itemize}

\section{Related Works}

\noindent\textbf{Policy learning at scale.} Following the success of CV~\citep{oquab2023dinov2, siglip}and NLP~\citep{touvron2023llama}, recent research~\citep{hpt,RT-1,RT-2, zheng2025universalactionsenhancedembodied,wang2024poco,niu2025pre} shows that robot policies improve when trained in large heterogeneous datasets. RT-1~\citep{RT-1} and RT-2~\citep{RT-2}, trained in large-scale real-world demonstrations, achieve strong in-domain accuracy and zero-shot transfer. Works such as Octo~\citep{octo} and RT-X~\citep{open_x_embodiment} extend this approach by aggregating the largest open-source manipulation datasets~\citep{open_x_embodiment}. Some methods, such as LAPA~\citep{lapa}, Seer~\citep{tian2024predictive}, and Moto~\citep{chen2024moto}, use video generation and inverse dynamics to learn scalable motor representations. In the VLA domain, models are typically initialized from pretrained vision-language models~\citep{openvla, qu2025spatialvla, RT-2} leveraging prior visual-linguistic alignment instead of learning from scratch. Further, methods such as RT-Trajectory~\citep{rt-trajectory} and GraspVLA~\citep{deng2025graspvla} jointly train intermediate manipulation representations such as trajectories or bounding boxes using a combination of real and simulated data to guide action generation and enhance generalization.

\noindent\textbf{Vision-language-action models.} Recent foundation models~\citep{RT-2, openvla, qu2025spatialvla, pi_0, chen2024moto, bjorck2025gr00t, pertsch2025fast,lingbot_vla} integrate perception, language, and robot manipulation into a single network, using two main architectures. Autoregressive models such as RT-2~\citep{RT-2}, OpenVLA~\citep{openvla} and SpatialVLA~\citep{qu2025spatialvla} treat actions as discrete tokens. LLARVA~\citep{niu2024llarva} introduces 2D trace for pretraining. FAST tokenization~\citep{pertsch2025fast} further compresses motion sequences. In contrast, flow-based VLAs avoid discretization; for example, $\pi_0$~\citep{pi_0} and GR00T~\citep{bjorck2025gr00t} generate actions through continuous flow matching~\citep{flowmatching}, while CogACT~\citep{cogact} and CronusVLA~\citep{li2025cronusvla} use diffusion~\citep{DiT}. Hybrid approaches, like RoboDual~\citep{robodual}, combine generalist action models with specialist action experts. Although flow-based methods~\citep{pi_0,bjorck2025gr00t,li2025cronusvla,cogact} often achieve superior performance, they typically neglect the integration of autoregressive text reasoning~\citep{RT-2}, which is crucial for leveraging the VLM’s semantic capabilities. In contrast, our model unifies autoregressive VLM language generation with the flow-based action generation, demonstrating efficient co-training of language and action.

\noindent\textbf{{Bringing step-by-step reasoning ability to manipulation.}} Bridging pre-trained world knowledge to enhance the generalization of robot policies is a promising direction. One line of work standardizes intermediate representations~\citep{li2026robointer}, such as primitive~\citep{rh20t}, trajectories~\citep{li2025hamster}, keypoints~\citep{li2024llara} and masks~\citep{huang2025roboground}. However, these approaches often rely on rule-based decomposition and hand-crafted planning heuristics, whose rigid separation from low-level control limits scalability and hinders end-to-end policy learning. CoT-VLA~\citep{zhao2025cot} instead treats future video generation as an implicit chain-of-thought, but predicting image tokens step-by-step introduces computational overhead, limiting practicality for fast closed-loop control. More recently, unified modeling of perception, reasoning, and manipulation~\citep{pi05,helix,lcb}, along with other generative formulations~\citep{pan2025transfer,zhou2024transfusion}, has demonstrated the potential of leveraging pre-trained VLMs and LLMs for reasoning-guided generation, revealing emerging capabilities~\citep{deng2025emerging}. Yet, many prior studies depend on closed-source data~\citep{pi05} or conduct limited evaluation in real-world settings~\citep{rth,zhou2025chatvla}, constraining reproducibility and large-scale assessment. Our work provides a pioneering exploration, supported by open data and benchmarks, to study \emph{reasoning-guided manipulation} through the integration of reasoning and action.

\section{InstructVLA}

We propose \textbf{InstructVLA} (\Cref{fig:model}), a unified model for joint language-action generation that also mitigates task interference and catastrophic forgetting. \Cref{sec: arch} describes the architecture, including dynamic switching between reasoning and execution modes, as well as inference strategies. \Cref{sec: recipe} presents the training paradigm for the instruction following of VLAs.

\subsection{Architecture}
\label{sec: arch}
\begin{figure*}[t]
    \centering
    \includegraphics[width=1\linewidth]{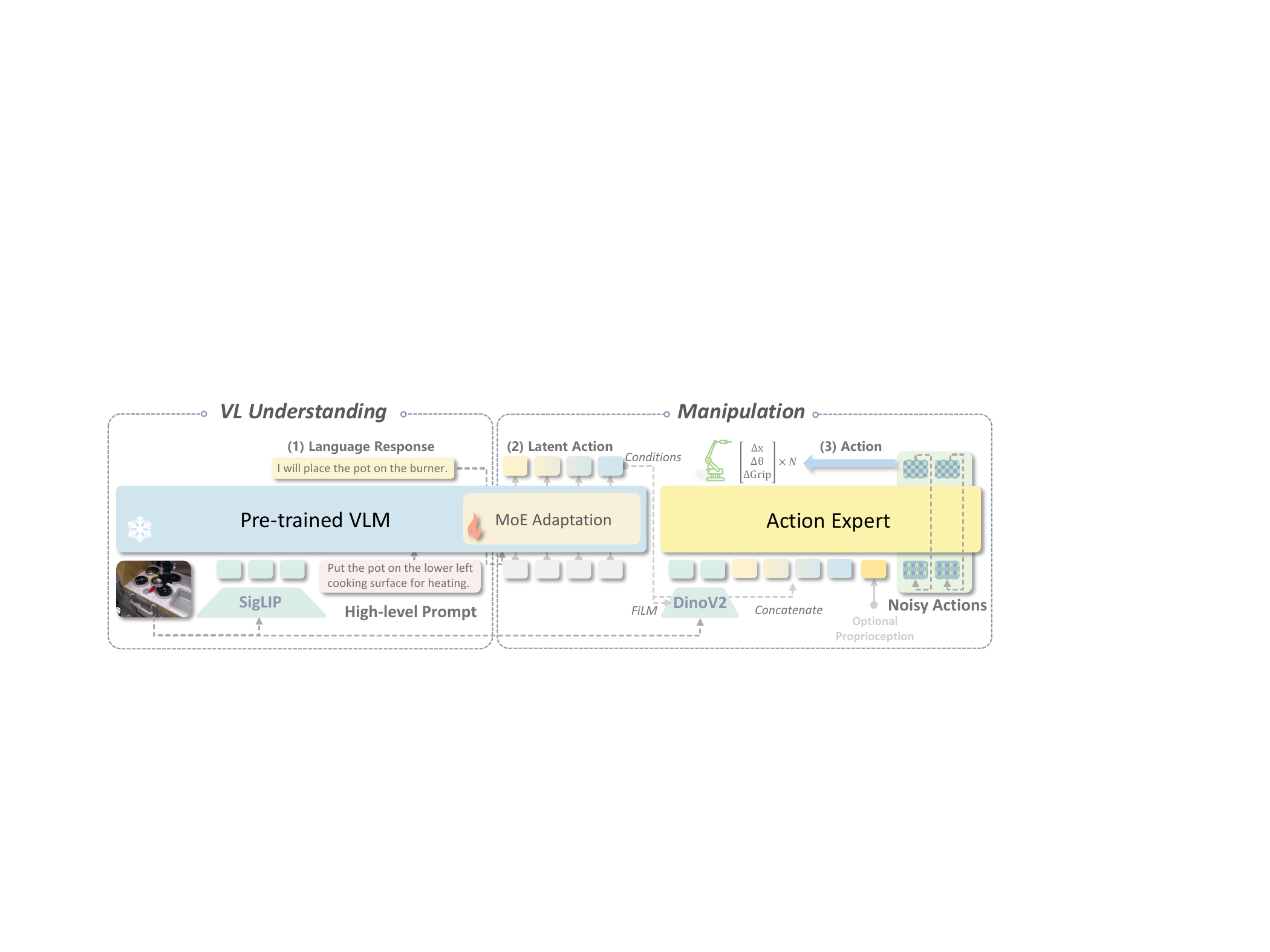}
\caption{\textbf{Overview of the InstructVLA.} InstructVLA integrates the multimodal reasoning capabilities of a vision-language model with robotic manipulation. Generation consists of three steps: (1) asynchronous auto-regressive reasoning by the VLM, (2) latent action generation, and (3) action decoding. A MoE adaptation enables the VLM to alternate between reasoning and latent action prediction. The flow matching action expert decodes the final actions, conditioned on latent actions.}
    % \vspace{-0.5cm}
    \label{fig:model}
\end{figure*}

\noindent\textbf{Embodied VLM for text and latent action generation.} We propose a unified framework that enables simultaneous multimodal reasoning and language-steered latent action planning using a single VLM (\Cref{fig:model} (1) and (2)). The model produces textual outputs to preserve the strong language understanding and multimodal inference capabilities of the pretrained VLM, while subsequently generating latent action representations for downstream manipulation. To support action planning, we introduce $N$ learnable action queries $Q \in \mathbb{R}^{N \times D}$, which attend to the VLM’s hidden states and extract task-relevant latent action $C \in \mathbb{R}^{N \times D}$, where $D$ is the VLM hidden dimension. Our implementation builds on the compact and efficient Eagle2-2B backbone~\citep{li2025eagle}, with a tailored training strategy described in~\Cref{sec: vla tuning}. The VLM is supervised with cross-entropy on language output with loss $\mathcal{L}_{LM}$.

\noindent\textbf{MoE adaptation to harmonize reasoning and action.} A key challenge is enabling the model to seamlessly alternate between reasoning and manipulation. To this end, we adopt a MoE design~\citep{moe}, which allows adaptive reweighting of expert modules based on input context and reasoning mode, thereby integrating multimodal reasoning with language-steered latent action. Specifically, LoRA~\citep{lora} modules are employed as experts within the LLM backbone, preserving pretrained capabilities while ensuring efficient inference. A scalar head~\citep{xlora} predicts gating coefficients $\lambda_i$ for each expert by classifying the hidden state, enabling the model to adaptively blend their outputs. The resulting hidden states for $K$ experts are computed as $h = W_0x + \sum_{i=0}^{K} B_i A_i x \cdot \alpha_i \cdot \lambda_i$, where $W_0$ is the original weight, $x$ denotes input, $A_i \in \mathbb{R}^{r \times d}$ and $B_i \in \mathbb{R}^{d \times r}$ are the LoRA parameters, $\alpha_i$ is the LoRA scalar factor, {as detailed in~\Cref{sec: moe ada}}.

\noindent\textbf{Flow model as an efficient action expert.} To further decouple low-level control from high-level understanding, the action expert is designed to generate actions from image observations conditioned on VLM-derived intentions. It takes image features from DINOv2~\citep{oquab2023dinov2} vision encoder, latent actions, noisy action embeddings and optional information such as proprioception, and fuses these with a simple transformer architecture~\citep{touvron2023llama} with block-wise causal attention. Specifically, non-causal attention is applied within each input, and causal attention between input types.  The vision encoder, further enhanced with feature-wise linear modulation (FiLM)~\citep{perez2018film}, plays a crucial role in directing actions to spatial and contextual input. The flow matching objective~\citep{pi_0} is used to supervise action learning, as detailed in~\Cref{sec: loss}.

\noindent\textbf{Inference.} InstructVLA integrates language and action generation in a single model with the following techniques to improve speed. 
(1) \textit{Decoding strategies.} To mitigate the latency of autoregressive decoding, textual responses are generated via greedy search until the first action query token appears. The remaining action queries are then decoded in parallel within a single forward pass of the VLM.
(2) \textit{Language response and latent action caching.} We decouple language response from action generation by caching textual outputs across multiple action steps, leveraging their temporal stability. InstructVLA also supports cache latent actions, which reduces the number of VLM forward with minimal performance impact compared with ECoT~\citep{ecot}  (see~\Cref{sec: sub dis}).

\subsection{Training Recipe}
\label{sec: recipe}
Direct co-training of vision, language, and action often leads to unstable optimization and slow convergence. We therefore adopt a principled two-stage training paradigm: first, action pretraining to align with the VLM’s latent action embeddings; second, vision-language-action instruction tuning to integrate multimodal reasoning with manipulation.

\noindent\textbf{Stage 1: Action pre-training.} InstructVLA is pre-trained using heterogeneous manipulation data~\citep{RT-1, Bridge_data}. To distill the knowledge from the VLM for manipulation, the model is trained to predict both actions and language motion (\Cref{sec: dataset}), with the latter supervised via cross-entropy loss. Due to the stability of flow matching and the next token prediction, the final loss is the direct sum of both losses as $\mathcal L = \mathcal{L}_{LM}+\mathcal L_{FM}$. During this stage, only the embedding of the latent action and action LoRA adapter on the LLM backbone are tuned, consisting of 650M parameters. The model trained is named the “Expert”.

\noindent\textbf{Stage 2: Vision-language-action instruction tuning.} \label{sec: vla tuning} 
We extend visual instruction tuning~\citep{llava} with a simple and efficient approach to train InstructVLA. Our key observation is that once the action expert has been pretrained to follow latent actions generated by the VLM, further adapting the LLM backbone enables the model to handle manipulation tasks with more complex instructions. {In this stage, a language LoRA and a scalar head are added, which together with the stage 1 action LoRA constitute the MoE adaptation~\citep{xlora}.} This MoE module is the only trainable component in Stage 2, totaling 220M parameters. We detail the data pipeline for vision-language-action instruction tuning in~\Cref{sec: instruction data}; this data bridges pretrained vision-language capabilities with embodied task scenarios. To further bootstrap multimodal understanding, we co-train the model with additional multimodal datasets~\citep{bunny}. The resulting model, referred to as the “Generalist”, integrates both vision-language reasoning and manipulation capabilities.

\section{VLA Dataset and Benchmark}

\subsection{InstructVLA Tuning Dataset}

\label{sec: dataset}
\begin{figure}
    \centering
    \includegraphics[width=1\linewidth]{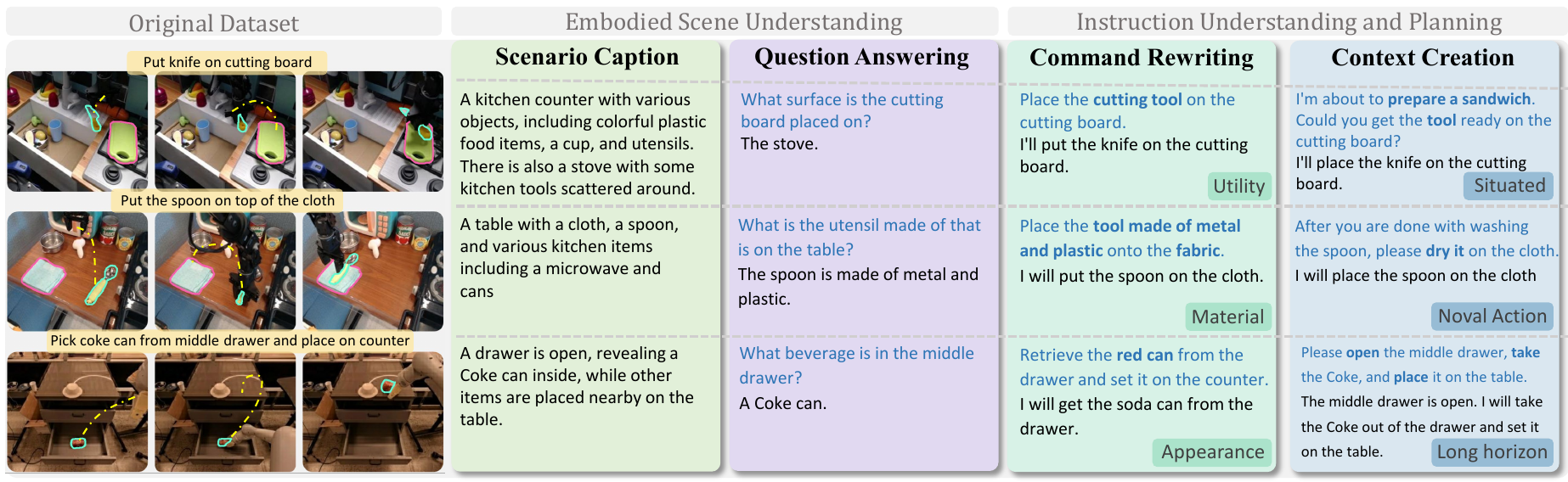}
    \caption{\textbf{Vision-language-action instruction tuning data examples.} Annotations focus on: \\ 
        (1) improving scene understanding and (2) learning instruction following and planning.}
    \label{fig:examples}
    \vspace{-0.5cm}
\end{figure}

We curate diverse hierarchical language annotations from large-scale manipulation datasets~\citep{RT-1, Bridge_data}, including language motion~\citep{rth} as detailed in~\Cref{sec: language motion}, along with the VLA-IT dataset for instruction tuning and reasoning transferring.

\noindent\textbf{Vision-language-action instruction tuning data.} \label{sec: instruction data} To enable language-steerable VLA models, it is essential to curate diverse instructions, model responses, and reasoning patterns. We categorize our data into four types as illustrated in~\Cref{fig:examples}. For embodied scene understanding: \textit{(1) Scenario captioning} provides descriptions of the robot's environment \textit{(2) Question answering} targets scene understanding through consistent QA pairs across an episode. Together, they bridge vision-language annotations with embodied scenes. For instruction understanding and latent action planning: \textit{(3) Command rewriting} introduces instructional diversity through paraphrasing, attribute-based references and varied vocabulary. 
\textit{(4) Context creation} generates implicit user goals or progress cues in multi-step tasks, requiring the robot to infer intent. These annotations support joint VLA reasoning.

We use GPT-4o~\citep{GPT-4} to annotate data with three frames from each episode, along with the corresponding instruction. Ground-truth instruction is crucial for annotation accuracy, emphasizing that even state-of-the-art VLMs can make errors in embodied tasks, leading to a performance gap when using GPT-4o as an instruction interpreter for such tasks. Additional details of the dataset analysis and prompt templates are provided in~\Cref{sec: Data Annotation Details and Analysis}.

\subsection{SimplerEnv-Instruct}
\label{sec: SimplerEnv-Instruct}

\begin{figure*}[t]
    \centering
    \includegraphics[width=1\linewidth]{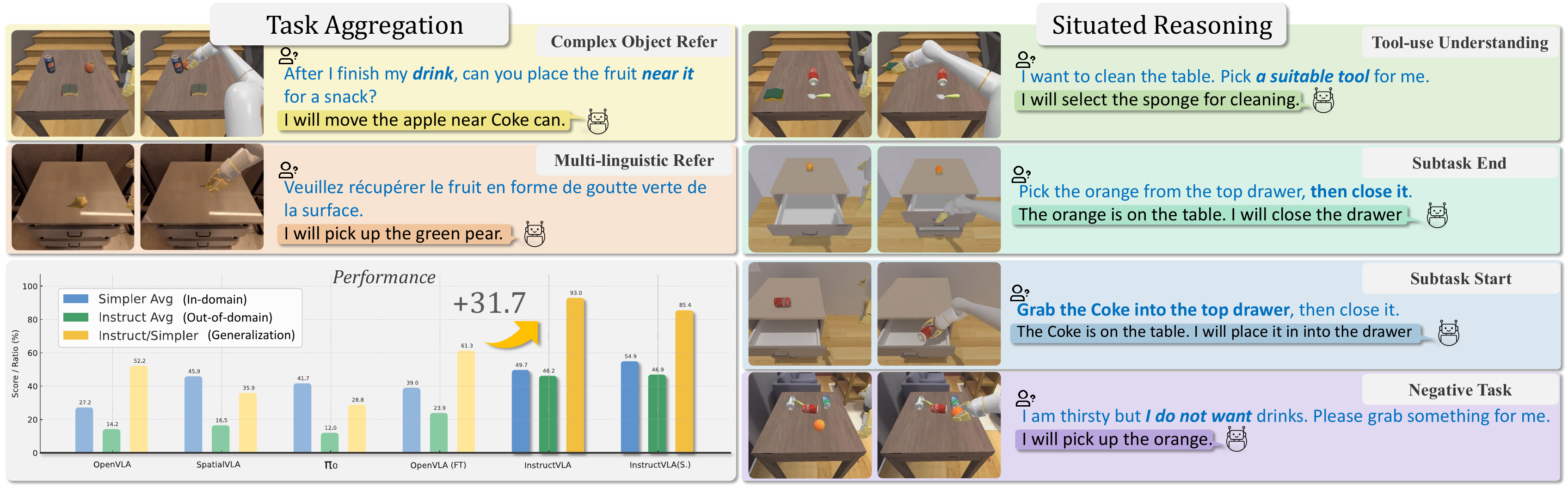}
    \caption{\textbf{Simpler-\textit{Instruct}.} Six representative test cases with instructions and InstructVLA responses. Prior VLAs exhibit limited generalization compared to InstructVLA.}
    \label{fig:simplerinstruct}
    % \vspace{-0.5cm}
\end{figure*}

Building upon the SimplerEnv platform, we introduce \textbf{SimplerEnv-Instruct}, a benchmark specifically designed to evaluate the instruction-following and reasoning capabilities of vision-language-action (VLA) models in a zero-shot setting. Unlike prior manipulation benchmarks that primarily focus on atomic actions or low-level control, SimplerEnv-Instruct captures two essential yet underexplored abilities: (1) policy generalization to linguistic and visual diversity, and (2) contextual reasoning in situated environments, evaluated in the \textit{situated reasoning} suite.  

\noindent\textbf{Task creation.} We remove trivial cases and design novel tasks requiring genuine generalization rather than memorization. Novel objects and instructions are strictly out-of-distribution from the originals, and all tasks are cross-validated by three annotators for clarity and consistency. In total, we curated 80 tasks with 1.1K trials, about one third the size of SimplerEnv, keeping evaluation practical.

\begin{itemize}[leftmargin=0.15in]
\item \textbf{Task aggregation.} (50 tasks; examples shown in~\Cref{fig:simplerinstruct}, left). This suite assesses a model’s ability to consistently interpret and execute core tasks based on both instructions and environmental context, despite variations in visual or linguistic forms. Tasks cover phenomena such as novel verbs, multilingual expressions, diverse object references, sentence rephrasings, and OOD objects.  

\item \textbf{Situated reasoning.} (30 tasks; examples shown in~\Cref{fig:simplerinstruct}, right). Beyond \textit{task aggregation}, this suite evaluates a model’s ability to reason over contextual cues or indirect instructions and to decompose commands into sub-goals. For example, ``I want to clean the table. Pick a suitable tool for me.'' requires selecting the correct object (e.g., a sponge) from context.  

\end{itemize}

Together, by leveraging the large-scale real-world training dataset, \textbf{SimplerEnv-Instruct} provides a reproducible benchmark that evaluates VLA generalization to unseen tasks. It achieves an affordable evaluation cost while systematically probing both task generalization and reasoning, filling a critical gap in VLA evaluation with a diagnostic, human-interpretable, and standardized benchmark.

\section{Experiment}

\begin{table*}[t]
\centering
\caption{\textbf{Multimodal understanding.} \#Params is the size of LLM backbone. S. denotes robot state.}
\label{tab: mm}
\resizebox{1\textwidth}{!}{%
    \begin{tabular}{lcccccccccccccc}
    \toprule
    \multicolumn{1}{l}{\multirow{2}{*}{Methods}} & \multicolumn{1}{c}{\multirow{2}{*}{\#Params}} & \multicolumn{7}{c}{Multi-modal Understanding Benchmarks}   & \multicolumn{6}{c}{VQA Benchmarks} \\ \cmidrule(lr){3-8}  \cmidrule(lr){9-15} 
    \multicolumn{1}{c}{}       & \multicolumn{1}{c}{}          & \multicolumn{1}{c}{MMMU$^{\text{Val}}$} & \multicolumn{1}{c}{MM-Vet} & \multicolumn{1}{c}{MMStar} & \multicolumn{1}{c}{MME$^P$} & \multicolumn{1}{c}{OCRBench} & \multicolumn{1}{c}{HallB} & \multicolumn{1}{c}{MMB} & \multicolumn{1}{c}{TextVQA} & \multicolumn{1}{c}{DocVQA} & \multicolumn{1}{c}{InfoVQA} & \multicolumn{1}{c}{AI2D} & \multicolumn{1}{c}{ChartQA} & \multicolumn{1}{c}{RWQA} \\ \midrule
    Bunny~\citep{bunny}          & 8B & 43.4 & 39.1 & 45.4 & \textbf{1987.7} & 444 & 37.7 & 72.9 & - & - & - & 69.4 & 30.1 & 60.4 \\
    PaliGemma~\citep{beyer2024paligemma} & 2B & 34.9 & 33.1& 48.3 & 1686.1 & 614 & 32.2 & 65.6 & 68.1 & 74.0 & 34.0 & 68.3 & 33.1 & 55.2 \\
    Eagle2~\citep{li2025eagle}   & 1.5B & 43.1 & \textbf{53.8} & \textbf{56.4} & 1572.1 & \underline{818} & \textbf{45.8} & 74.9 & 79.1 & \underline{88.0} & 65.8 & \textbf{79.3} & \underline{82.3} & \underline{63.1} \\
    Qwen2-VL~\citep{qwen2vl}     & 1.5B & 41.1 & 51.5 & 48.0 & \underline{1872.0} & 809 & 41.7 & 74.9 & 74.9 & \textbf{88.6} & 61.4 & 74.7 & 73.5 & 62.9 \\
    \midrule
    OpenVLA~\citep{openvla}      & 7B & 0.0 & 0.0 & 0.0 & 0.0 & 0.0 & 0.0 & 0.0 & 0.0 & 0.0 & 0.0 & 0.0 & 0.0 & 0.0 \\
    OpenVLA (FT)                & 7B & 26.0 & 9.1 & 28.2 & 87.6 & 2.5 & 8.4 & 18.9 & 2.5 & 29.2 & 43.4 & 35.8 & 1.4 & 47.2 \\
    ECoT~\citep{ecot}            & 7B & 16.2 & 0.0 & 19.1 & 0.0 & 0.0 & 3.1 & 0.9 & 0.0 & 2.2 & 0.0 & 0.0 & 0.0 & 29.8 \\
    ChatVLA~\cite{zhou2025chatvla} & 1.5B & 37.4 & - & 47.2 & 1435.2 & 729 & 39.9 & 69.0 & 71.2 & 83.3 & 53.3 & 67.6 & 59.9 & 57.0 \\
    Magma~\citep{magma}  & 8B & 38.8 & 34.1 & 41.3 & 1496.5 & 518 & 38.0 & 69.7 & 66.5 & 65.4 & 45.2 & 66.1 & 61.8 & 56.5 \\
    \midrule
    \textbf{InstructVLA-Generalist}  & 1.5B & \textbf{44.2} & 51.7 & \underline{56.2} & 1529.6 & 814 & \underline{45.6} & \underline{76.1} & \underline{77.7} & 85.8 & \textbf{63.7} & \underline{79.1} & 81.7 & \underline{63.1} \\
    \textbf{InstructVLA-Generalist(S.)}  & 1.5B & \underline{43.8} & \textbf{54.0} & 56.0 & 1548.0 & \textbf{829} & 42.8 & \textbf{76.3} & \textbf{78.2} & 86.0 & \textbf{63.7} & 78.9 & \textbf{82.9} & \textbf{63.5} \\
    \bottomrule
    \end{tabular}
}
\vspace{8pt}
\caption{\textbf{Robotic manipulation.} Google and WidowX Robot denote two embodiments in SimplerEnv. For SimplerEnv-Instruct, we focus on two reasoning levels instead of embodiments. Magma$^{\dagger}$ denotes evaluation with sampling. The results of InstructVLA are averaged over three random seeds.}
\label{tab: manip}
\resizebox{1\textwidth}{!}{%
    \begin{tabular}{lccccccccccccccc}
    \toprule
    \multirow{3}{*}{Methods} & \multicolumn{8}{c}{\textbf{Google Robot}} & \multicolumn{3}{c}{\textbf{WidowX Robot}} & \multirow{3}{*}{\textbf{Avg}} & \multicolumn{3}{c}{\textbf{SimplerEnv-Instruct}} \\ \cmidrule(lr){2-12} \cmidrule(lr){14-16} 
     & \multicolumn{2}{c}{\begin{tabular}[c]{@{}c@{}}Open/Close \\ Drawer\end{tabular}} & \multicolumn{2}{c}{\begin{tabular}[c]{@{}c@{}}Put in \\ Drawer\end{tabular}} & \multicolumn{2}{c}{\begin{tabular}[c]{@{}c@{}}Pick \\ Coke Can\end{tabular}} & \multicolumn{2}{c}{\begin{tabular}[c]{@{}c@{}}Move\\ Near\end{tabular}} & \begin{tabular}[c]{@{}c@{}}Put \\ Spoon\end{tabular} & \begin{tabular}[c]{@{}c@{}}Put \\ Carrot\end{tabular} & \begin{tabular}[c]{@{}c@{}}Stack \\ Blocks\end{tabular} &  & \multicolumn{1}{c}{\multirow{2}{*}{\begin{tabular}[c]{@{}c@{}}Task\\ Aggregation\end{tabular}}} & \multicolumn{1}{c}{\multirow{2}{*}{\begin{tabular}[c]{@{}c@{}}Situated\\ Reasoning\end{tabular}}} & \multicolumn{1}{c}{\multirow{2}{*}{\textbf{Avg}}} \\ \cmidrule(lr){2-12}%  \cmidrule(lr){14-16}
     & VM & VA & VM & VA & VM & VA & VM & VA & \multicolumn{3}{c}{VM} &  & \multicolumn{1}{c}{} & \multicolumn{1}{c}{} & \multicolumn{1}{c}{} \\ \midrule
    RT-1-X~\citep{open_x_embodiment} & 59.7 & 29.4 & 21.3 & 10.1 & 56.7 & 49.0 & 31.7 & 32.3 & 0.0 & 4.2 & 0.0 & 26.8 & - & - & - \\ 
    RT-2-X~\citep{open_x_embodiment} & 25.0 & 35.5 & 3.7 & 20.6 & 78.7 & 82.3 & 77.9 & 79.2 & - & - & - & - & - & - & - \\ 
    RoboVLMs-2B~{(S.)}~\citep{robovlms} & 43.5 & 10.6 & 27.8 & 0.0 & 77.3 & 75.6 & 61.7 & 60.0 & 45.8 & 20.8 & 4.2 & 38.8 & - & - & - \\ 
    OpenVLA-7B~\citep{openvla} & 63.0 & 28.8 & 0.0 & 0.0 & 18.0 & 60.8 & 56.3 & 67.7 & 4.2 & 0.0 & 0.0 & 27.2 & 14.8 & 13.6 & 14.2 \\ 
    SpatialVLA-3B~\citep{qu2025spatialvla} & 57.4 & 41.8 & 0.9 & 9.1 & 86.0 & 88.0 & 77.9 & 72.7 & 16.7 & 25.0 & 29.2 & 45.9 & 23.6 & 9.8 & 16.5 \\
    GR00T-N1.5-3B~{(S.)}~\citep{bjorck2025gr00t} & 27.8 & 13.2 & 7.4 & 2.2 & 51.7 & 63.6 & 51.0 & 54.0 & 62.5 & 45.8 & 16.7 & 36.0 & - & - & - \\
    $\pi_0$-3B~{(S.)}~\citep{pi_0} & 64.8 & 48.4 & 13.9 & 15.4 & 70.3 & 44.7 & 41.0 & 35.5 & 37.5 & 50.0 & 37.5 & 41.7 & 12.1 & 11.8 & 12.0 \\
    \textbf{InstructVLA-Expert} & 52.3 & 61.7 & 50.3 & 33.1 & 79.6 & 92.3 & 68.3 & 71.9 & 43.1 & 40.4 & 9.7 & \underline{50.9} & $21.6\pm1.4$ & $12.9\pm0.4$ & \underline{17.3} \\ 
    \textbf{InstructVLA-Expert(S.)} & 46.8 & 54.1 & 45.7 & 70.0 & 96.0 & 95.9 & 79.7 & 82.4 & 61.1 & 54.2 & 36.1 & \textbf{61.2} & $20.9\pm0.3$ & $20.5\pm1.0$ & \textbf{20.7} \\ 
    \midrule
    Magma-8B~\citep{magma} & 9.7 & 5.8 & 0.0 & 0.0 & 46.0 & 46.4 & 60.0 & 82.0 & 45.8 & 33.3 & 8.3 & 30.5 & 15.5 & 9.9 & 12.7 \\ 
    Magma-8B$^{\dagger}$~\citep{magma} & 56.0 & 53.4 & 6.4 & 18.5 & 83.7 & 68.8 & 65.4 & 65.7 & 35.5 & 31.0 & 12.7 & 43.6 & 26.2 & 21.4 & 23.8 \\ 
    OpenVLA (FT) 7B & 63.9 & 42.6 & 3.7 & 6.9 & 62.3 & 88.7 & 65.8 & 67.7 & 12.5 & 33.3 & 4.2 & 39.0 & 28.3 & 19.5 & 23.9 \\ 

    OpenVLA (FT\&GPT) & - & - & - & - & - & - & - & - & - & - & - & - & 38.8 & 32.4 & 35.6 \\ 
    \midrule
    \textbf{InstructVLA-Generalist} & 64.5 & 61.7 & 38.3 & 27.5 & 81.7 & 91.8 & 55.8 & 69.7 & 31.9 & 34.7 & 12.5 & \underline{49.7} & $43.6\pm1.4$ & $48.8\pm0.8$ & \underline{46.2} \\ 
    \textbf{InstructVLA-Generalist(S.)} & 39.8 & 51.1 & 45.7 & 57.3 & 91.0 & 93.0 & 71.7 & 78.3 & 62.4 & 48.6 & 15.3 & \textbf{54.9} & $48.2\pm1.3$ & $45.6\pm0.5$ & \textbf{46.9} \\ 
    \bottomrule
    \end{tabular}
 }
 % \vspace{-0.5cm}
\end{table*}

\noindent\textbf{Benchmarks.} \textit{(a) Multimodal:} We adopt automatic evaluation from VLMEvalKit~\citep{duan2024vlmevalkit}, as detailed in~\Cref{sec: multimodal benchmarks}. \textit{(b) SimplerEnv:} This benchmark~\citep{simpleenv} provides real-to-sim evaluation on large-scale manipulation datasets, incorporating visual matching and variance aggregation to assess generalization. \textit{(c) SimplerEnv-Instruct:} As described in~\Cref{sec: SimplerEnv-Instruct}, this extension of SimplerEnv introduces novel objects, tasks, and instructions, offering a broader testbed for evaluating instruction generalization in VLAs. In addition, we assess embodied understanding in~\Cref{sec: vl evaluation} and manipulation performance on the LIBERO~\citep{Libero} benchmark in~\Cref{sec: libero}.

\noindent\textbf{Training details.}
The VLM is trained with a resolution of $448 \times 448$ following~\citet{li2025eagle}, while the action expert operates at $224 \times 224$ as in~\citep{openvla}, using a fixed learning rate of 5e-5 without warm-up. The action expert employs a 12-layer transformer backbone with a hidden size of 768. Following~\citet{pi_0}, a $\beta$ distribution is used to enhance accuracy on the noisier time steps. During Stage 2 finetuning, manipulation and multimodal understanding are trained in an interleaved manner. Owing to our training paradigm, multimodal capabilities are preserved easily. We adopt a 1:7 multimodal-to-action training ratio, twice the ratio in ECoT and ChatVLA (1:3), reducing the additional computation needed to maintain multimodal ability. More details are in \Cref{sec: Model Design and Training Details}.

\noindent\textbf{Baselines.} We categorize the baselines into three groups: (1) \textit{Multimodal VLMs}, including Bunny\citep{bunny}, PaliGemma~\citep{beyer2024paligemma}, Eagle2~\citep{li2025eagle}, and Qwen2-VL~\citep{qwen2vl}; (2) \textit{VLA models}, including RT-1-X and RT-2-X~\citep{open_x_embodiment}, RoboVLMs~\citep{robovlms}, SpatialVLA~\citep{qu2025spatialvla}, $\pi_{0}$~\citep{pi_0}, GR00T-N1.5~\citep{bjorck2025gr00t}, and OpenVLA~\citep{openvla};
(3) \textit{Generalist VLA models}, including Magma~\citep{magma}, OpenVLA fine-tuned (FT) from generalist pretrained model on both robotic and multimodal data, and ECoT(Bridge)~\citep{ecot}. During evaluation, InstructVLA and other baselines use a temperature of 0 without sampling to expedite generation. We re-evaluate Magma with official checkpoint\footnote{We observe a notable performance gain for Magma when using sampling. Accordingly, we report its official score on SimplerEnv and re-evaluate its performance on SimplerEnv-Instruct under the sampling setting.}. For ECoT, we report only its multimodal results due to its real-to-sim domain gap.

\subsection{Main Results}

We present our main results in~\Cref{tab: manip,tab: mm}. In~\Cref{tab: mm}, using the same generalist model InstructVLA (generalist), it not only outperforms the co-trained baseline Magma, but is also comparable to its base model Eagle2 and Bunny~(VLM data corpus). InstructVLA further demonstrates stronger embodied understanding as detailed in~\Cref{sec: vl evaluation}. In~\Cref{tab: manip}, InstructVLA (expert) outperforms the expert baseline SpatialVLA by 33.3\% on SimplerEnv. Meanwhile, InstructVLA (generalist) not only maintains strong performance on SimplerEnv’s atomic instructions but also achieves a 31.7\% {relative} improvement on SimplerEnv-Instruct over the state-of-the-art baseline (OpenVLA with GPT-4o).

However, we observe that finetuning OpenVLA on multimodal and manipulation datasets does not fully restore its original multimodal capabilities, although it does improve task performance. Its performance can be further enhanced by integrating GPT-4o as an API-based system-2 module to rephrase instructions (OpenVLA (FT\&GPT)). However, GPT-4o faces the same challenges in accurate instruction rewriting as noted in~\Cref{sec: instruction data}, and fails to outperform InstructVLA (Generalist). Methods such as Magma, which co-train both abilities of the VLM, better preserve multimodal ability, but still fail to match the performance of our approach. Although it also adapts two-stage training, ECoT relies solely on textual chain-of-thought reasoning over manipulation datasets and lacks the capability for multimodal question answering. We observe that it consistently generates manipulation-style CoT responses, without demonstrating effective instruction-following ability.

\subsection{Real-world Experiments}

\begin{figure*}[t]
    \centering
    \includegraphics[width=1\linewidth]{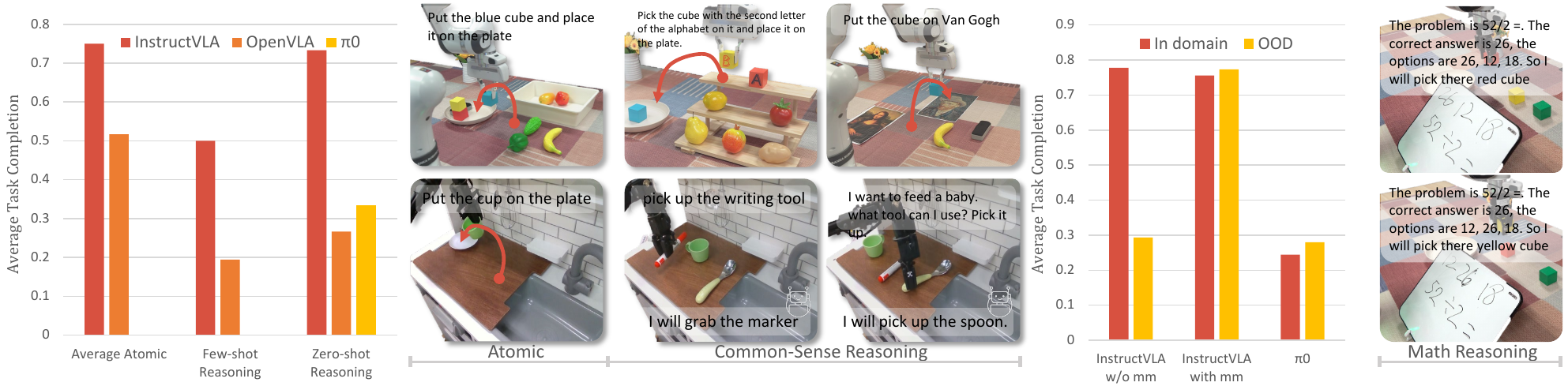}
    \vspace{-0.2cm}
    \caption{\textbf{Real-world experiments.} ``Atomic'' refers to atomic instructions. For the Kitchen and math settings, InstructVLA’s responses are presented.}
    \label{fig:realworld}
    % \vspace{-0.2cm}
\end{figure*}

To evaluate InstructVLA in real-world scenarios, we conduct zero-shot experiments on the WidowX-250 Arm and few-shot experiments on the Franka Research 3 robot, as shown in~\Cref{fig:realworld}. The few-shot tasks involve spatial pick-and-place from a rack and cluttered tabletop setting and math-centric tasks detailed in~\Cref{sec:realworld_ablation} to demonstrate the role of multimodal data. The zero-shot tasks are set in a kitchen environment following the Bridge dataset. InstructVLA is fine-tuned using the proposed training recipe, while OpenVLA is jointly trained on atomic skill and VLA-IT datasets with extra language supervision. The $\pi_0$ is finetuned using the official repository.

Each scenario includes both atomic and reasoning instructions. Atomic tasks emphasize in-domain objects and instructions with a focus on spatial generalization to assess baseline VLA capabilities. Both models perform comparably on direct in-domain instructions, but InstructVLA achieves a 23.3\% improvement over OpenVLA. For reasoning tasks such as celebrity recognition, OCR, and tool-use inference, OpenVLA shows a substantial performance drop, whereas InstructVLA outperforms it by 41.7\% in few-shot and 46.7\% in zero-shot settings. On reasoning and math tasks, InstructVLA achieves a $2.5\times$ improvement over $\pi_0$, which behaves close to random guessing. Additional ablations and experimental setups are provided in~\Cref{sec:realworld_ablation,sec: Real-world Experiments Setup}.

\subsection{Ablation Studies}

We conduct ablation studies guided by two central questions: (1)~\Cref{sec:cotraining}. How can manipulation and multimodal understanding be effectively integrated into a single model through architectural design and training strategies? (2)~\Cref{sec:transfer}. To what extent does vision-language comprehension enhance manipulation performance in complex scenarios? Through targeted ablations, we examine the impact of key architectural and training decisions on these capabilities.

\begin{figure*}[t]
    \centering
    \begin{minipage}[t]{0.38\linewidth}
        \vspace{2pt}
        \centering
        \captionof{table}{\textbf{Ablation of action expert vision design and language motion.} ``w/o Lang.'' denotes without using language motion. ``w/o FiLM'' denotes using only DINO. ``w/o DINO'' denotes action expert without the vision input.}
        \label{tab:ablation_1}
        \resizebox{\textwidth}{!}{%
            \begin{tabular}{lccc}
            \toprule
            Experts    & WidowX Bot & Google Bot & Ave. \\ \midrule
            w/o DINO        & 4.2            & 32.4         & 23.0    \\
            w/o FiLM        & 25.0           & 56.3         & 45.9    \\
            w/o Lang.       & 15.3           & 65.0         & \underline{48.4}    \\ \midrule
            InstructVLA     & 29.1           & 64.8         & \textbf{52.9}   \\ \bottomrule
            \end{tabular}
        }
    \end{minipage}
    \hfill
    \begin{minipage}[t]{0.6\linewidth}
        \vspace{0pt}
        \centering
        \includegraphics[width=\linewidth]{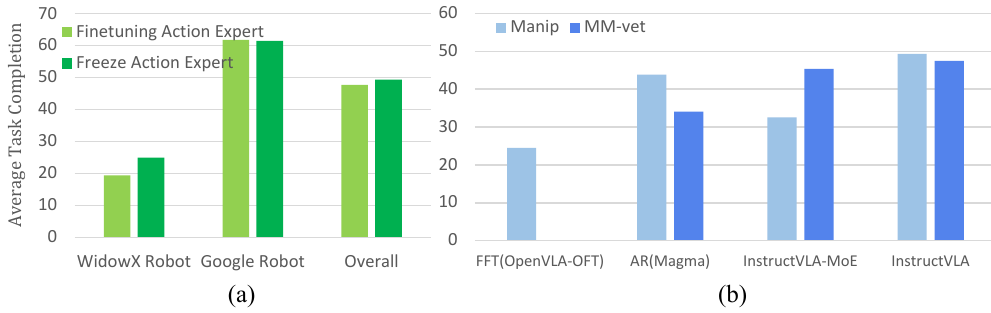}
        \vspace{-15pt}
        \caption{\textbf{Finetuning strategies.} (a) Freezing or finetuning the action head during VLA-IT training. (b) Training strategies when multimodal and manipulation tasks co-exist. ``FFT'' denotes full finetuning. ``AR'' denotes auto-regressive.}
        \label{fig:ablation_instruct}
    \end{minipage}
    % \vspace{0.3cm}
\end{figure*}

\subsubsection{Multimodal and Manipulation Co-training} 
\label{sec:cotraining}

\noindent\textbf{Strategies for multimodal and manipulation co-training.}
As shown in~\Cref{fig:ablation_instruct} (b), four paradigms are compared. (1) Following OpenVLA-OFT, FFT denotes full finetuning of the model with latent actions but without MoE adaptation and multi-stage training. With comparable computational resources, this setting yields suboptimal performance on both manipulation and understanding tasks. (2) The AR paradigm (Magma, RT-2) supports co-training but has limited performance. (3) Removing the MoE design while keeping the training paradigm preserves multimodal performance but reduces manipulation capability. (4) In contrast, InstructVLA leverages our proposed architecture and two-stage training strategy, achieving a 12.5\% improvement over Magma on SimplerEnv.

% \noindent\textbf{Effects of language motion data for pre-training.}
\noindent\textbf{Language motion helps action understanding.} As shown in~\Cref{tab:ablation_1}, introducing ``language motion'' (textual descriptions of low-level actions) supervision enhances the VLM’s ability to associate visual cues with manipulation primitives, leading to a 9.3\% improvement in overall success rate.

\noindent\textbf{Enhanced expert perception helps policy learning.} Incorporating richer perception into the action expert is efficient due to its compact design compared to the VLM backbone. As shown in~\Cref{tab:ablation_1}, while the base VLM offers general visual understanding, fine-grained perception for manipulation tasks demands richer representations. Removing the DINOv2-based ViT encoder from the action expert results in a 50.0\% performance drop, highlighting its critical role in capturing task-relevant visual cues. Incorporating FiLM to the ViT encoder yields a further 15.3\% improvement by modulating visual features with latent actions. As shown in~\Cref{tab: manip} the expert model with robot state generally performs better.

% ----------- Figure (single column) -----------
\begin{figure}[t]
    \centering
    \includegraphics[width=\linewidth]{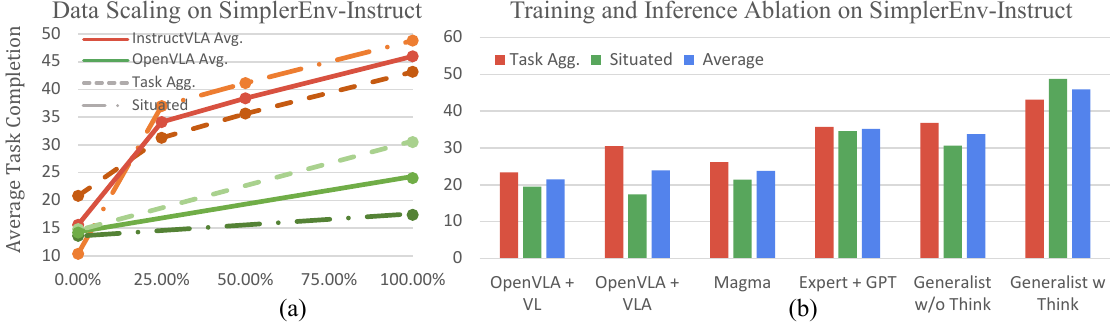}
    \caption{\textbf{Data scaling and multimodal training.} Impact of scaling and training strategies on manipulation with multimodal reasoning.}
    \label{fig:ablation_instruct_1}
\end{figure}

\subsubsection{Multimodal Ability Transfers to Manipulation}
\label{sec:transfer}

\noindent\textbf{VLM-only instruction tuning handles situdated understanding.} As shown in~\Cref{fig:ablation_instruct}(a), we examine the effect of VLA instruction tuning by comparing two configurations: (1) finetuning only the VLM, and (2) jointly finetuning both the VLM and the action expert. Freezing the action expert achieves performance comparable to joint finetuning while substantially reducing the number of trainable parameters. This suggests that InstructVLA can effectively adapt complex and situated manipulation tasks by fine-tuning only the VLM, without altering the pretrained action expert.

\begin{wraptable}{r}{0.34\linewidth}
    \centering
    \vspace{-10pt}
    \caption{\textbf{Effect of data dievrsity.} ``T.A.'' denotes task aggregation, and ``S.R.'' denotes situated reasoning on SimplerEnv-Instruct.}
    \label{tab:ablation_qa}
    \resizebox{\linewidth}{!}{%
        \begin{tabular}{cccc}
        \toprule
        QA \& Cap.  & T.A. & S.R. & Ave. \\ \midrule
        \ding{55}   & 40.7 & 42.7 & 41.7 \\
        \ding{51}   & 43.6 & 48.8 & 46.2 \\ \bottomrule
        \end{tabular}
    }
    \vspace{-5pt}
\end{wraptable}

\noindent\textbf{How VLA-IT scale and multimodal diversity affect reasoning-guided manipulation.} As shown in~\Cref{fig:ablation_instruct_1}(a), we first evaluate the \textbf{scaling behavior} of VLA-IT annotations on the SimplerEnv-Instruct benchmark. Situated reasoning tasks, which require grounding objects and goals in context, benefit most from larger annotation sets, highlighting the bootstrapped reasoning abilities inherited from VLMs. In contrast, pretrained OpenVLA fine-tuned on VLA-IT gains primarily from increased instruction diversity but shows limited improvement on situated reasoning tasks due to catastrophic forgetting of VL capabilities. These findings suggest that two-stage methods such as ECoT may be insufficient for fully leveraging the multimodal capacity of VLMs. We also examine the \textbf{effect of annotation diversity}, as shown in~\Cref{tab:ablation_qa}, where adding QA and captioning improves generalization of InstructVLA by 10.8\%. Additional ablations are provided in~\Cref{sec: openvla data ablation}.

\noindent\textbf{Ablating strategies for incorporating reasoning into manipulation.} As shown in~\Cref{fig:ablation_instruct_1} (b), (1) Simply combining manipulation and general multimodal ability through co-training does not yield significant benefits. Magma, despite co-training on multimodal datasets, shows limited transfer of vision-language capability to reasoning tasks on SimplerEnv-Instruct. Although OpenVLA suffers from catastrophic forgetting when finetuned with VLA-IT corpus, it still achieves better performance than Magma. (2) Multimodal ability can implicitly benefit manipulation when preserved through embodied reasoning annotation. Our generalist model, trained on the VLA-IT corpus, surpasses fine-tuned OpenVLA and Magma on the SimplerEnv-Instruct benchmark, even without explicit textual reasoning (\textit{Think}). (3) Explicit textual reasoning further enhances manipulation. Enabling thinking in the generalist model brings a 36.1\% performance gain over direct instruction execution and even outperforms InstructVLA-expert paired with GPT-4o as an external interpreter. Further analysis of the role of thinking is presented in~\Cref{sec: sub dis}.

\section{Conclusion}

We present InstructVLA, a unified VLA model that integrates multimodal reasoning and action generation. We further demonstrate how the embodied understanding ability can directly benefit the maipulation tasks. Our data and training pipeline enables leading performance across manipulation tasks, multimodal benchmarks, and real-world deployments, paving the way for more generalizable, interpretable, and interactive robots.

\bibliography{iclr2026_conference}

@String(AAAI = {AAAI})

@article{li2025cronusvla,
  title={CronusVLA: Transferring Latent Motion Across Time for Multi-Frame Prediction in Manipulation},
  author={Li, Hao and Yang, Shuai and Chen, Yilun and Tian, Yang and Yang, Xiaoda and Chen, Xinyi and Wang, Hanqing and Wang, Tai and Zhao, Feng and Lin, Dahua and others},
  journal={arXiv preprint arXiv:2506.19816},
  year={2025}
}

@inproceedings{huang2025roboground,
  title={RoboGround: Robotic Manipulation with Grounded Vision-Language Priors},
  author={Huang, Haifeng and Chen, Xinyi and Chen, Yilun and Li, Hao and Han, Xiaoshen and Wang, Zehan and Wang, Tai and Pang, Jiangmiao and Zhao, Zhou},
  booktitle={Proceedings of the IEEE/CVF Conference on Computer Vision and Pattern Recognition},
  pages={22540--22550},
  year={2025}
}

@article{li2026robointer,
  title={RoboInter: A Holistic Intermediate Representation Suite Towards Robotic Manipulation},
  author={Li, Hao and Wang, Ziqin and Ding, Zi-han and Yang, Shuai and Chen, Yilun and Tian, Yang and Hu, Xiaolin and Wang, Tai and Lin, Dahua and Zhao, Feng and others},
  journal={arXiv preprint arXiv:2602.09973},
  year={2026}
}

@article{robovlms,
  title={Towards generalist robot policies: What matters in building vision-language-action models},
  author={Li, Xinghang and Li, Peiyan and Liu, Minghuan and Wang, Dong and Liu, Jirong and Kang, Bingyi and Ma, Xiao and Kong, Tao and Zhang, Hanbo and Liu, Huaping},
  journal={arXiv preprint arXiv:2412.14058},
  year={2024}
}

@article{simpleenv,
  title={Evaluating Real-World Robot Manipulation Policies in Simulation},
  author={Li, Xuanlin and Hsu, Kyle and Gu, Jiayuan and Pertsch, Karl and Mees, Oier and Walke, Homer Rich and Fu, Chuyuan and Lunawat, Ishikaa and Sieh, Isabel and Kirmani, Sean and others},
  journal={arXiv preprint arXiv:2405.05941},
  year={2024}
}

@article{GPT-4,
  title={Gpt-4 technical report},
  author={OpenAI},
  journal={arXiv:2303.08774},
  year={2023}
}

@article{RT-1,
  title={Rt-1: Robotics transformer for real-world control at scale},
  author={Brohan, Anthony and Brown, Noah and Carbajal, Justice and Chebotar, Yevgen and Dabis, Joseph and Finn, Chelsea and Gopalakrishnan, Keerthana and Hausman, Karol and Herzog, Alex and Hsu, Jasmine and others},
  journal={arXiv preprint arXiv:2212.06817},
  year={2022}
}

@article{RT-2,
  title={Rt-2: Vision-language-action models transfer web knowledge to robotic control},
  author={Brohan, Anthony and Brown, Noah and Carbajal, Justice and Chebotar, Yevgen and Chen, Xi and Choromanski, Krzysztof and Ding, Tianli and Driess, Danny and Dubey, Avinava and Finn, Chelsea and others},
  journal={arXiv preprint arXiv:2307.15818},
  year={2023}
}

@inproceedings{octo,
    title={Octo: An Open-Source Generalist Robot Policy},
    author = {{Octo Model Team} and Dibya Ghosh and Homer Walke and Karl Pertsch and Kevin Black and Oier Mees and Sudeep Dasari and Joey Hejna and Charles Xu and Jianlan Luo and Tobias Kreiman and {You Liang} Tan and Pannag Sanketi and Quan Vuong and Ted Xiao and Dorsa Sadigh and Chelsea Finn and Sergey Levine},
    booktitle = {Proceedings of Robotics: Science and Systems},
    year = {2024},
}

@article{Libero,
  title={Libero: Benchmarking knowledge transfer for lifelong robot learning},
  author={Liu, Bo and Zhu, Yifeng and Gao, Chongkai and Feng, Yihao and Liu, Qiang and Zhu, Yuke and Stone, Peter},
  journal={Advances in Neural Information Processing Systems},
  volume={36},
  year={2024}
}

@article{Bridge_data,
  title={Bridge data: Boosting generalization of robotic skills with cross-domain datasets},
  author={Ebert, Frederik and Yang, Yanlai and Schmeckpeper, Karl and Bucher, Bernadette and Georgakis, Georgios and Daniilidis, Kostas and Finn, Chelsea and Levine, Sergey},
  journal={arXiv preprint arXiv:2109.13396},
  year={2021}
}

@article{khazatsky2024droid,
  title={Droid: A large-scale in-the-wild robot manipulation dataset},
  author={Khazatsky, Alexander and Pertsch, Karl and Nair, Suraj and Balakrishna, Ashwin and Dasari, Sudeep and Karamcheti, Siddharth and Nasiriany, Soroush and Srirama, Mohan Kumar and Chen, Lawrence Yunliang and Ellis, Kirsty and others},
  journal={arXiv preprint arXiv:2403.12945},
  year={2024}
}

@inproceedings{siglip,
  title={Sigmoid loss for language image pre-training},
  author={Zhai, Xiaohua and Mustafa, Basil and Kolesnikov, Alexander and Beyer, Lucas},
  booktitle={Proceedings of the IEEE/CVF International Conference on Computer Vision},
  pages={11975--11986},
  year={2023}
}

@article{oquab2023dinov2,
  title={Dinov2: Learning robust visual features without supervision},
  author={Oquab, Maxime and Darcet, Timoth{\'e}e and Moutakanni, Th{\'e}o and Vo, Huy and Szafraniec, Marc and Khalidov, Vasil and Fernandez, Pierre and Haziza, Daniel and Massa, Francisco and El-Nouby, Alaaeldin and others},
  journal={arXiv preprint arXiv:2304.07193},
  year={2023}
}

@misc{open_x_embodiment,
title={Open {X-E}mbodiment: Robotic Learning Datasets and {RT-X} Models},
author = {Open X-Embodiment Collaboration and Abby O'Neill and Abdul Rehman and Abhinav Gupta and Abhiram Maddukuri and Abhishek Gupta and Abhishek Padalkar and Abraham Lee and Acorn Pooley and Agrim Gupta and Ajay Mandlekar and Ajinkya Jain and Albert Tung and Alex Bewley and Alex Herzog and Alex Irpan and Alexander Khazatsky and Anant Rai and Anchit Gupta and Andrew Wang and Andrey Kolobov and Anikait Singh and Animesh Garg and Aniruddha Kembhavi and Annie Xie and Anthony Brohan and Antonin Raffin and Archit Sharma and Arefeh Yavary and Arhan Jain and Ashwin Balakrishna and Ayzaan Wahid and Ben Burgess-Limerick and Beomjoon Kim and Bernhard Schölkopf and Blake Wulfe and Brian Ichter and Cewu Lu and Charles Xu and Charlotte Le and Chelsea Finn and Chen Wang and Chenfeng Xu and Cheng Chi and Chenguang Huang and Christine Chan and Christopher Agia and Chuer Pan and Chuyuan Fu and Coline Devin and Danfei Xu and Daniel Morton and Danny Driess and Daphne Chen and Deepak Pathak and Dhruv Shah and Dieter Büchler and Dinesh Jayaraman and Dmitry Kalashnikov and Dorsa Sadigh and Edward Johns and Ethan Foster and Fangchen Liu and Federico Ceola and Fei Xia and Feiyu Zhao and Felipe Vieira Frujeri and Freek Stulp and Gaoyue Zhou and Gaurav S. Sukhatme and Gautam Salhotra and Ge Yan and Gilbert Feng and Giulio Schiavi and Glen Berseth and Gregory Kahn and Guangwen Yang and Guanzhi Wang and Hao Su and Hao-Shu Fang and Haochen Shi and Henghui Bao and Heni Ben Amor and Henrik I Christensen and Hiroki Furuta and Homanga Bharadhwaj and Homer Walke and Hongjie Fang and Huy Ha and Igor Mordatch and Ilija Radosavovic and Isabel Leal and Jacky Liang and Jad Abou-Chakra and Jaehyung Kim and Jaimyn Drake and Jan Peters and Jan Schneider and Jasmine Hsu and Jay Vakil and Jeannette Bohg and Jeffrey Bingham and Jeffrey Wu and Jensen Gao and Jiaheng Hu and Jiajun Wu and Jialin Wu and Jiankai Sun and Jianlan Luo and Jiayuan Gu and Jie Tan and Jihoon Oh and Jimmy Wu and Jingpei Lu and Jingyun Yang and Jitendra Malik and João Silvério and Joey Hejna and Jonathan Booher and Jonathan Tompson and Jonathan Yang and Jordi Salvador and Joseph J. Lim and Junhyek Han and Kaiyuan Wang and Kanishka Rao and Karl Pertsch and Karol Hausman and Keegan Go and Keerthana Gopalakrishnan and Ken Goldberg and Kendra Byrne and Kenneth Oslund and Kento Kawaharazuka and Kevin Black and Kevin Lin and Kevin Zhang and Kiana Ehsani and Kiran Lekkala and Kirsty Ellis and Krishan Rana and Krishnan Srinivasan and Kuan Fang and Kunal Pratap Singh and Kuo-Hao Zeng and Kyle Hatch and Kyle Hsu and Laurent Itti and Lawrence Yunliang Chen and Lerrel Pinto and Li Fei-Fei and Liam Tan and Linxi "Jim" Fan and Lionel Ott and Lisa Lee and Luca Weihs and Magnum Chen and Marion Lepert and Marius Memmel and Masayoshi Tomizuka and Masha Itkina and Mateo Guaman Castro and Max Spero and Maximilian Du and Michael Ahn and Michael C. Yip and Mingtong Zhang and Mingyu Ding and Minho Heo and Mohan Kumar Srirama and Mohit Sharma and Moo Jin Kim and Naoaki Kanazawa and Nicklas Hansen and Nicolas Heess and Nikhil J Joshi and Niko Suenderhauf and Ning Liu and Norman Di Palo and Nur Muhammad Mahi Shafiullah and Oier Mees and Oliver Kroemer and Osbert Bastani and Pannag R Sanketi and Patrick "Tree" Miller and Patrick Yin and Paul Wohlhart and Peng Xu and Peter David Fagan and Peter Mitrano and Pierre Sermanet and Pieter Abbeel and Priya Sundaresan and Qiuyu Chen and Quan Vuong and Rafael Rafailov and Ran Tian and Ria Doshi and Roberto Mart{'i}n-Mart{'i}n and Rohan Baijal and Rosario Scalise and Rose Hendrix and Roy Lin and Runjia Qian and Ruohan Zhang and Russell Mendonca and Rutav Shah and Ryan Hoque and Ryan Julian and Samuel Bustamante and Sean Kirmani and Sergey Levine and Shan Lin and Sherry Moore and Shikhar Bahl and Shivin Dass and Shubham Sonawani and Shubham Tulsiani and Shuran Song and Sichun Xu and Siddhant Haldar and Siddharth Karamcheti and Simeon Adebola and Simon Guist and Soroush Nasiriany and Stefan Schaal and Stefan Welker and Stephen Tian and Subramanian Ramamoorthy and Sudeep Dasari and Suneel Belkhale and Sungjae Park and Suraj Nair and Suvir Mirchandani and Takayuki Osa and Tanmay Gupta and Tatsuya Harada and Tatsuya Matsushima and Ted Xiao and Thomas Kollar and Tianhe Yu and Tianli Ding and Todor Davchev and Tony Z. Zhao and Travis Armstrong and Trevor Darrell and Trinity Chung and Vidhi Jain and Vikash Kumar and Vincent Vanhoucke and Wei Zhan and Wenxuan Zhou and Wolfram Burgard and Xi Chen and Xiangyu Chen and Xiaolong Wang and Xinghao Zhu and Xinyang Geng and Xiyuan Liu and Xu Liangwei and Xuanlin Li and Yansong Pang and Yao Lu and Yecheng Jason Ma and Yejin Kim and Yevgen Chebotar and Yifan Zhou and Yifeng Zhu and Yilin Wu and Ying Xu and Yixuan Wang and Yonatan Bisk and Yongqiang Dou and Yoonyoung Cho and Youngwoon Lee and Yuchen Cui and Yue Cao and Yueh-Hua Wu and Yujin Tang and Yuke Zhu and Yunchu Zhang and Yunfan Jiang and Yunshuang Li and Yunzhu Li and Yusuke Iwasawa and Yutaka Matsuo and Zehan Ma and Zhuo Xu and Zichen Jeff Cui and Zichen Zhang and Zipeng Fu and Zipeng Lin},
howpublished  = {\url{https://arxiv.org/abs/2310.08864}},
year = {2023},
}

@article{rt-trajectory,
  title={Rt-trajectory: Robotic task generalization via hindsight trajectory sketches},
  author={Gu, Jiayuan and Kirmani, Sean and Wohlhart, Paul and Lu, Yao and Arenas, Montserrat Gonzalez and Rao, Kanishka and Yu, Wenhao and Fu, Chuyuan and Gopalakrishnan, Keerthana and Xu, Zhuo and others},
  journal={arXiv preprint arXiv:2311.01977},
  year={2023}
}

@article{touvron2023llama,
  title={Llama: Open and efficient foundation language models},
  author={Touvron, Hugo and Lavril, Thibaut and Izacard, Gautier and Martinet, Xavier and Lachaux, Marie-Anne and Lacroix, Timoth{\'e}e and Rozi{\`e}re, Baptiste and Goyal, Naman and Hambro, Eric and Azhar, Faisal and others},
  journal={arXiv preprint arXiv:2302.13971},
  year={2023}
}

@article{rh20t,
  title={RH20T-P: A Primitive-Level Robotic Dataset Towards Composable Generalization Agents},
  author={Chen, Zeren and Shi, Zhelun and Lu, Xiaoya and He, Lehan and Qian, Sucheng and Fang, Hao Shu and Yin, Zhenfei and Ouyang, Wanli and Shao, Jing and Qiao, Yu and others},
  journal={arXiv preprint arXiv:2403.19622},
  year={2024}
}

@article{qwen,
  title={Qwen-vl: A frontier large vision-language model with versatile abilities},
  author={Bai, Jinze and Bai, Shuai and Yang, Shusheng and Wang, Shijie and Tan, Sinan and Wang, Peng and Lin, Junyang and Zhou, Chang and Zhou, Jingren},
  journal={arXiv preprint arXiv:2308.12966},
  year={2023}
}

@inproceedings{clip,
  title={Learning transferable visual models from natural language supervision},
  author={Radford, Alec and Kim, Jong Wook and Hallacy, Chris and Ramesh, Aditya and Goh, Gabriel and Agarwal, Sandhini and Sastry, Girish and Askell, Amanda and Mishkin, Pamela and Clark, Jack and others},
  booktitle={International Conference on Machine Learning},
  pages={8748--8763},
  year={2021},
  organization={PMLR}
}

@article{pi_0,
  title={$\backslash pi_0 $: A Vision-Language-Action Flow Model for General Robot Control},
  author={Black, Kevin and Brown, Noah and Driess, Danny and Esmail, Adnan and Equi, Michael and Finn, Chelsea and Fusai, Niccolo and Groom, Lachy and Hausman, Karol and Ichter, Brian and others},
  journal={arXiv preprint arXiv:2410.24164},
  year={2024}
}

@article{pi05,
  title={pi-0.5: a Vision-Language-Action Model with Open-World Generalization},
  author={Intelligence, Physical and Black, Kevin and Brown, Noah and Darpinian, James and Dhabalia, Karan and Driess, Danny and Esmail, Adnan and Equi, Michael and Finn, Chelsea and Fusai, Niccolo and others},
  journal={arXiv preprint arXiv:2504.16054},
  year={2025}
}

@article{liu2024robomamba,
  title={Robomamba: Multimodal state space model for efficient robot reasoning and manipulation},
  author={Liu, Jiaming and Liu, Mengzhen and Wang, Zhenyu and Lee, Lily and Zhou, Kaichen and An, Pengju and Yang, Senqiao and Zhang, Renrui and Guo, Yandong and Zhang, Shanghang},
  journal={arXiv e-prints},
  pages={arXiv--2406},
  year={2024}
}

@article{openvla,
  title={OpenVLA: An Open-Source Vision-Language-Action Model},
  author={Kim, Moo Jin and Pertsch, Karl and Karamcheti, Siddharth and Xiao, Ted and Balakrishna, Ashwin and Nair, Suraj and Rafailov, Rafael and Foster, Ethan and Lam, Grace and Sanketi, Pannag and others},
  journal={arXiv preprint arXiv:2406.09246},
  year={2024}
}

@article{cogact,
  title={Cogact: A foundational vision-language-action model for synergizing cognition and action in robotic manipulation},
  author={Li, Qixiu and Liang, Yaobo and Wang, Zeyu and Luo, Lin and Chen, Xi and Liao, Mozheng and Wei, Fangyun and Deng, Yu and Xu, Sicheng and Zhang, Yizhong and others},
  journal={arXiv preprint arXiv:2411.19650},
  year={2024}
}

@article{rth,
  title={Rt-h: Action hierarchies using language},
  author={Belkhale, Suneel and Ding, Tianli and Xiao, Ted and Sermanet, Pierre and Vuong, Quon and Tompson, Jonathan and Chebotar, Yevgen and Dwibedi, Debidatta and Sadigh, Dorsa},
  journal={arXiv preprint arXiv:2403.01823},
  year={2024}
}

@article{hirobot,
  title={Hi robot: Open-ended instruction following with hierarchical vision-language-action models},
  author={Shi, Lucy Xiaoyang and Ichter, Brian and Equi, Michael and Ke, Liyiming and Pertsch, Karl and Vuong, Quan and Tanner, James and Walling, Anna and Wang, Haohuan and Fusai, Niccolo and others},
  journal={arXiv preprint arXiv:2502.19417},
  year={2025}
}

@article{flowmatching,
  title={Flow matching for generative modeling},
  author={Lipman, Yaron and Chen, Ricky TQ and Ben-Hamu, Heli and Nickel, Maximilian and Le, Matt},
  journal={arXiv preprint arXiv:2210.02747},
  year={2022}
}

@inproceedings{lora,
  title={Lora: Low-rank adaptation of large language models.},
  author={Hu, Edward J and Shen, Yelong and Wallis, Phillip and Allen-Zhu, Zeyuan and Li, Yuanzhi and Wang, Shean and Wang, Lu and Chen, Weizhu and others},
  booktitle={International Conference on Learning Representations},
  year={2022}
}

@article{li2025eagle,
  title={Eagle 2: Building Post-Training Data Strategies from Scratch for Frontier Vision-Language Models},
  author={Li, Zhiqi and Chen, Guo and Liu, Shilong and Wang, Shihao and VS, Vibashan and Ji, Yishen and Lan, Shiyi and Zhang, Hao and Zhao, Yilin and Radhakrishnan, Subhashree and others},
  journal={arXiv preprint arXiv:2501.14818},
  year={2025}
}

@article{act,
  title={Learning fine-grained bimanual manipulation with low-cost hardware},
  author={Zhao, Tony Z and Kumar, Vikash and Levine, Sergey and Finn, Chelsea},
  journal={arXiv preprint arXiv:2304.13705},
  year={2023}
}

@article{moe,
  title={Mixture-of-experts with expert choice routing},
  author={Zhou, Yanqi and Lei, Tao and Liu, Hanxiao and Du, Nan and Huang, Yanping and Zhao, Vincent and Dai, Andrew M and Le, Quoc V and Laudon, James and others},
  journal={Advances in Neural Information Processing Systems},
  volume={35},
  pages={7103--7114},
  year={2022}
}

@article{xlora,
  title={X-LoRA: Mixture of low-rank adapter experts, a flexible framework for large language models with applications in protein mechanics and molecular design},
  author={Buehler, Eric L and Buehler, Markus J},
  journal={APL Machine Learning},
  volume={2},
  number={2},
  year={2024},
  publisher={AIP Publishing}
}

@article{ecot,
  title={Robotic control via embodied chain-of-thought reasoning},
  author={Zawalski, Micha{\l} and Chen, William and Pertsch, Karl and Mees, Oier and Finn, Chelsea and Levine, Sergey},
  journal={arXiv preprint arXiv:2407.08693},
  year={2024}
}

@article{young2023cider,
  title={CIDER: Context sensitive sentiment analysis for short-form text},
  author={Young, James C and Arthur, Rudy and Williams, Hywel TP},
  journal={arXiv preprint arXiv:2307.07864},
  year={2023}
}

@article{chen2024moto,
  title={Moto: Latent motion token as the bridging language for robot manipulation},
  author={Chen, Yi and Ge, Yuying and Li, Yizhuo and Ge, Yixiao and Ding, Mingyu and Shan, Ying and Liu, Xihui},
  journal={arXiv preprint arXiv:2412.04445},
  year={2024}
}

@article{MMstar,
    title={Are We on the Right Way for Evaluating Large Vision-Language Models?},
    author={Chen, Lin and Li, Jinsong and Dong, Xiaoyi and Zhang, Pan and Zang, Yuhang and Chen, Zehui and Duan, Haodong and Wang, Jiaqi and Qiao, Yu and Lin, Dahua and others},
    journal={arXiv preprint arXiv:2403.20330},
    year={2024}
}

@inproceedings{mmmu,
    title={MMMU: A Massive Multi-discipline Multimodal Understanding and Reasoning Benchmark for Expert AGI},
    author={Xiang Yue and Yuansheng Ni and Kai Zhang and Tianyu Zheng and Ruoqi Liu and Ge Zhang and Samuel Stevens and Dongfu Jiang and Weiming Ren and Yuxuan Sun and Cong Wei and Botao Yu and Ruibin Yuan and Renliang Sun and Ming Yin and Boyuan Zheng and Zhenzhu Yang and Yibo Liu and Wenhao Huang and Huan Sun and Yu Su and Wenhu Chen},
    booktitle={Proceedings of the IEEE/CVF Conference on Computer Vision and Pattern Recognition},
    year={2024},
  }

@article{ocrbench,
    title={OCRBench: on the hidden mystery of OCR in large multimodal models},
    volume={67},
    ISSN={1869-1919},
    url={http://dx.doi.org/10.1007/s11432-024-4235-6},
    DOI={10.1007/s11432-024-4235-6},
    number={12},
    journal={Science China Information Sciences},
    publisher={Springer Science and Business Media LLC},
    author={Liu, Yuliang and Li, Zhang and Huang, Mingxin and Yang, Biao and Yu, Wenwen and Li, Chunyuan and Yin, Xu-Cheng and Liu, Cheng-Lin and Jin, Lianwen and Bai, Xiang},
    year={2024},
    month=dec }

@article{pertsch2025fast,
  title={Fast: Efficient action tokenization for vision-language-action models},
  author={Pertsch, Karl and Stachowicz, Kyle and Ichter, Brian and Driess, Danny and Nair, Suraj and Vuong, Quan and Mees, Oier and Finn, Chelsea and Levine, Sergey},
  journal={arXiv preprint arXiv:2501.09747},
  year={2025}
}

@article{qu2025spatialvla,
  title={SpatialVLA: Exploring Spatial Representations for Visual-Language-Action Model},
  author={Qu, Delin and Song, Haoming and Chen, Qizhi and Yao, Yuanqi and Ye, Xinyi and Ding, Yan and Wang, Zhigang and Gu, JiaYuan and Zhao, Bin and Wang, Dong and others},
  journal={arXiv preprint arXiv:2501.15830},
  year={2025}
}

@article{bjorck2025gr00t,
  title={GR00T N1: An Open Foundation Model for Generalist Humanoid Robots},
  author={Bjorck, Johan and Casta{\~n}eda, Fernando and Cherniadev, Nikita and Da, Xingye and Ding, Runyu and Fan, Linxi and Fang, Yu and Fox, Dieter and Hu, Fengyuan and Huang, Spencer and others},
  journal={arXiv preprint arXiv:2503.14734},
  year={2025}
}

@article{lapa,
  title={Latent action pretraining from videos},
  author={Ye, Seonghyeon and Jang, Joel and Jeon, Byeongguk and Joo, Sejune and Yang, Jianwei and Peng, Baolin and Mandlekar, Ajay and Tan, Reuben and Chao, Yu-Wei and Lin, Bill Yuchen and others},
  journal={arXiv preprint arXiv:2410.11758},
  year={2024}
}

@inproceedings{lcb,
  title={From LLMs to Actions: latent codes as bridges in hierarchical robot control},
  author={Shentu, Yide and Wu, Philipp and Rajeswaran, Aravind and Abbeel, Pieter},
  booktitle={2024 IEEE/RSJ International Conference on Intelligent Robots and Systems (IROS)},
  pages={8539--8546},
  year={2024},
  organization={IEEE}
}

@article{sun2024emma,
  title={Emma-X: An Embodied Multimodal Action Model with Grounded Chain of Thought and Look-ahead Spatial Reasoning},
  author={Sun, Qi and Hong, Pengfei and Pala, Tej Deep and Toh, Vernon and Tan, U and Ghosal, Deepanway and Poria, Soujanya and others},
  journal={arXiv preprint arXiv:2412.11974},
  year={2024}
}

@article{zhou2024transfusion,
  title={Transfusion: Predict the next token and diffuse images with one multi-modal model},
  author={Zhou, Chunting and Yu, Lili and Babu, Arun and Tirumala, Kushal and Yasunaga, Michihiro and Shamis, Leonid and Kahn, Jacob and Ma, Xuezhe and Zettlemoyer, Luke and Levy, Omer},
  journal={arXiv preprint arXiv:2408.11039},
  year={2024}
}

@article{pan2025transfer,
  title={Transfer between Modalities with MetaQueries},
  author={Pan, Xichen and Shukla, Satya Narayan and Singh, Aashu and Zhao, Zhuokai and Mishra, Shlok Kumar and Wang, Jialiang and Xu, Zhiyang and Chen, Jiuhai and Li, Kunpeng and Juefei-Xu, Felix and others},
  journal={arXiv preprint arXiv:2504.06256},
  year={2025}
}

@article{llava,
  title={Visual instruction tuning},
  author={Liu, Haotian and Li, Chunyuan and Wu, Qingyang and Lee, Yong Jae},
  journal={Advances in Neural Information Processing Systems},
  volume={36},
  pages={34892--34916},
  year={2023}
}

@article{bunny,
  title={Efficient multimodal learning from data-centric perspective},
  author={He, Muyang and Liu, Yexin and Wu, Boya and Yuan, Jianhao and Wang, Yueze and Huang, Tiejun and Zhao, Bo},
  journal={arXiv preprint arXiv:2402.11530},
  year={2024}
}

@article{openvla-oft,
  title={Fine-tuning vision-language-action models: Optimizing speed and success},
  author={Kim, Moo Jin and Finn, Chelsea and Liang, Percy},
  journal={arXiv preprint arXiv:2502.19645},
  year={2025}
}

@inproceedings{perez2018film,
  title={Film: Visual reasoning with a general conditioning layer},
  author={Perez, Ethan and Strub, Florian and De Vries, Harm and Dumoulin, Vincent and Courville, Aaron},
  booktitle={Proceedings of the AAAI Conference on Artificial Intelligence},
  year={2018}
}

@article{magma,
  title={Magma: A foundation model for multimodal ai agents},
  author={Yang, Jianwei and Tan, Reuben and Wu, Qianhui and Zheng, Ruijie and Peng, Baolin and Liang, Yongyuan and Gu, Yu and Cai, Mu and Ye, Seonghyeon and Jang, Joel and others},
  journal={arXiv preprint arXiv:2502.13130},
  year={2025}
}

@article{hpt,
  title={Scaling proprioceptive-visual learning with heterogeneous pre-trained transformers},
  author={Wang, Lirui and Chen, Xinlei and Zhao, Jialiang and He, Kaiming},
  journal={Advances in Neural Information Processing Systems},
  volume={37},
  pages={124420--124450},
  year={2024}
}

@article{robodual,
  title={Towards Synergistic, Generalized, and Efficient Dual-System for Robotic Manipulation},
  author={Bu, Qingwen and Li, Hongyang and Chen, Li and Cai, Jisong and Zeng, Jia and Cui, Heming and Yao, Maoqing and Qiao, Yu},
  journal={arXiv preprint arXiv:2410.08001},
  year={2024}
}

@inproceedings{DiT,
  title={Scalable diffusion models with transformers},
  author={Peebles, William and Xie, Saining},
  booktitle={Proceedings of the IEEE/CVF International Conference on Computer Vision},
  pages={4195--4205},
  year={2023}
}

@article{cot,
  title={Chain-of-thought prompting elicits reasoning in large language models},
  author={Wei, Jason and Wang, Xuezhi and Schuurmans, Dale and Bosma, Maarten and Xia, Fei and Chi, Ed and Le, Quoc V and Zhou, Denny and others},
  journal={Advances in Neural Information Processing Systems},
  volume={35},
  pages={24824--24837},
  year={2022}
}

@inproceedings{duan2024vlmevalkit,
  title={Vlmevalkit: An open-source toolkit for evaluating large multi-modality models},
  author={Duan, Haodong and Yang, Junming and Qiao, Yuxuan and Fang, Xinyu and Chen, Lin and Liu, Yuan and Dong, Xiaoyi and Zang, Yuhang and Zhang, Pan and Wang, Jiaqi and others},
  booktitle={Proceedings of the 32nd ACM International Conference on Multimedia},
  pages={11198--11201},
  year={2024}
}

@misc{mme,
      title={MME: A Comprehensive Evaluation Benchmark for Multimodal Large Language Models}, 
      author={Chaoyou Fu and Peixian Chen and Yunhang Shen and Yulei Qin and Mengdan Zhang and Xu Lin and Jinrui Yang and Xiawu Zheng and Ke Li and Xing Sun and Yunsheng Wu and Rongrong Ji},
      year={2024},
      eprint={2306.13394},
      archivePrefix={arXiv},
      primaryClass={cs.CV},
      url={https://arxiv.org/abs/2306.13394}, 
}

@inproceedings{guan2024hallusionbench,
  title={Hallusionbench: an advanced diagnostic suite for entangled language hallucination and visual illusion in large vision-language models},
  author={Guan, Tianrui and Liu, Fuxiao and Wu, Xiyang and Xian, Ruiqi and Li, Zongxia and Liu, Xiaoyu and Wang, Xijun and Chen, Lichang and Huang, Furong and Yacoob, Yaser and others},
  booktitle={Proceedings of the IEEE/CVF Conference on Computer Vision and Pattern Recognition},
  pages={14375--14385},
  year={2024}
}

@inproceedings{liu2024mmbench,
  title={Mmbench: Is your multi-modal model an all-around player?},
  author={Liu, Yuan and Duan, Haodong and Zhang, Yuanhan and Li, Bo and Zhang, Songyang and Zhao, Wangbo and Yuan, Yike and Wang, Jiaqi and He, Conghui and Liu, Ziwei and others},
  booktitle={European Conference on Computer Vision},
  pages={216--233},
  year={2024},
  organization={Springer}
}

@inproceedings{singh2019towards,
    title={Towards VQA Models That Can Read},
    author={Singh, Amanpreet and Natarjan, Vivek and Shah, Meet and Jiang, Yu and Chen, Xinlei and Parikh, Devi and Rohrbach, Marcus},
    booktitle={Proceedings of the IEEE/CVF Conference on Computer Vision and Pattern Recognition},
    pages={8317-8326},
    year={2019}
}

@inproceedings{mathew2021docvqa,
  title={Docvqa: A dataset for vqa on document images},
  author={Mathew, Minesh and Karatzas, Dimosthenis and Jawahar, CV},
  booktitle={Proceedings of the IEEE/CVF Winter Conference on Applications of Computer Vision},
  pages={2200--2209},
  year={2021}
}

@inproceedings{mathew2022infographicvqa,
  title={Infographicvqa},
  author={Mathew, Minesh and Bagal, Viraj and Tito, Rub{\`e}n and Karatzas, Dimosthenis and Valveny, Ernest and Jawahar, CV},
  booktitle={Proceedings of the IEEE/CVF Winter Conference on Applications of Computer Vision},
  pages={1697--1706},
  year={2022}
}

@article{qwen2vl,
  title={Qwen2-vl: Enhancing vision-language model's perception of the world at any resolution},
  author={Wang, Peng and Bai, Shuai and Tan, Sinan and Wang, Shijie and Fan, Zhihao and Bai, Jinze and Chen, Keqin and Liu, Xuejing and Wang, Jialin and Ge, Wenbin and others},
  journal={arXiv preprint arXiv:2409.12191},
  year={2024}
}

@inproceedings{ai2d,
title={A diagram is worth a dozen images},
author={Kembhavi, Aniruddha and Salvato, Mike and Kolve, Eric and Seo, Minjoon and Hajishirzi, Hannaneh and Farhadi, Ali},
booktitle={European Conference on Computer Vision},
pages={235--251},
year={2016},
organization={Springer}
}

@article{masry2022chartqa,
  title={Chartqa: A benchmark for question answering about charts with visual and logical reasoning},
  author={Masry, Ahmed and Long, Do Xuan and Tan, Jia Qing and Joty, Shafiq and Hoque, Enamul},
  journal={arXiv preprint arXiv:2203.10244},
  year={2022}
}

@misc{rwqa,
  title={RealWorldQA},
  author={RealWorld Team},
  year={2024},
  url={https://x.ai/news/grok-1.5v}
}

@misc{helix,
  title={Helix},
  author={Figure AI},
  year={2024},
  url={https://www.figure.ai/news/helix}
}

@inproceedings{karamcheti2024prismatic,
  title={Prismatic vlms: Investigating the design space of visually-conditioned language models},
  author={Karamcheti, Siddharth and Nair, Suraj and Balakrishna, Ashwin and Liang, Percy and Kollar, Thomas and Sadigh, Dorsa},
  booktitle={Forty-first International Conference on Machine Learning},
  year={2024}
}

@article{beyer2024paligemma,
  title={Paligemma: A versatile 3b vlm for transfer},
  author={Beyer, Lucas and Steiner, Andreas and Pinto, Andr{\'e} Susano and Kolesnikov, Alexander and Wang, Xiao and Salz, Daniel and Neumann, Maxim and Alabdulmohsin, Ibrahim and Tschannen, Michael and Bugliarello, Emanuele and others},
  journal={arXiv preprint arXiv:2407.07726},
  year={2024}
}

@article{zhou2025chatvla,
  title={Chatvla: Unified multimodal understanding and robot control with vision-language-action model},
  author={Zhou, Zhongyi and Zhu, Yichen and Zhu, Minjie and Wen, Junjie and Liu, Ning and Xu, Zhiyuan and Meng, Weibin and Cheng, Ran and Peng, Yaxin and Shen, Chaomin and others},
  journal={arXiv preprint arXiv:2502.14420},
  year={2025}
}

@article{niu2024llarva,
  title={Llarva: Vision-action instruction tuning enhances robot learning},
  author={Niu, Dantong and Sharma, Yuvan and Biamby, Giscard and Quenum, Jerome and Bai, Yutong and Shi, Baifeng and Darrell, Trevor and Herzig, Roei},
  journal={arXiv preprint arXiv:2406.11815},
  year={2024}
}

@article{tian2024predictive,
  title={Predictive inverse dynamics models are scalable learners for robotic manipulation},
  author={Tian, Yang and Yang, Sizhe and Zeng, Jia and Wang, Ping and Lin, Dahua and Dong, Hao and Pang, Jiangmiao},
  journal={arXiv preprint arXiv:2412.15109},
  year={2024}
}

@article{french1999catastrophic,
  title={Catastrophic forgetting in connectionist networks},
  author={French, Robert M},
  journal={Trends in Cognitive Sciences},
  volume={3},
  number={4},
  pages={128--135},
  year={1999},
  publisher={Elsevier}
}

@misc{zheng2025universalactionsenhancedembodied,
      title={Universal Actions for Enhanced Embodied Foundation Models}, 
      author={Jinliang Zheng and Jianxiong Li and Dongxiu Liu and Yinan Zheng and Zhihao Wang and Zhonghong Ou and Yu Liu and Jingjing Liu and Ya-Qin Zhang and Xianyuan Zhan},
      year={2025},
      eprint={2501.10105},
      archivePrefix={arXiv},
      primaryClass={cs.RO},
      url={https://arxiv.org/abs/2501.10105}, 
}

@article{wang2024poco,
  title={Poco: Policy composition from and for heterogeneous robot learning},
  author={Wang, Lirui and Zhao, Jialiang and Du, Yilun and Adelson, Edward H and Tedrake, Russ},
  journal={arXiv preprint arXiv:2402.02511},
  year={2024}
}

@inproceedings{bleu,
  title={Bleu: a method for automatic evaluation of machine translation},
  author={Papineni, Kishore and Roukos, Salim and Ward, Todd and Zhu, Wei-Jing},
  booktitle={Proceedings of the 40th annual meeting of the Association for Computational Linguistics},
  pages={311--318},
  year={2002}
}

@inproceedings{meteor,
  title={METEOR: An automatic metric for MT evaluation with improved correlation with human judgments},
  author={Banerjee, Satanjeev and Lavie, Alon},
  booktitle={Proceedings of the {ACL} Workshop on Intrinsic and Extrinsic Evaluation Measures for Machine Translation and/or Summarization},
  pages={65--72},
  year={2005}
}

@article{Sentence-BERT,
  title={Sentence-BERT: Sentence Embeddings using Siamese BERT-Networks},
  author={Reimers, N},
  journal={arXiv preprint arXiv:1908.10084},
  year={2019}
}

@article{simcse,
  title={Simcse: Simple contrastive learning of sentence embeddings},
  author={Gao, Tianyu and Yao, Xingcheng and Chen, Danqi},
  journal={arXiv preprint arXiv:2104.08821},
  year={2021}
}

@article{li2025hamster,
  title={Hamster: Hierarchical action models for open-world robot manipulation},
  author={Li, Yi and Deng, Yuquan and Zhang, Jesse and Jang, Joel and Memmel, Marius and Yu, Raymond and Garrett, Caelan Reed and Ramos, Fabio and Fox, Dieter and Li, Anqi and others},
  journal={arXiv preprint arXiv:2502.05485},
  year={2025}
}

@article{deng2025graspvla,
  title={Graspvla: a grasping foundation model pre-trained on billion-scale synthetic action data},
  author={Deng, Shengliang and Yan, Mi and Wei, Songlin and Ma, Haixin and Yang, Yuxin and Chen, Jiayi and Zhang, Zhiqi and Yang, Taoyu and Zhang, Xuheng and Cui, Heming and others},
  journal={arXiv preprint arXiv:2505.03233},
  year={2025}
}

@inproceedings{gao2025genmanip,
  title={GENMANIP: LLM-driven Simulation for Generalizable Instruction-Following Manipulation},
  author={Gao, Ning and Chen, Yilun and Yang, Shuai and Chen, Xinyi and Tian, Yang and Li, Hao and Huang, Haifeng and Wang, Hanqing and Wang, Tai and Pang, Jiangmiao},
  booktitle={Proceedings of the IEEE/CVF Conference on Computer Vision and Pattern Recognition},
  pages={12187--12198},
  year={2025}
}

@article{niu2025pre,
  title={Pre-training auto-regressive robotic models with 4d representations},
  author={Niu, Dantong and Sharma, Yuvan and Xue, Haoru and Biamby, Giscard and Zhang, Junyi and Ji, Ziteng and Darrell, Trevor and Herzig, Roei},
  journal={arXiv preprint arXiv:2502.13142},
  year={2025}
}

@article{driess2025knowledge,
  title={Knowledge insulating vision-language-action models: Train fast, run fast, generalize better},
  author={Driess, Danny and Springenberg, Jost Tobias and Ichter, Brian and Yu, Lili and Li-Bell, Adrian and Pertsch, Karl and Ren, Allen Z and Walke, Homer and Vuong, Quan and Shi, Lucy Xiaoyang and others},
  journal={arXiv preprint arXiv:2505.23705},
  year={2025}
}

@article{deng2025emerging,
  title={Emerging properties in unified multimodal pretraining},
  author={Deng, Chaorui and Zhu, Deyao and Li, Kunchang and Gou, Chenhui and Li, Feng and Wang, Zeyu and Zhong, Shu and Yu, Weihao and Nie, Xiaonan and Song, Ziang and others},
  journal={arXiv preprint arXiv:2505.14683},
  year={2025}
}

@article{mu2025comprehensive,
  title={A comprehensive survey of mixture-of-experts: Algorithms, theory, and applications},
  author={Mu, Siyuan and Lin, Sen},
  journal={arXiv preprint arXiv:2503.07137},
  year={2025}
}

@inproceedings{yao2023react,
  title={React: Synergizing reasoning and acting in language models},
  author={Yao, Shunyu and Zhao, Jeffrey and Yu, Dian and Du, Nan and Shafran, Izhak and Narasimhan, Karthik and Cao, Yuan},
  booktitle={International Conference on Learning Representations},
  year={2023}
}

@article{team2025longcat,
  title={LongCat-Flash Technical Report},
  author={Team, Meituan LongCat and Li, Bei and Lei, Bingye and Wang, Bo and Rong, Bolin and Wang, Chao and Zhang, Chao and Gao, Chen andf Zhang, Chen and Sun, Cheng and others},
  journal={arXiv preprint arXiv:2509.01322},
  year={2025}
}

@article{liu2024deepseek,
  title={Deepseek-v3 technical report},
  author={Liu, Aixin and Feng, Bei and Xue, Bing and Wang, Bingxuan and Wu, Bochao and Lu, Chengda and Zhao, Chenggang and Deng, Chengqi and Zhang, Chenyu and Ruan, Chong and others},
  journal={arXiv preprint arXiv:2412.19437},
  year={2024}
}

@article{li2024llara,
  title={Llara: Supercharging robot learning data for vision-language policy},
  author={Li, Xiang and Mata, Cristina and Park, Jongwoo and Kahatapitiya, Kumara and Jang, Yoo Sung and Shang, Jinghuan and Ranasinghe, Kanchana and Burgert, Ryan and Cai, Mu and Lee, Yong Jae and others},
  journal={arXiv preprint arXiv:2406.20095},
  year={2024}
}

@inproceedings{zhao2025cot,
  title={Cot-vla: Visual chain-of-thought reasoning for vision-language-action models},
  author={Zhao, Qingqing and Lu, Yao and Kim, Moo Jin and Fu, Zipeng and Zhang, Zhuoyang and Wu, Yecheng and Li, Zhaoshuo and Ma, Qianli and Han, Song and Finn, Chelsea and others},
  booktitle={Proceedings of the IEEE/CVF Conference on Computer Vision and Pattern Recognition},
  pages={1702--1713},
  year={2025}
}

@article{lingbot_vla,
  title={A Pragmatic VLA Foundation Model},
  author={Wu, Wei and Lu, Fan and Wang, Yunnan and Yang, Shuai and Liu, Shi and Wang, Fangjing and Zhu, Qian and Sun, He and Wang, Yong and Ma, Shuailei and others},
  journal={arXiv preprint arXiv:2601.18692},
  year={2026}
}
\bibliographystyle{iclr2026_conference}

\clearpage
\newpage
\appendix
\vspace{2em}
\begin{center}
  {\huge \bfseries Appendix}
\end{center}
\vspace{1em}

\let\addcontentsline\oldaddcontentsline
\tableofcontents
% \clearpage
% \newpage
\vspace{2cm}
The supplementary material is organized as follows:
\begin{itemize}[leftmargin=0.2in]
    % \item We provide a \textbf{Video Demo} to demonstrate the rollout episodes of our model.
    \item \Cref{sec: Extra experiment and analysis} presents: (1) extended analysis of InstructVLA, (2) additional benchmarks on embodied understanding, (3) extra simulation benchmark and ablation study, (4) finetuning of OpenVLA under the same settings as InstructVLA, and (5) extra real-world ablation study.
    \item \Cref{sec: extra related works} discusses related concepts to InstructVLA and the proposed vision-language-action instruction tuning methods.
    \item \Cref{sec: failure cases analysis} provides additional case analysis for InstructVLA, OpenVLA, and GPT-4o System2.
    \item \Cref{sec: Data Annotation Details and Analysis} lists data annotation details, including GPT-4o prompt and dataset statistics. We further analyse the distribution of the instructions from two dimensions: task diversity and language diversity.
    \item \Cref{sec: Benchmark Task Descriptions and Visualization} visualizes the SimplerEnv-Instruct benchmark and the acknowledgements of 3D assets.
    \item \Cref{sec: Model Design and Training Details} details the model architecture, training configurations, inference speeds under different settings, and compute resources used.
    \item \Cref{sec: Multimodal Examples} shows several multimodal question answering examples.
    \item \Cref{sec: Real-world Experiments Setup} describes the real-world experimental setup and provides example executions.
    \item \Cref{sec: Broader Impacts} discusses the broader impacts, limitations, and outlines future directions for InstructVLA.
\end{itemize}

\newpage

\section{More Experiments and Analysis}
\label{sec: Extra experiment and analysis}

\subsection{Further Discussions}

Our further analysis is threefold. First, we present visualizations and scaling curves to examine the MoE and latent action designs. Second, we provide a detailed analysis of reasoning gains in manipulation tasks and case studies. Finally, we demonstrate that InstructVLA supports zero-shot dual-frequency generation to accelerate inference and compare the dataset scales used across different studies.

\label{sec: sub dis}
\subsubsection{Extra Model Design Analysis}

The MoE and latent action are our key design components. We present an example illustrating the role of MoE under different task settings, including simple and reasoning instructions, with and without model reasoning. For latent action, we analyze its scaling behavior to guide future tuning.

\begin{figure}[h]
    \centering
    \includegraphics[width=1\linewidth]{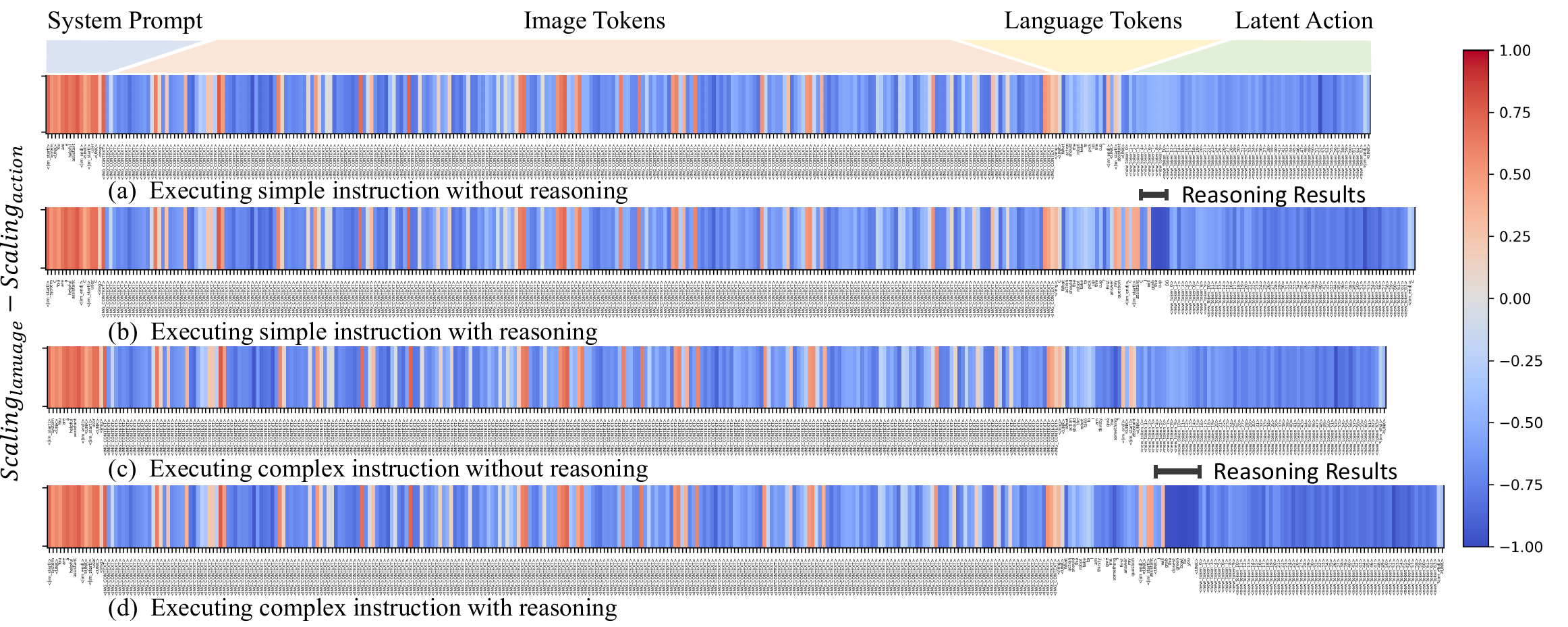}
    \caption{\textbf{Activation visualization.} We evaluate a WidowX zero-shot example across four settings. Red indicates stronger activation in the language adapter, while blue indicates stronger activation in the action adapter. The horizontal axis lists each language token. The generated tokens are marked.}
    \label{fig:supp_moe}
\end{figure}

\noindent\textbf{Analysis of MoE gating.} From the example in~\Cref{fig:supp_moe}, we draw the following intuitive conclusions:  

\begin{itemize}[leftmargin=2em]
    \item System prompts are primarily processed by the language adapter, reflecting its close connection to pretraining. 
    \item Visual information is processed by both the language and action adapters, indicating that both semantic understanding and manipulation decision-making require visual inputs. 
    \item During language generation, the model engages not only in multimodal reasoning but also in manipulation planning, as evidenced by the activation of the action expert. Notably, the action expert attends more strongly to nouns and verbs in the generated tokens, highlighting its role in instruction following.  
    \item During latent action generation, the language expert plays a less prominent role. Instead, with multimodal reasoning, the model concentrates more effectively on action generation, as shown by the stronger activation of the action expert (deeper blue).  
\end{itemize}

To conclude, the MoE has demonstrated its effectiveness in improving efficiency and handling heterogeneous datasets~\citep{mu2025comprehensive,xlora,moe,team2025longcat,liu2024deepseek}. In InstructVLA, we further investigate how the MoE facilitates interleaved multimodal reasoning and manipulation decision making.

\begin{figure}[ht]
    \centering
    \includegraphics[width=0.4\linewidth]{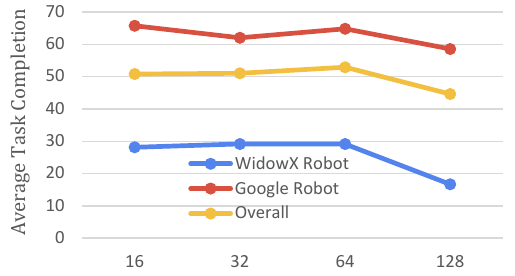}
    \caption{Impact of latent action token quantity on robot performance.}
    \label{fig:supp_latent_action}
\end{figure}

\noindent\textbf{Effects of latent action.} Latent action tokens are a key design component for decoupling high-level VLM planning from low-level action generation. As shown in~\Cref{fig:supp_latent_action}, we vary the number of tokens from 16 to 128. Too few tokens limit behavioral diversity, while too many reduce training efficiency. A setting of 64 offers a good trade-off under our current configuration.

\subsubsection{{Extra Reasoning-Manipulation Analysis}}

In this section, we discuss the efficiency and design choices of VLA-IT training. We then analyze how multimodal reasoning benefits manipulation through fine-grained evaluation, examine its role in cross-embodiment generalization, and present a case study illustrating how a unique multimodal capability addresses challenging tasks.

\begin{figure}
    \centering
    \includegraphics[width=1\linewidth]{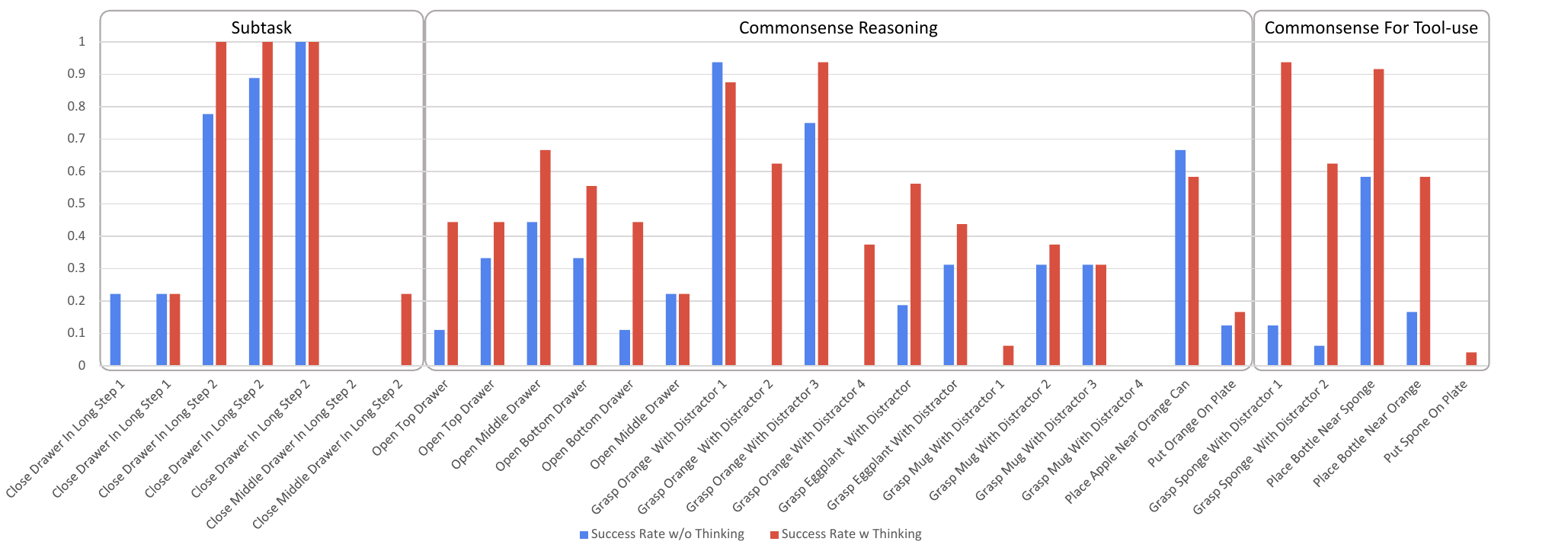}
    \caption{\textbf{Performance visualization} of 30 situated reasoning tasks with and without reasoning enabled. Activating reasoning in our generalist model generally improves performance. For clarity, tasks are grouped into three categories: \textit{Subtask}, involving subtask identification; \textit{Commonsense Reasoning}, requiring broad world knowledge; and \textit{Commonsense for Tool Use}, focusing on tool-related reasoning.}
    \label{fig:supp_thinking}
\end{figure}

\noindent\textbf{Effect of VLA-IT on Scaling and Reasoning.}
As shown in~\Cref{tab: manip}, although the InstructVLA-expert model does not outperform the OpenVLA(OXE) on Situated Reasoning of SimplerEnv-Instruct, which benefits from direct full fine-tuning of the VLM backbone, InstructVLA-expert shows promising scaling ability in understanding complex instructions and performing test-time thinking after stage-2 VLA-IT training. This result reflects a deliberate design choice in InstructVLA, where latent action learning during pretraining focuses on querying from visual and simple instruction features rather than relying on the full semantic space of the VLM too early. This design offers two significant advantages. First, it preserves the original semantic space of the pretrained VLM, maintaining its vision-language capabilities. Second, it enables the model to integrate diverse reasoning contexts during VLA-IT training. These properties contribute to the strong performance gains achieved by our generalist model and demonstrate the effectiveness of this training paradigm.

\begin{figure}[h]
    \centering
    \includegraphics[width=1\linewidth]{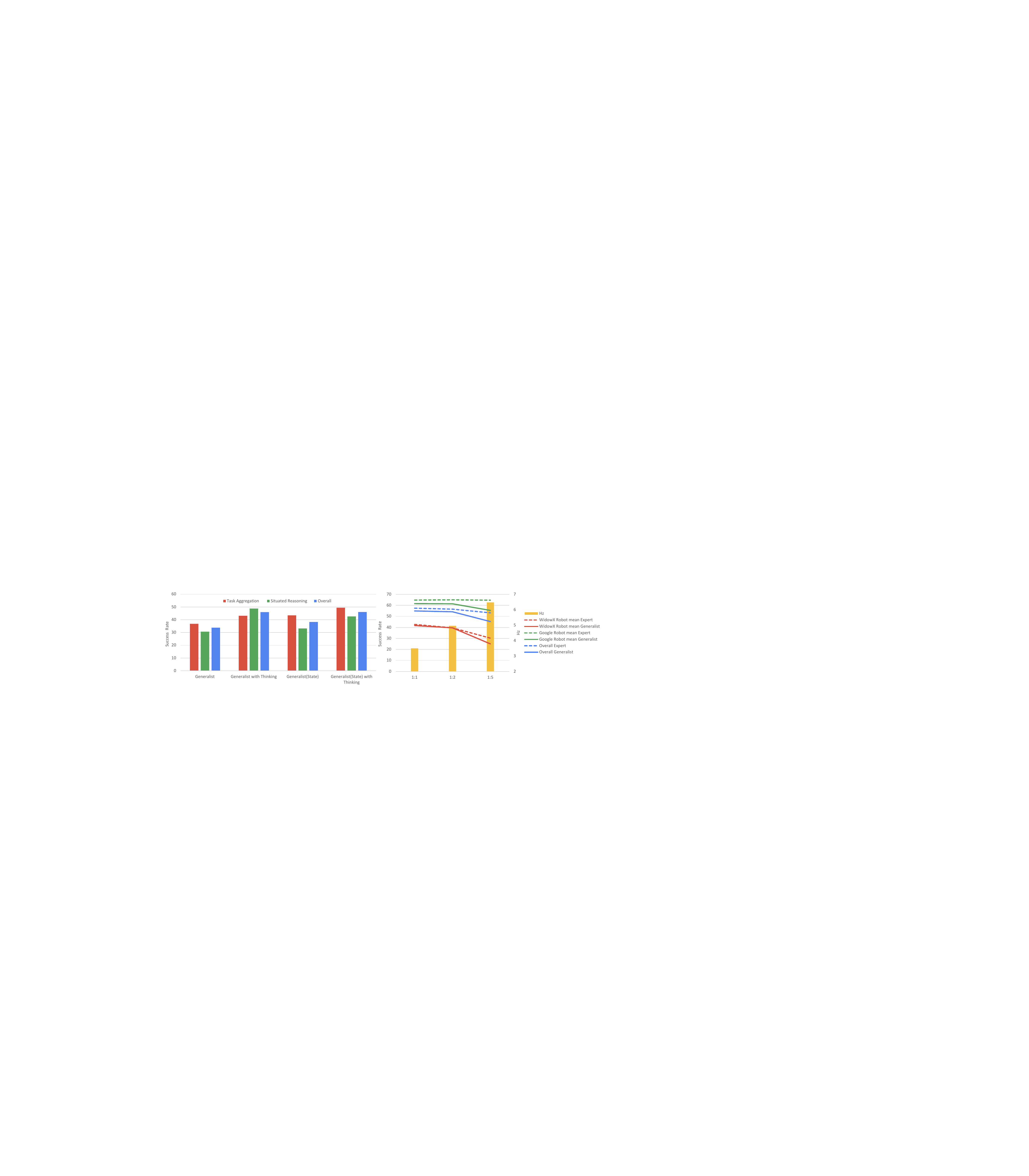}
    \caption{\textbf{Test-time tinking and dual-frequency evaluation.} ``Expert'' refers to the model after action pretraining, while ``Generalist'' denotes the model after VLA-IT tuning. For dual-frequency evaluation, the horizontal axis represents the ratio of VLM executions to expert model executions.}
    \label{fig:dual}
\end{figure}

\noindent\textbf{Embodied reasoning helps manipulation.} Allowing the model to perform test-time thinking by generating textual analysis of the given instruction can improve performance, particularly on situated reasoning tasks, as shown in~\Cref{fig:dual} (left). Notably, while the model with access to robot state outperforms the one without state when no instruction response is required, it provides limited performance gains when instruction following is involved. We hypothesize that state information helps the model retain manipulation skills but compromises its generalization to OOD environments and instructions.

\textbf{Fine-grained analysis of reasoning gains in manipulation tasks.} We compare the performance of the generalist model on SimplerEnv-Instruct with and without vision language reasoning, as shown in~\Cref{fig:supp_thinking}. A clear performance gap emerges in tasks involving commonsense tool use and interaction with articulated objects. This may result from instructions that do not explicitly state the intended actions and objects. For example, retrieving a cleaning tool from a drawer requires the robot to infer whether the prerequisite of an open drawer is satisfied, and to identify the sponge as the appropriate tool among several options. In addition to these cases, the reasoning process also improves performance on other situated reasoning tasks by grounding unfamiliar instructions using the pretrained in-domain knowledge of the vision language model.

\begin{figure}[h]
    \centering
    \includegraphics[width=1\linewidth]{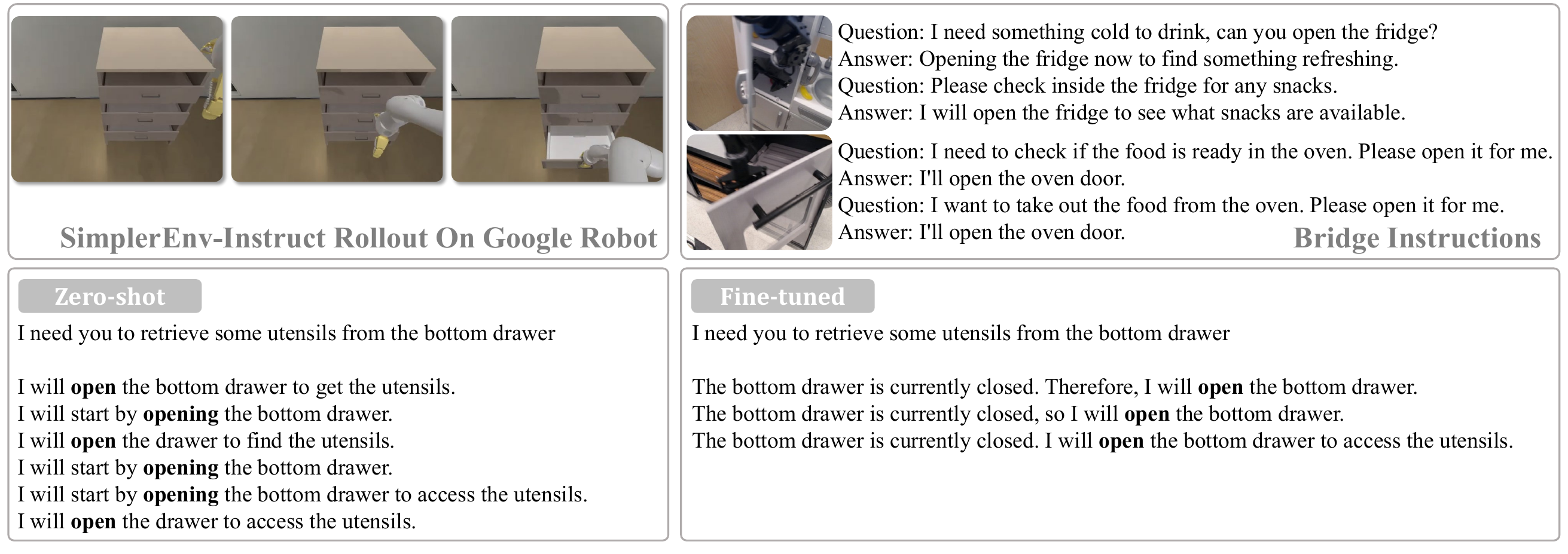}
    \caption{\textbf{Case study on cross-embodiment.} Top left: rollouts on SimplerEnv-Instruct. Top right: similar scenarios from the Bridge dataset with corresponding instructions. Bottom left: zero-shot results trained only on Bridge instructions. Bottom right: rollouts from the fine-tuned model.}
    \label{fig:cross_embodiment}
\end{figure}

\noindent\textbf{VLA instruction tuning for cross-embodiment understanding.} To assess whether InstructVLA retains this capability, we evaluate three variants on SimplerEnv-Instruct (see~\Cref{tab: cross embodiment}): InstructVLA-Expert, trained solely on atomic instructions without test-time thinking; InstructVLA Generalist (Bridge), trained with the VLA-IT dataset on Bridge and the original Fractal dataset; and InstructVLA Generalist, trained with the full VLA-IT datasets across both environments. Adding the Bridge dataset results in a 139.4\% improvement in Situated Reasoning performance for Generalist (Bridge) over the expert baseline, while task aggregation performance remains comparable. This discrepancy reflects differing generalization requirements: task aggregation emphasizes linguistic robustness, whereas Situated Reasoning demands vision-language grounding prior to action. The latter particularly benefits from the preserved reasoning capabilities of the pretrained VLM. As illustrated in~\Cref{fig:cross_embodiment}, the zero-shot model generates more diverse and accurate outputs than its fine-tuned counterpart.

\begin{table}[ht]
\centering
\caption{\textbf{Instruction tuning data ablation.} We evaluate three settings: without VLA-IT data, with data only on Bridge, and with VLA-IT data on both Fractal and Bridge. This ablation examines the contribution of the VLA-IT dataset and the cross-embodiment generalization of InstructVLA on SimplerEnv-Instruct.}
\resizebox{\textwidth}{!}{%
    \begin{tabular}{cccccc}
    \toprule
    \multicolumn{2}{c}{Instruction Tuning Data}     & \multirow{2}{*}{Name} & \multirow{2}{*}{Task Aggregation} & \multirow{2}{*}{Situated Reasoning} & \multirow{2}{*}{Overall} \\
    Bridge & Fractal & & & \\ \midrule
    \ding{55} & \ding{55} & Expert                & 20.8    & 10.4    & 15.6   \\
    \ding{51} & \ding{55} & Generalist (Bridge)   & 18.4    & 24.9    & 21.7   \\
    \ding{51} & \ding{51} & Generalist            & 43.3    & 48.8    & 46.0   \\ \bottomrule
    \end{tabular}
    \label{tab: cross embodiment}
}
\end{table}

\begin{figure}
    \centering
    \includegraphics[width=1\linewidth]{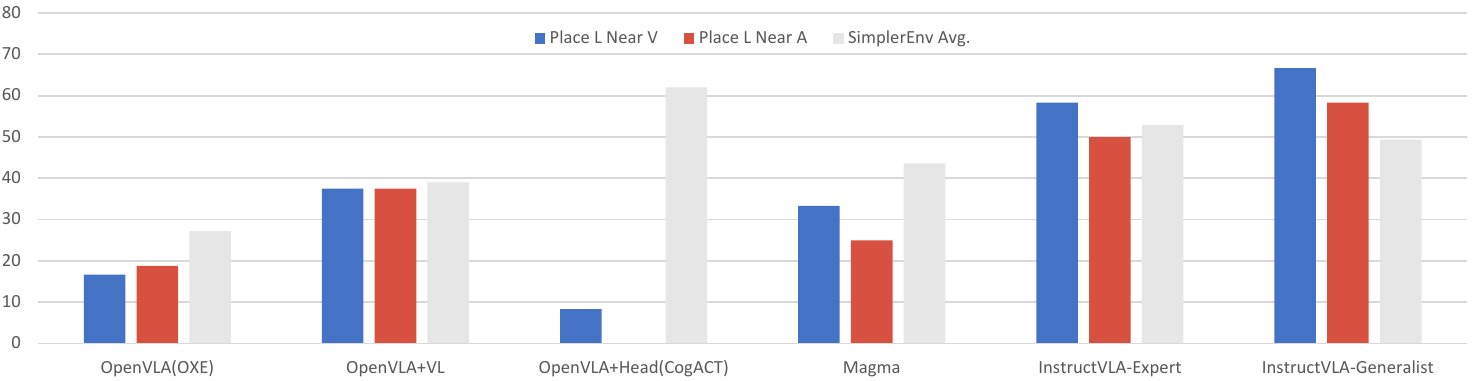}
    \caption{\textbf{Case study on multimodal capabilities.} OCR represents a unique multimodal skill of VLMs that is absent from typical manipulation datasets. We evaluate two tasks from the task aggregation set in SimplerEnv-Instruct, involving moving one letter to another (see~\Cref{fig:openvla}(1)). By comparing different finetuning paradigms, we assess how effectively multimodal capabilities are integrated into VLA models.}

    \label{fig:ocr}
\end{figure}

\noindent\textbf{Case study on multimodal capability transfer.} As shown in~\Cref{fig:ocr}, we compare InstructVLA with OpenVLA~\citep{openvla}, Magma~\citep{magma}, and CogACT~\citep{cogact}, all using the same input (language instruction and a single image). \textbf{InstructVLA-Expert, though trained without multimodal and VLA-IT datasets, retains the OCR capability of the underlying VLM and achieves the best performance among baselines trained solely on manipulation data.} Finetuning InstructVLA-Expert into InstructVLA-Generalist with multimodal and VLA-IT datasets further enhances performance. For autoregressive models such as OpenVLA and Magma, multimodal finetuning improves OCR ability. In contrast, CogACT, when fine-tuned from OpenVLA(OXE) only on manipulation data with an action head, shows improved in-domain performance (on SimplerEnv) but suffers in generalization.

\subsubsection{Extra Inference and Training analysis}

\noindent\textbf{Dual frequency inference.} To further analyze the relationship between latent actions generated by the VLM and the final decoded actions, we decouple the inference frequencies of the VLM and the action expert, as illustrated in~\Cref{fig:dual} right. The results show that performance remains stable at a 1:2 ratio (VLM:expert), but begins to degrade at higher ratios. This suggests that latent actions offer relatively stable guidance to the action expert, reducing the need for frequent VLM queries.

\noindent\textbf{Training at scale.} A generalist VLA model with vision-language capabilities should be scalable across both manipulation and multimodal datasets. In this context, we compare datasets used by models claiming generalist abilities, as shown in~\Cref{tab: data comparison}. RoboMamba~\citep{liu2024robomamba} utilizes a limited manipulation dataset compared to other methods, while the dataset for ChatVLA~\citep{zhou2025chatvla} is not reported. $\pi_{0.5}$~\citep{pi05} employs a significantly larger multimodal dataset than other approaches, though its multimodal performance is not disclosed. Magma uses more robot and multimodal data but achieves slightly worse performance on both multimodal and manipulation benchmarks compared to InstructVLA.

\begin{table}[h]
\centering
\caption{\textbf{Data comparison of different methods.} ``Trans.'' denotes transitions.}
\label{tab: data comparison}
\resizebox{\textwidth}{!}{%
\begin{tabular}{lcccccc}
\toprule
\multicolumn{1}{c}{} & Magma\citep{magma}  & ChatVLA\citep{zhou2025chatvla}         & RoboMamba\citep{liu2024robomamba}      & $\pi_{0.5}$\citep{pi05}   & InstructVLA \\
\midrule
Manipulation Data    & 9.4M Trans. & -  & 10K Trans. & >10000 Hours   & 469 Hours/~5.9M Trans.  \\
Multimodal Data     & 1.2M Images + 4M Videos & 54K      & 1.5M            & \textgreater{}7M  & 2M          \\
\bottomrule
\end{tabular}
}
\end{table}

\clearpage
\newpage

\subsection{Embodied Understanding Evaluation}

\label{sec: vl evaluation}
\begin{table}[h]
\centering
\caption{\textbf{VLA-IT captioning evaluation.} ``Sentence-BERT'' and ``SimCSE'' represent learning-based evaluation methods, while the remaining metrics are traditional n-gram-based evaluations focused on word distribution.}
\label{tab: Captioning}

\resizebox{\textwidth}{!}{
    \begin{tabular}{llcccccc}%ccccccc
    \toprule
    Methods                 & \# Params &  Sentence-BERT & SimCSE & BLEU-1 & BLEU-4 & METEOR & CIDER \\
    \midrule
    Qwen2-VL~\citep{qwen2vl} & 1.5B & 61.3 & 67.5 & 16.8 & 1.5 & 12.4 & 0.30 \\
    GPT4o~\citep{GPT-4}      &  - & 60.7 & 67.1 & 16.3 & 1.8 & 16.2 & 0.09 \\
    \midrule
    OpenVLA(VLA-IT)~\citep{openvla}& 7B & 0.0 & 0.0 & 0.0 & 0.0 & 0.0 & 0.00 \\
    Magma~\citep{magma}& 8B & 59.8 & 66.7 & 12.4 & 1.2 & 12.3 & 0.12 \\
    \midrule
    InstructVLA(Generalist) & 1.5B & 72.0 & 77.0 & 44.3 & 8.2 & 18.7 & 0.84 \\
    \bottomrule
    \end{tabular}
}
\vspace{4pt}
\caption{\textbf{VLA-IT question-answering evaluation.}}
\label{tab: qa}

\resizebox{\textwidth}{!}{
    \begin{tabular}{llccccccc}%ccccccc
    \toprule
    Methods                 & \# Params & Sentence-BERT & SimCSE & BLEU-1 & BLEU-4 & METEOR & CIDER \\
    \midrule
    Qwen2-VL~\citep{qwen2vl} & 1.5B & 51.9 & 53.4 & 15.3 & 2.8 & 17.9 & 0.82 \\
    GPT4o~\citep{GPT-4}      & - & 63.6 & 63.6 & 29.6 & 19.9 & 9.8 & 1.16 \\
    \midrule
    OpenVLA(VLA-IT)~\citep{openvla}& 7B & 0.0 & 0.0 & 0.0 & 0.0 & 0.0 & 0.00 \\
    Magma~\citep{magma}& 8B & 53.5 & 54.5 & 23.7 & 5.7 & 21.6 & 1.04 \\
    \midrule
    InstructVLA(Generalist) & 1.5B & 64.9 & 65.9 & 44.6 & 17.4 & 23.5 & 1.85 \\
    \bottomrule
    \end{tabular}
}
\vspace{4pt}

\caption{\textbf{VLA-IT instruction response evaluation.} We use ``context creation'' annotations, as they present a more challenging and diverse set of instructions.}
\label{tab: instruction}

\resizebox{\textwidth}{!}{
    \begin{tabular}{llccccccc}
    \toprule
    Methods                 & \# Params & Sentence-BERT & SimCSE & BLEU-1 & BLEU-4 & METEOR & CIDER \\
    \midrule
    Qwen2-VL~\citep{qwen2vl} & 1.5B & 52.3 & 54.0 & 5.6 & 1.5 & 11.6 & 0.09 \\
    GPT4o~\citep{GPT-4}      & - & 52.8 & 54.1 & 17.8 & 4.2 & 20.6 & 1.02 \\
    \midrule
    OpenVLA(VLA-IT)~\citep{openvla}& 7B & 0.0 & 0.0 & 0.0 & 0.0 & 0.0 & 0.00 \\
    Magma~\citep{magma}& 8B & 10.9 & 13.6 & 3.7 & 0.8 & 1.6 & 0.00 \\
    \midrule
    InstructVLA(Generalist) & 1.5B & 71.6 & 73.1 & 50.2 & 24.1 & 25.8 & 2.26 \\
    \bottomrule
    \end{tabular}
}
\end{table}

In addition to the multimodal and closed-loop evaluations presented in the main results, we conduct supplementary language evaluations on the proposed VLA-IT dataset. This evaluation uses manually verified VLA-IT annotations on the Bridge dataset~\citep{Bridge_data}, chosen for its diversity and distinct validation split. We generate 1,000 annotations following the method described in the VLA-IT dataset generation section. Two evaluation metrics are employed: (1) learning-based methods~\citep{Sentence-BERT,simcse}, and (2) traditional metrics~\citep{bleu,young2023cider,meteor}.

The captioning, question-answering and instruction-following results are presented in~\Cref{tab: Captioning,tab: qa,tab: instruction}. We select Qwen2-VL~\citep{qwen2vl} and GPT-4o~\citep{GPT-4} as zero-shot VLM baselines, and include Magma~\citep{magma} (zero-shot) and OpenVLA~\citep{openvla} fine-tuned on the VLA-IT dataset as baselines for VLA models.

\begin{figure*}[t]
    \centering
    \includegraphics[width=0.7\linewidth]{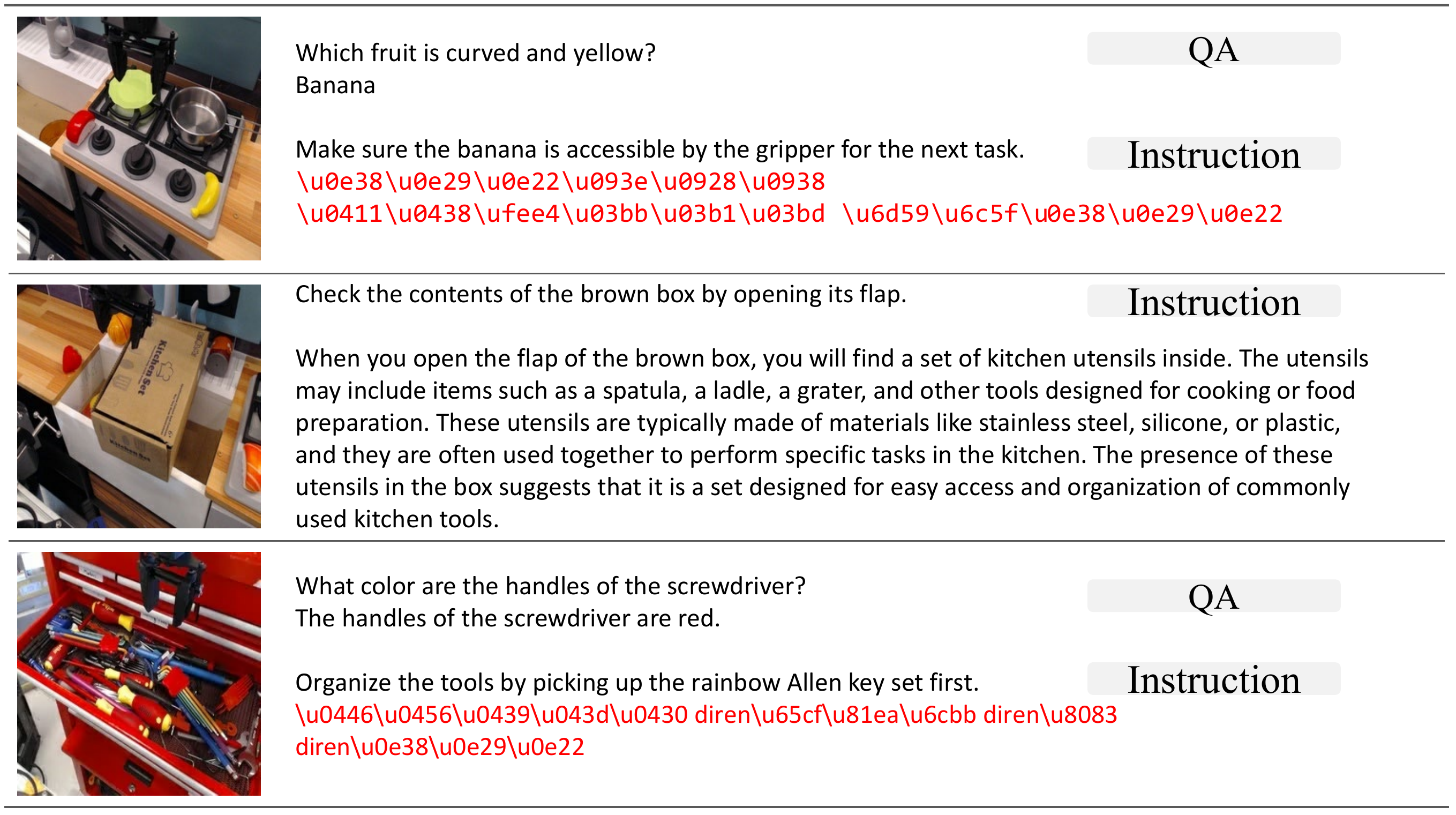}
    \caption{\textbf{Magma results.} Magma's responses collapse when given instructions resembling those in its manipulation tasks, possibly due to learned actions interfering with its language latent space.}
    \label{fig:magma}
\end{figure*}

Although OpenVLA is fine-tuned on the VLA-IT dataset, it fails to generate complete sentences under the same evaluation setting as InstructVLA, despite the performance on multiple-choice benchmarks reported in our main results. This suggests a significant loss of its free-form dialogue capability. Magma performs well on question answering and captioning tasks. However, it struggles with instruction response (\Cref{fig:magma}), often generating outputs misaligned with the given image. \textit{We hypothesize that this failure stems from the similarity between these instructions and the atomic commands used in finetuning manipulation datasets, which disrupts the coherence of the language latent space near the action latent space.} This suggests a limited capacity to interpret and generalize free-form instructions, hindering effective transfer of vision-language capabilities.

\begin{figure*}[t]
    \centering
    \includegraphics[width=1\linewidth]{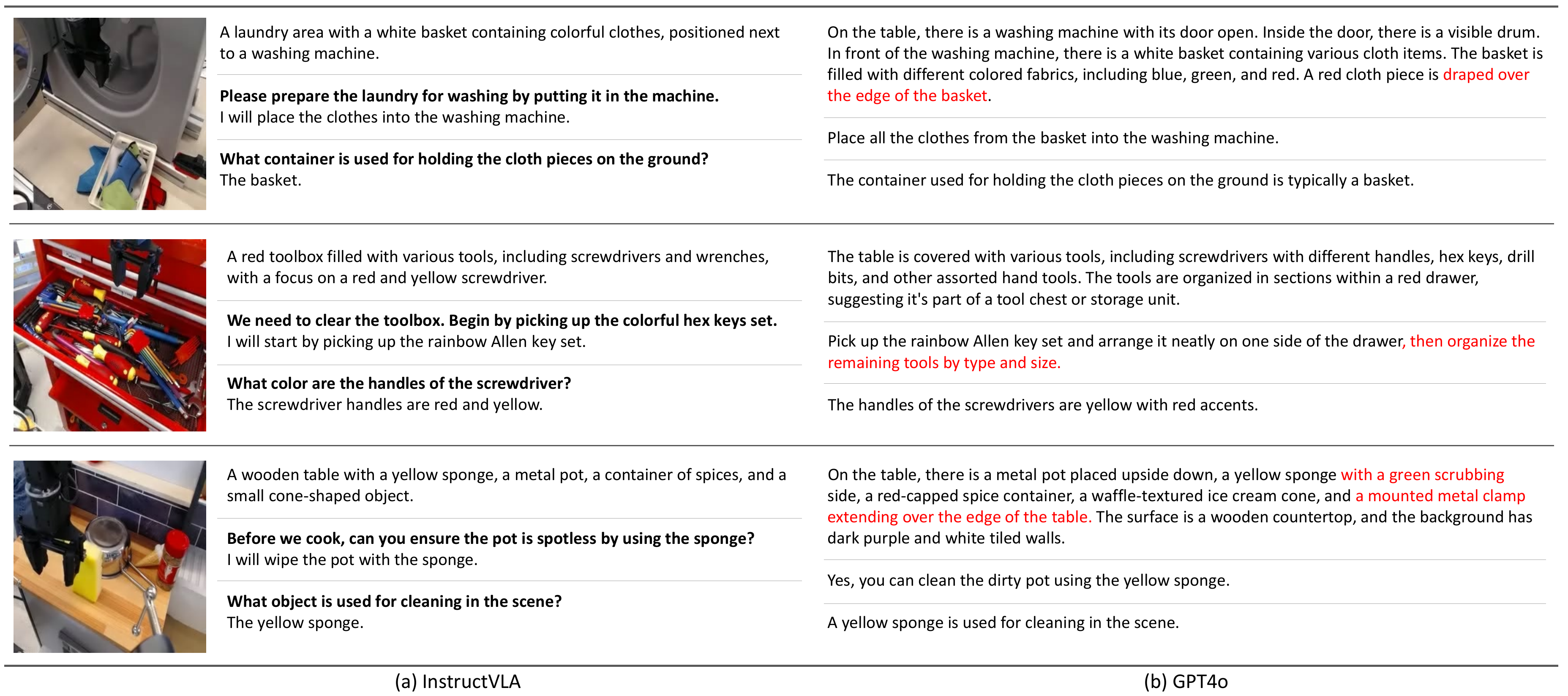}
    \caption{\textbf{Comparison with GPT-4o.} We visualize three examples from the VLA-IT language validation set. Each example includes a scenario caption (top), instruction response (middle), and question answering (bottom). The GPT-4o column displays \textbf{responses only}, as the instructions are identical across models.}
    \label{fig:language}
\end{figure*}

InstructVLA achieves state-of-the-art performance, while GPT4o demonstrates competitive results. We visualize three episodes in~\Cref{fig:language}. GPT-4o generates more detailed captions but occasionally exhibits minor hallucinations. In the instruction response task, InstructVLA produces clearer and more grounded responses compared to GPT-4o, benefiting from the integration of ground-truth atomic instructions during the data annotation process, as discussed in~\Cref{sec: Ground Truth Instruction For Data annotation}.

\clearpage
\newpage

\subsection{{Extra Manipulation Benchmark}}
\label{sec: libero}
\begin{table}[ht]
\centering
\caption{\textbf{LIBERO benchmark results.} We present the success rate and standard error for each method across four task suites, which are averaged over three random seeds with 500 trials. ``KI'' denotes knowledge insulating from\citep{driess2025knowledge}.}
 \label{tab:libero}
\resizebox{\textwidth}{!}{
\begin{tabular}{lccccc}
\toprule
& Spatial       & Object        & Goal          & 10 (Long)      & Average \\
                                    \midrule 
OpenVLA-7B~\citep{openvla} & $84.7\pm 0.9$ & $88.4\pm0.8$ & $79.2\pm 1.0$ & $53.7\pm 1.3$ & $76.5\pm  0.6$ \\
OpenVLA-OFT-7B~\citep{openvla-oft} & $97.6\pm0.9$     &$98.4\pm0.8$      & $97.9\pm1.0$ & $94.5\pm 1.3$ & $97.1 \pm0.6$ \\
SpatialVLA-2B~\citep{qu2025spatialvla}  & $88.2\pm0.5$ & $89.9\pm0.7$ & $78.6\pm0.6$ & $55.5\pm1.0$ & $78.1\pm0.7$ \\
$\pi_0$-2B~\citep{pi_0} & $96.8\pm0.8$          & $98.8\pm0.9$ & $95.8\pm1.1$ & $85.2\pm1.2$ & $94.2\pm0.9$\\
$\pi_0$-FAST-2B~\citep{pertsch2025fast}   & $96.4\pm0.7$ & $96.8\pm 0.7 $ & $88.6\pm1.0$ & $60.2\pm1.4 $ & $85.5\pm1.0$ \\
CoT-VLA~\cite{zhao2025cot} & {$87.5\pm1.4$} & {$91.6\pm0.5$} & {$87.6\pm0.6$} & {$69.0\pm0.8$} & {$81.1\pm0.6$} \\ \midrule
GR00T-N1-1.34B~\citep{bjorck2025gr00t}         & $94.4\pm0.9$ & $97.6\pm 1.0$ & $93.0\pm1.2$ & $90.6\pm1.0$ & $93.9\pm1.1$ \\ 
$\pi_{0.5}$ + KI (from scratch)~\citep{pi05} & 96.6          & 97.2          & 94.6          & 84.8          &  93.3 \\
$\pi_{0.5}$ + KI (from generalist model)~\citep{pi05} & 98.0 & 97.8          & 95.6          & 85.8          &  94.3 \\ 
% DexVLA-1.5B~\cite{wen2025dexvla} & 97.2 & 99.1 & 95.6 & - & - \\
\midrule
InstructVLA (w/o wrist view) & 92.4 & 95.6	& 92.0 & 76.6 & 89.2 \\

InstructVLA-1.5B                       & $97.3\pm 0.5$ & $99.6\pm0.0$ & $96.5\pm 0.5$ & $89.8\pm 1.6$      &   $95.8\pm 0.4$ \\   
\bottomrule
\end{tabular}
}
\end{table}

\noindent\textbf{Benchmarks and baselines.} We evaluate InstructVLA on the LIBERO simulation benchmark~\citep{Libero}, which includes diverse robotic manipulation tasks in simulated environments. Following OpenVLA~\citep{openvla}, we conduct experiments on four task suites, each containing 10 tasks with 50 human-teleoperated demonstrations. These suites assess spatial reasoning (LIBERO-Spatial), object type understanding (LIBERO-Object), task-oriented behaviors (LIBERO-Goal), and generalization to long-horizon tasks involving diverse objects, layouts, and goals (LIBERO-Long).

Our baselines fall into two categories: (i) generalist manipulation policies, including OpenVLA~\citep{openvla}, OpenVLA-OFT~\citep{openvla-oft}, SpatialVLA~\citep{qu2025spatialvla}, $\pi_0$\citep{pi_0}, and $\pi_0$-FAST\citep{pertsch2025fast}; and (ii) manipulation policies with multimodal ability, including GR00T-N1~\citep{bjorck2025gr00t}, and $\pi_{0.5}$\citep{pi05} with knowledge insulation\citep{driess2025knowledge}.

\noindent\textbf{Training details.} We augment InstructVLA with wrist-view images from the LIBERO training set~\citep{Libero}. Specifically, both the main and wrist-view images are provided to the VLM and the action expert. To reduce the tokenized input length, the two images are concatenated and resized into a single frame for VLM. Training follows the same hyperparameters as the Simpler-Env experiments and is performed on a single A800 node with 8 GPUs using a global batch size of 256, with evaluation every 1.5K steps.

\noindent\textbf{Results.} As shown in~\Cref{tab:libero}, InstructVLA achieves competitive performance despite not being pretrained on large-scale manipulation datasets like $\pi_{0.5}$\citep{pi05,driess2025knowledge} and using a much smaller VLM backbone than OpenVLA-OFT\citep{openvla-oft}. Compared with recent VLAs such as $\pi_0$, InstructVLA attains higher performance with a substantially smaller action model (134M versus 300M).

\subsection{Data Ablation on OpenVLA}
\label{sec: openvla data ablation}
\begin{table}[h]
\centering
\caption{\textbf{Data ablation on OpenVLA.} ``+VL'' indicates finetuning OpenVLA with the same multimodal dataset used by InstructVLA. ``+VLA-IT'' refers to finetuning OpenVLA with the same VLA-IT dataset as InstructVLA. ``+GPT4o'' denotes using GPT4o as system 2 to translate free-form instructions into atomic ones.}
\label{tab: openvla data}
\resizebox{\textwidth}{!}{
    \begin{tabular}{lccccc}
    \toprule
                    & OpenVLA (OXE) & OpenVLA + VL & OpenVLA + VL + VLA-IT & OpenVLA + VL + GPT4o & InstructVLA\\
                    \midrule
    Task Aggregation & 14.8 & 28.3 & 30.5 & 38.8 & 43.3\\
    Situated Reasoning      & 13.6 & 19.5 & 17.4 & 32.4 & 48.8\\
    Average                 & 14.2 & 23.9 & 24.0 & 35.6 & 46.0\\
    \bottomrule
    \end{tabular}
}
\end{table}

To investigate whether the performance gain of VLA-IT arises solely from the dataset itself, we reimplement the training procedure of the InstructVLA on OpenVLA~\citep{openvla}, which represents a class of models trained under the action-only paradigm. As shown in~\Cref{tab: openvla data}, OpenVLA benefits from both vision-language and VLA instruction tuning data, with the latter showing greater improvement in the task aggregation setting. This is attributed to exposure to more diverse instructions. However, performance on the situated reasoning setting remains unchanged, likely due to catastrophic forgetting caused by the action-only training paradigm, which limits OpenVLA’s ability to leverage the VLM's reasoning ability through simple finetuning.

The greatest performance gain is observed when GPT-4o is introduced as an auxiliary System 2 in both evaluation settings. However, overall performance remains inferior to InstructVLA, as GPT-4o cannot fully ground free-form instructions to the atomic skills on which OpenVLA is pretrained.

\subsection{Real-world Ablation}
\label{sec:realworld_ablation}
\begin{figure*}[ht]
    \centering
    \includegraphics[width=0.9\linewidth]{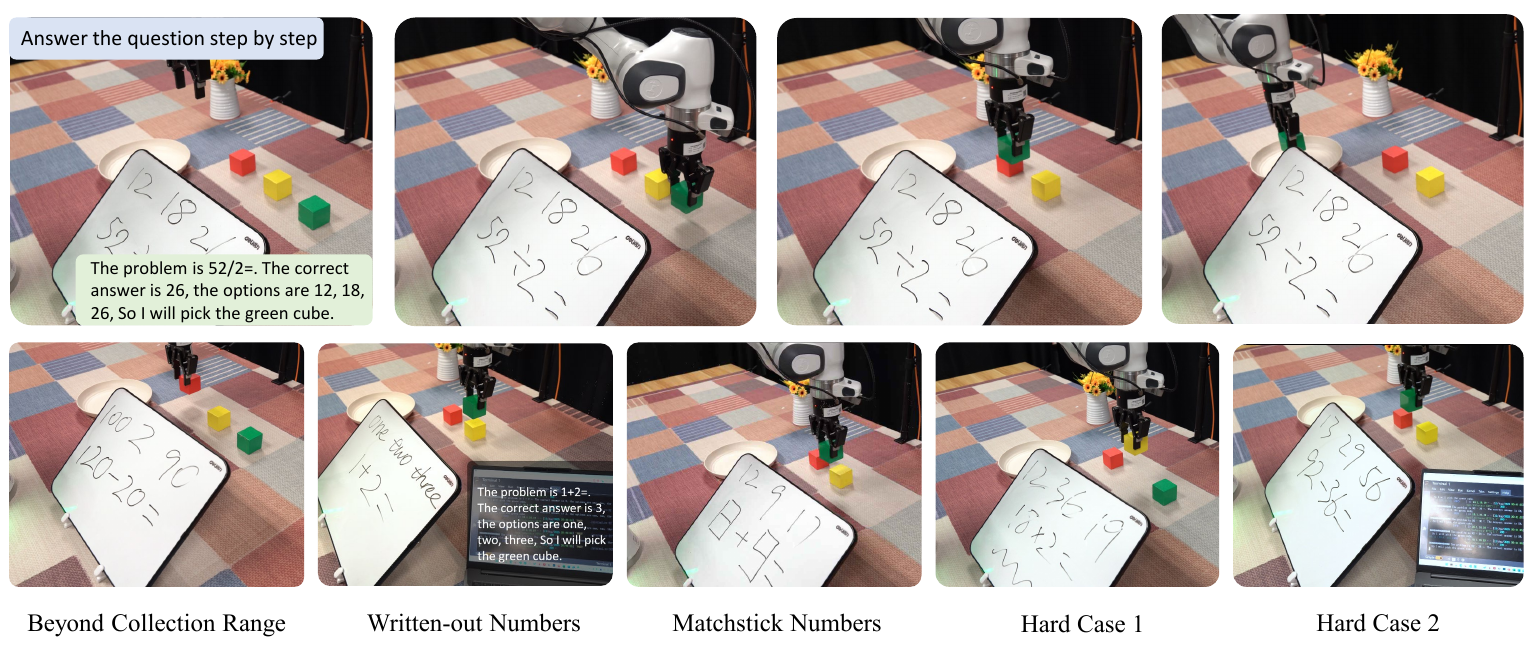}
    \caption{\textbf{Real-world ablation study.} The first row depicts the reasoning responses and the rolled-out actions, while the second row illustrates five categories of generalization.}
    \label{fig:supp_math}
\end{figure*}

\begin{figure*}[ht]
    \centering
    \includegraphics[width=0.9\linewidth]{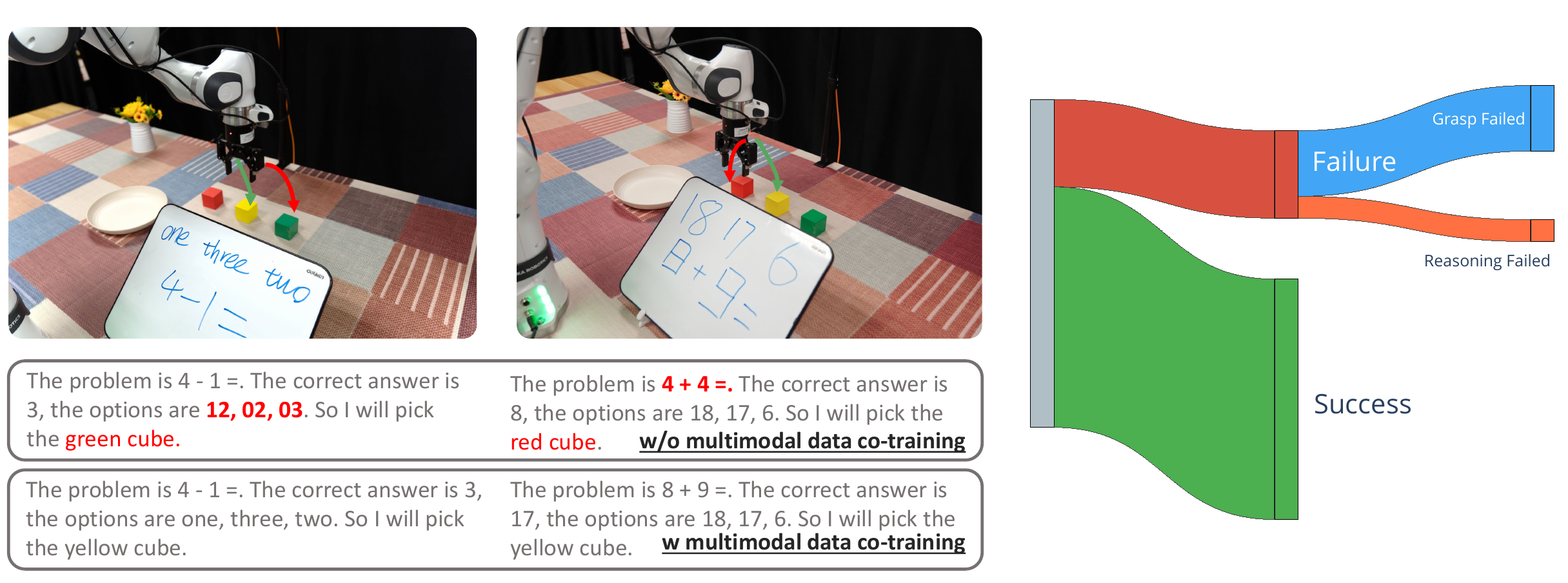}
    \caption{\textbf{Reasoning examples.} Two evaluation cases are presented to illustrate the role of multimodal datasets. We further summarize the results of InstructVLA in a Sankey diagram.}
    \label{fig:supp_math_example}
\end{figure*}

\noindent\textbf{Setup.} This case study evaluates the role of multimodal datasets in manipulation tasks. The robot setup follows our few-shot Frank evaluation. As shown in~\Cref{fig:supp_math,fig:supp_math_example}, the model must first perform OCR to recognize the formula on the board and its answer options, then compute the result, and finally control the robot to grasp the correct object. This task mirrors a shopping scenario where robots often need to read prices and perform simple calculations to satisfy a requirement. The study jointly assesses OCR and calculation abilities, which are expected to benefit from multimodal data. To reduce bias, each case is evaluated three times with different target objects. In total, 250 training cases are collected but excluded from evaluation.

The in-domain tasks are defined as calculations within the range of the training data and written in a similar format. Generalization tasks are divided into five types: (1) Beyond Collection Range, (2) Written-out Numbers, (3) Matchstick Numbers, (4) Hard Case 1 (digits partially occluded with superimposed lines), and (5) Hard Case 2 (involving more complex calculations). 

\noindent\textbf{Analysis.} By co-training with a general multimodal dataset, we observe that InstructVLA performs better on the tasks of \textit{Written-out Numbers}, \textit{Matchstick Numbers}, and \textit{Hard Case 1.} We attribute this improvement to the inclusion of general OCR data within the multimodal dataset. Although the multimodal dataset is unfiltered (i.e., identical to the corpus used for training a VLM such as Bunny), it nonetheless enhances the instruction generalization for these specific tasks. 

The SOTA VLA $\pi_0$~\citep{pi_0}, although pretrained on DROID~\cite{khazatsky2024droid}, however, produces near-random results: although each grasp is executed precisely, the model frequently selects the wrong target object. Interestingly, when the third-view camera, which capturing the board with expressions and options, is masked, $\pi_0$ still behaves similarly. This suggests that $\pi_0$ largely ignores reasoning cues and overfits to the wrist view. While it performs precise grasping, the overall outcomes remain unsatisfactory.

\section{Extra Related Works}

\label{sec: extra related works}

In this section, we delineate the distinctions between InstructVLA and several similarly named methods that differ substantially in their conceptual foundations and objectives.

\subsection{Embodied Instruction Tuning}

\noindent\textbf{Vision-Action Instruction Tuning.} The concept of Vision-Action Instruction Tuning is introduced in LLARVA~\citep{niu2024llarva}, which unifies robotic tasks through structured prompts and 2D trace supervision for cross-embodiment pretraining. In contrast, InstructVLA extends this idea by focusing on preserving the multimodal knowledge of VLMs and bridging high-level human instructions with low-level manipulation skills, enabling generalization to diverse tasks that require common-sense reasoning.

\noindent\textbf{Visuomotor Instruction Tuning.} The concept of Visuomotor Instruction Tuning is purposed in LLaRA~\citep{li2024llara}. This approach formulates robot policies as visuo-textual conversations and produces 2D keypoints and rotations for manipulation. However, it functions primarily as a high-level planner, and its outputs require additional adaptation before being directly executed on robots.

\subsection{Multi-stage Training}

\noindent\textbf{OpenVLA-OFT.}
OpenVLA-OFT~\citep{openvla-oft} extends OpenVLA~\citep{openvla} by incorporating FiLM layers, Parallel decoding, MLP action head, and has been applied to fine-tuning on smaller simulation datasets such as LIBERO~\cite{Libero}. This approach demonstrates the effectiveness of architectural enhancements for improving manipulation performance in specific domains. However, while these techniques improve in-domain performance, they fall short in reasoning-centric settings such as SimplerEnv-Instruct, as shown in~\Cref{fig:ablation_instruct} (b).
In contrast, our work moves beyond architectural modifications by emphasizing generalizable manipulation with textual reasoning through MoE adaptation, latent action methods, and a comprehensive data and evaluation pipeline. With the proposed VLA-IT training paradigm, our generalist model achieves nearly a 2$\times$ improvement over models that rely solely on architectural designs.

\noindent\textbf{Embodied Chain-of-Thought.}
ECoT~\citep{ecot} introduces chain-of-thought (CoT) supervision to link reasoning with manipulation and follows a standard ``pretrain-then-instruction-tune'' paradigm. However, it relies on full-model pretraining fine-tuning, as in OpenVLA~\citep{openvla}, which leads to catastrophic forgetting of vision-language capabilities.
\textit{In contrast, InstructVLA adopts a two-stage design: the first stage injects action-generation ability while deliberately preserving the multimodal knowledge of the pretrained VLM.} This approach ensures that the model retains open-world understanding and general multimodal reasoning, both of which are largely lost in ECoT. The second stage then strengthens multimodal reasoning and manipulation alignment. Consequently, InstructVLA supports broader inference modes (reasoning + manipulation, direct manipulation, and multimodal VQA) and achieves stronger performance with substantially fewer trainable parameters.

{
\noindent\textbf{Visual Chain-of-Thought.} CoT-VLA~\citep{zhao2025cot} enhances manipulation by generating future image frames as visual chain-of-thought goals before predicting actions. While effective for goal specification, this approach relies on heavy video-generation supervision and does not exploit strong VLM pretraining for visual-language reasoning.
}
\clearpage
\newpage

\section{Case Study}
\label{sec: failure cases analysis}

\subsection{Reasoning Cases in SimplerEnv-Instruct}

\begin{figure}[ht]
    \centering
    \includegraphics[width=1\linewidth]{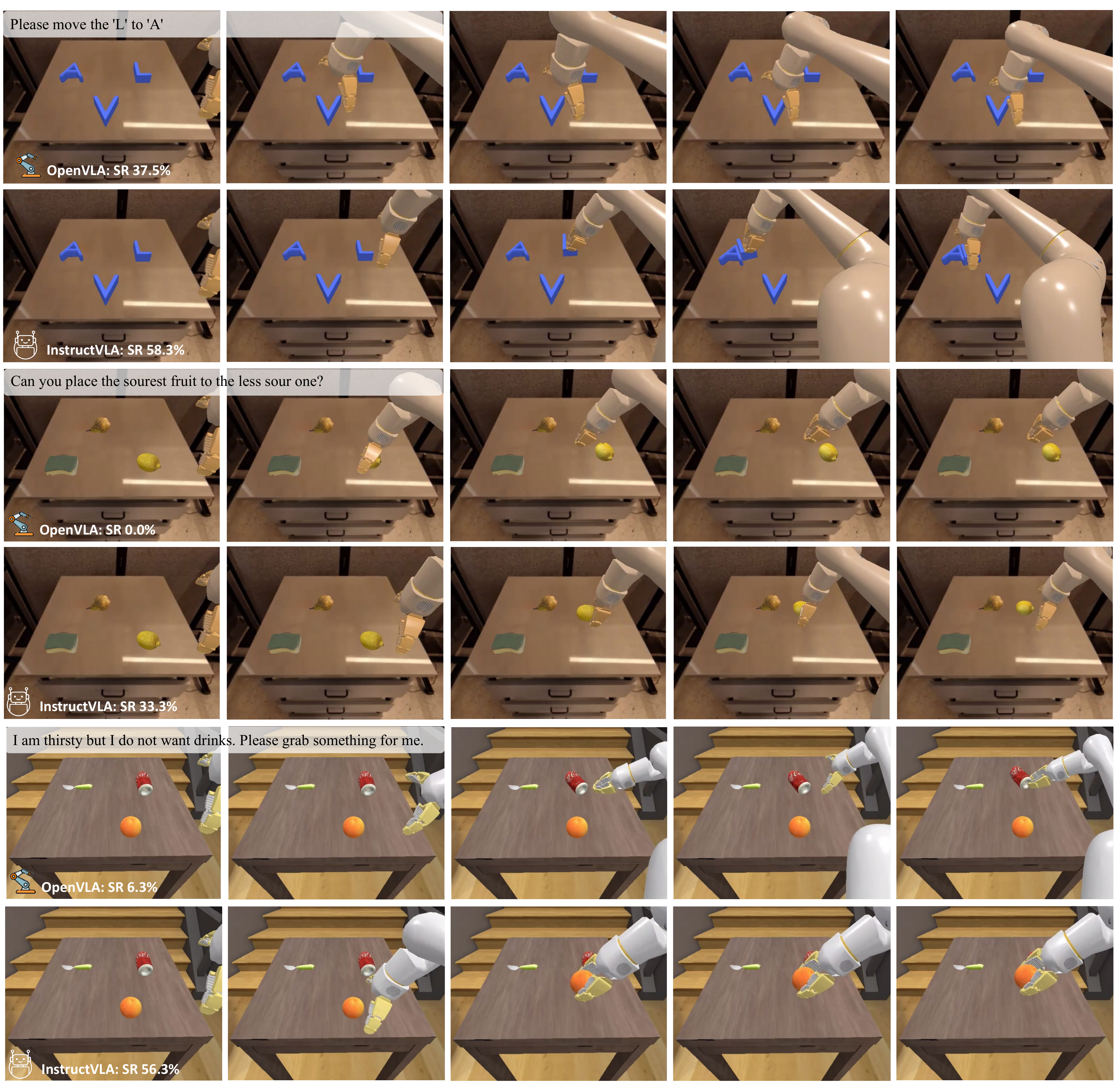}
    \caption{\textbf{Reasoning cases in SimplerEnv-Instruct.} Three cases of the VL fine-tuned OpenVLA and InstructVLA-Generalist. ``SR'' denotes success rate.}
    \label{fig:openvla}
\end{figure}

We present three representative reasoning cases in~\Cref{fig:openvla}. In the first example, OpenVLA fails to associate the letters ``V'' and ``L'' with their corresponding shapes in the image, resulting in consistent failure to grasp in all similar scenarios. In the second case, OpenVLA does not correctly associate the concept of ``sour'' with the corresponding fruit. As a result, its action is influenced by both the pear and lemon, leading to a grasp attempt between them that strikes the table. In the final example, OpenVLA fails to interpret the negation in the instruction and incorrectly grasps Coke instead of orange.

\clearpage
\newpage

\subsection{Failure Cases}

\begin{figure}[h]
    \centering
    \includegraphics[width=1\linewidth]{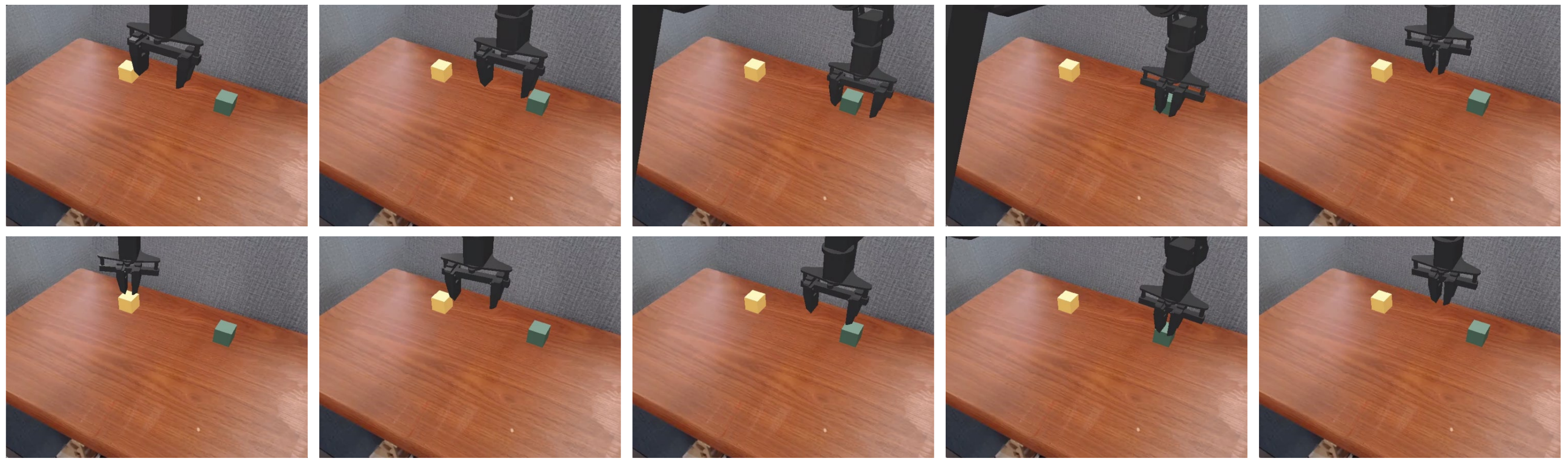}
    \caption{\textbf{Failure case 1 of InstructVLA.} The model receives only a third-person view image as visual input, making it difficult to estimate depth or the gripper’s relative position to the object. Consequently, it fails to grasp the object accurately, despite the gripper appearing aligned with the target in the image.}
    \label{fig:failure_case1}
\end{figure}

\begin{figure}[h]
    \centering
    \includegraphics[width=1\linewidth]{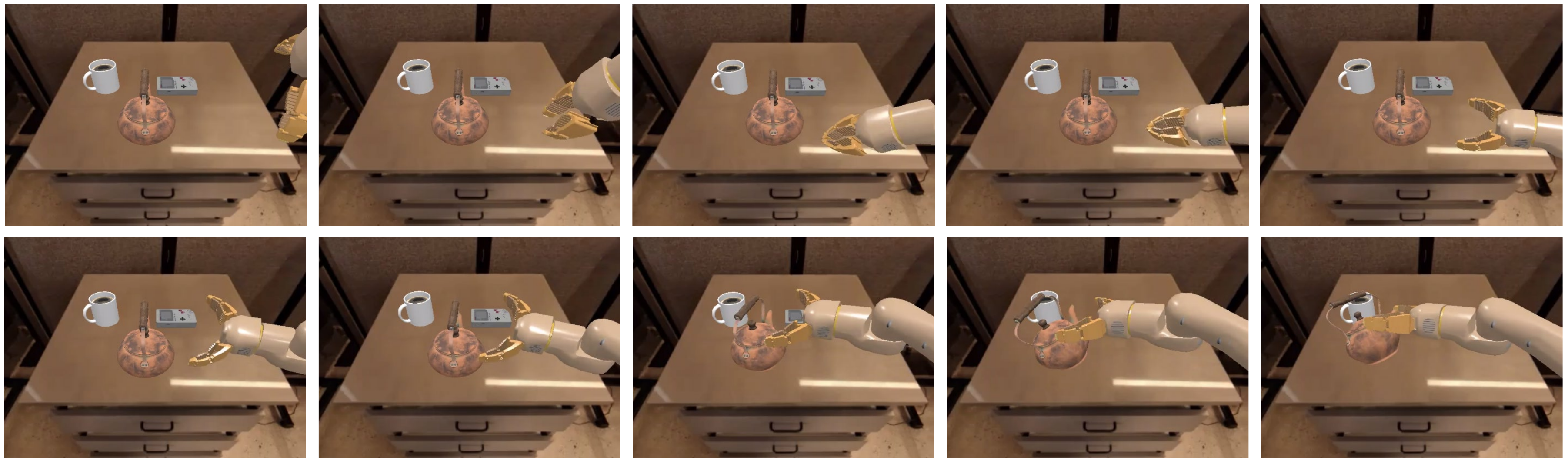}
    \caption{\textbf{Failure case 2 of InstructVLA.} The model fails to accurately estimate depth due to the real-to-sim gap, specifically the absence of arm reflection on the table, which causes the robot to become stuck in an out-of-distribution position.}
    \label{fig:failure_case2}
\end{figure}

We illustrate two representative failure cases of InstructVLA in~\Cref{fig:failure_case1,fig:failure_case2}. While some failures may result from the real-to-sim gap, incorporating additional sensory inputs such as depth information and robot state may enhance performance. We leave this exploration for future work. Additionally, we observe that the model achieves higher success rates in language responses than in action execution, suggesting that multimodal understanding is more readily transferable than manipulation skills. This highlights a fundamental challenge in the development of embodied models.
 
\clearpage
\newpage

\subsection{GPT4o as the Auxiliary System 2}

\begin{figure}[h]
    \centering
    \includegraphics[width=1\linewidth]{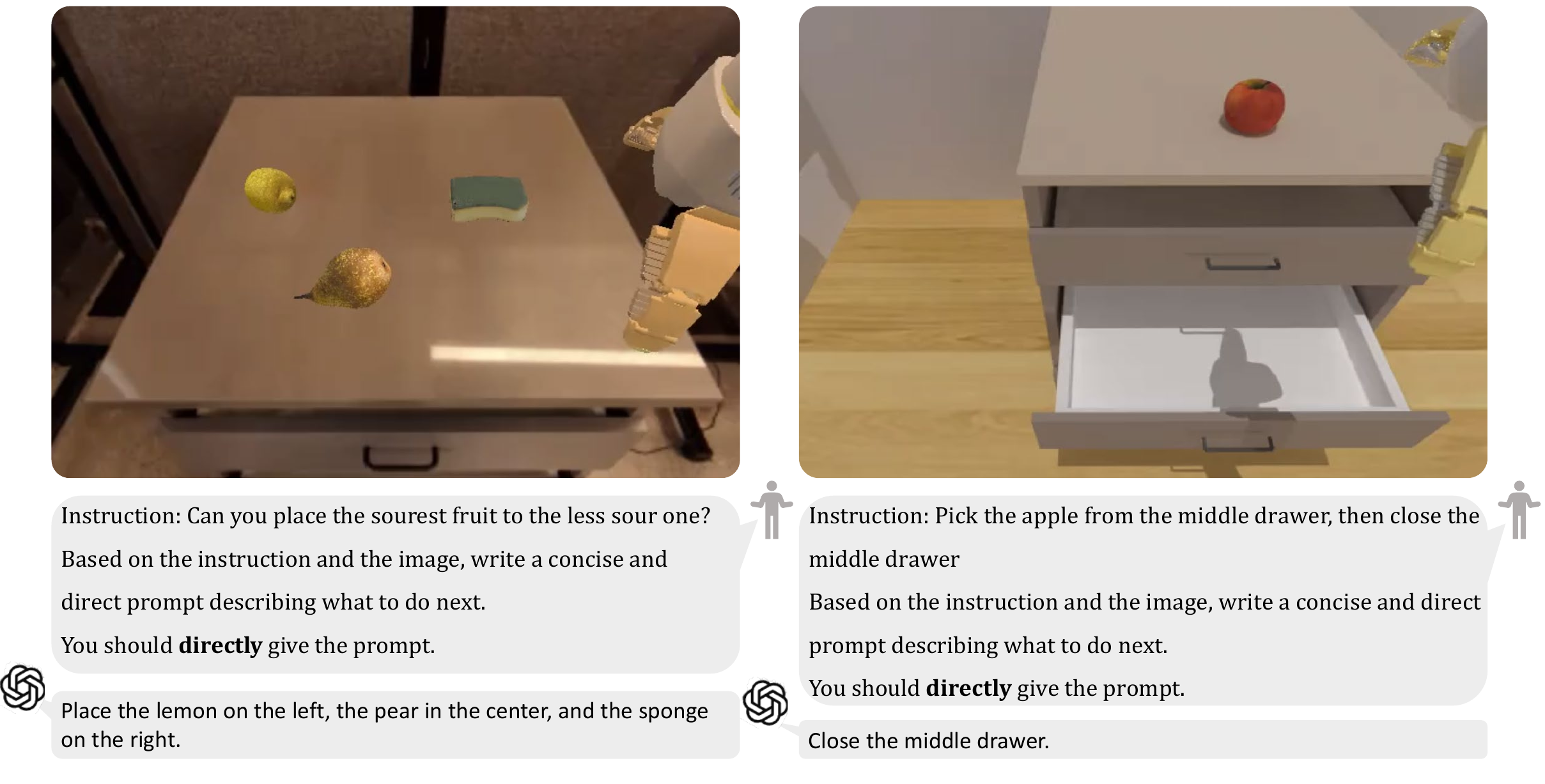}
    \caption{\textbf{GPT-4o as the auxiliary system 2.} We prompt GPT-4o with the first image from the environment along with the instruction, asking it to rewrite the prompt in a simple and clear format.}
    \label{fig:gpt sys2}
\end{figure}

A strong baseline for InstructVLA integrates an expert model capable of executing atomic instructions with GPT-4o as an instruction parser to decompose complex, free-form commands for decision-making~\citep{hirobot,gao2025genmanip}. The prompt used is listed in Prompt~\hyperref[prompt: gpt system2]{1}, and it was evaluated and refined on 20 test cases from the task aggregation to ensure reliable performance. Results on additional test cases are presented in~\Cref{fig:gpt sys2}. GPT-4o successfully identified the atomic instruction in the second case but failed in the first. 

During evaluation, GPT-4o is invoked only in the initial step to ensure an unobstructed view of the scene and to generate a free-form instruction. We do not provide a closed set of task-relevant instructions for selection, as the training set (\Cref{fig:data_analysis}) lacks sufficient diversity in instructions and objects, and therefore does not adequately cover the evaluation settings. Across 80 evaluation cases, GPT-4o frequently fails in physical grounding, maintaining coherence, and accurately interpreting the scene.

{\scriptsize
\phantomsection
\label{prompt: gpt system2}
\begin{tcolorbox}[colback=white, colframe=black!75!white, title=GPT-4o System-2 Prompt]

Instruction: Can you place the sourest fruit to the least sour one?

Based on the instruction and the image, write a concise and direct prompt to describe what to do next. 

You should \textbf{directly} give the prompt.

\end{tcolorbox}

}

\clearpage
\newpage

\section{Data Annotation Details and Analysis}
\label{sec: Data Annotation Details and Analysis}

 The data analysis and GPT4o prompt are listed as follows (\Cref{fig:data_analysis} and Prompt~\hyperref[prompt: data anno]{2}).

\subsection{Language motion pre-training data}

\label{sec: language motion}
 
Language motion~\citep{rth} provides intuitive linguistic descriptions of basic end-effector movements, which can be distilled into latent actions.
{
For each episode, we extract a sequence of low-level motion primitives from the robot state trajectory.
Let $s_t \in \mathbb{R}^8$ denote the state at time $t$, consisting of the end-effector position $p_t \in \mathbb{R}^3$, orientation quaternion $q_t \in \mathbb{R}^4$ (in $xyzw$ order), and the scalar gripper state $g_t \in \mathbb{R}$
\begin{equation}
s_t = (p_t, q_t, g_t).
\end{equation}
We process overlapping windows of length $n$, $(s_t, \dots, s_{t+N})$, and summarize each window by the displacement between the first and last state
\begin{equation}
\Delta p = p_{t+N} - p_t, \quad
\Delta q = q_{t+N} \otimes q_t^{-1}, \quad
\Delta g = g_{t+N} - g_t.
\end{equation}

 The rotational displacement $\Delta q$ is converted to Euler angles $(r, \phi, \psi)$ in the \texttt{xyz} convention (roll, pitch, yaw). We then form a 6D continuous motion descriptor
\begin{equation}
d = (d_x, d_y, d_z, d_{\text{pitch}}, d_{\text{yaw}}, d_{\text{grip}}),
\end{equation}
where $d_{x,y,z}$ correspond to the clipped $\Delta p$, and $d_{\text{grip}} = \Delta g$. Finally, we quantize each dimension with a symmetric threshold $\theta$:
\[
v_i = 
\begin{cases}
+1, & d_i > \theta,\\
0,  & |d_i| \le \theta,\\
-1, & d_i < -\theta,
\end{cases}
\]
obtaining a discrete motion code $v \in \{-1,0,1\}^6$. This code is then mapped to a natural-language description (e.g., “move forward”, ``tilt up'', ``close gripper'') using a fixed vocabulary, as shown in~\Cref{tab:motion_primitives}. If all dimensions are zero, the primitive is labeled as \textit{stop}. We visualize example annotations from a representative episode in~\Cref{fig:language motion}. The language motions are concatenated into the model’s response along with the corresponding user prompt.

\begin{table}[t]
    \centering
    \caption{Definition of the 6D motion-primitive code $v \in \{-1,0,1\}^6$ and its natural-language verbalization. If no motion is detected, the default label is \textit{stop}.}
    \label{tab:motion_primitives}
    \begin{tabular}{cclll}
        \toprule
        Index & Symbol & Physical meaning & Values & Verbalization \\
        \midrule
        0 & $v_x$ & Translation along $x$ & $-1,0,1$ & backward / --- / forward \\
        1 & $v_y$ & Translation along $y$ & $-1,0,1$ & right / --- / left \\
        2 & $v_z$ & Translation along $z$ & $-1,0,1$ & down / --- / up \\
        3 & $v_{\text{pitch}}$ & Pitch rotation & $-1,0,1$ & tilt down / --- / tilt up \\
        4 & $v_{\text{yaw}}$ & Yaw rotation & $-1,0,1$ & rotate clockwise / --- / rotate counterclockwise \\
        5 & $v_{\text{grip}}$ & Gripper motion & $-1,0,1$ & open gripper / --- / close gripper \\
        \bottomrule
    \end{tabular}
\end{table}
\begin{figure}[ht]
    \centering
    \includegraphics[width=1\linewidth]{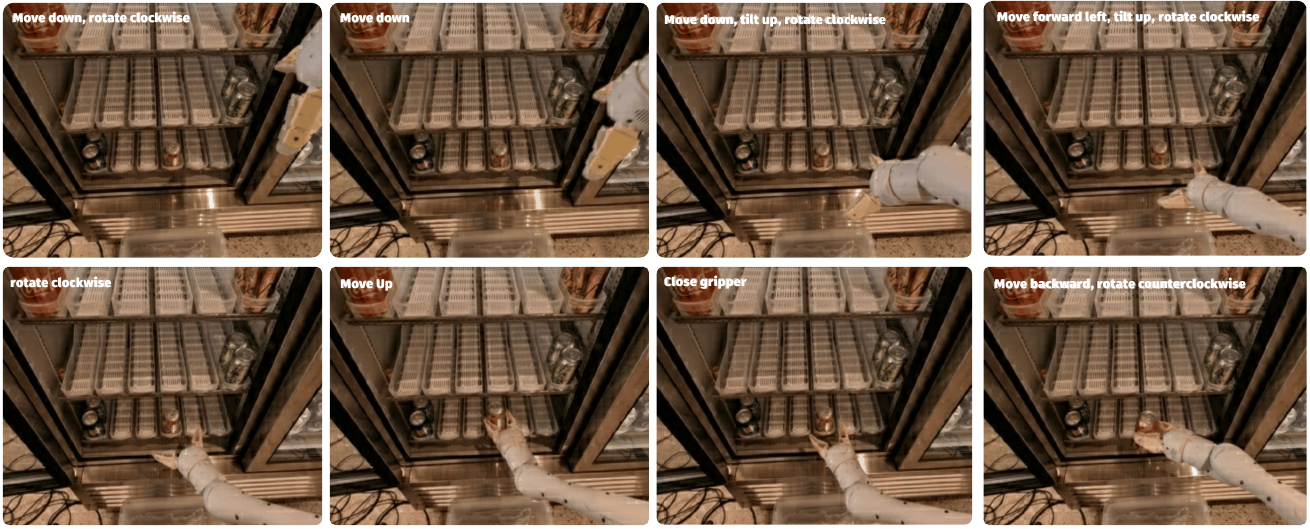}
    \caption{Language motion examples}
    \label{fig:language motion}
\end{figure}
}
 
\subsection{Task Diversity Analysis}

We categorize tasks into two broad classes: \textbf{Command Rewriting / Context Creation} and \textbf{Question Answering}. Each class includes several common task types:

\subsubsection*{Command Rewriting / Context Creation}
\begin{itemize}[leftmargin=0.2in]
    \item \textbf{Complex Object Referencing:} Uses attributes, pronouns, or relational terms to reference an object.\\
    \emph{Example:} ``Place the red item next to the box.''
    \item \textbf{Novel Action Referencing:} Rephrases a previously known action using a different verb or motion.\\
    \emph{Example:} ``Shut the drawer'' (instead of ``Close the drawer'').
    \item \textbf{Negative Task Specification:} Specifies the correct action by negating incorrect alternatives.\\
    \emph{Example:} ``I'm thirsty, but I don't want sparkling water—bring me something else.''
    \item \textbf{Subtask Identification:} Isolates a step from a multi-step instruction with a clear sequential order.\\
    \emph{Example:} From ``Take the spoon out of the top drawer,'' execute only the first step.
    \item \textbf{Situated Task Identification:} Infers the required action based on contextual cues or situational conditions.\\
    \emph{Example:} ``I want to clean the table. What should I use?''
    \item \textbf{Direct Instruction:} Provides an explicit and unambiguous command.\\
    \emph{Example:} ``Organize the drinks by putting the green can next to the Coke can.''
    \item \textbf{Tool-Use Understanding:} Refers to an object by its utility or function rather than its name.\\
    \emph{Example:} ``Hand me something to cut with'' (instead of ``Use the knife'').
\end{itemize}

\subsubsection*{Question Answering}
\begin{itemize}[leftmargin=0.2in]
    \item \textbf{Quantitative Identification:} Requires determining the number or quantity of items.\\
    \emph{Example:} ``How many apples are on the table?''
    \item \textbf{Spatial Identification:} Involves spatial relationships between objects or with the user.\\
    \emph{Example:} ``Is the cup on the left or the right of the plate?''
    \item \textbf{Visual Identification:} Focuses on appearance-based attributes such as color or shape.\\
    \emph{Example:} ``Which one is the metallic silver object?''
    \item \textbf{Commonsense Answering:} Requires everyday reasoning or world knowledge.\\
    \emph{Example:} ``Which of these would you use to cut paper?''
    \item \textbf{State Identification:} Determines the current condition or status of an object.\\
    \emph{Example:} ``Is the drawer currently open or closed?''
\end{itemize}

\begin{figure}
    \centering

    \begin{minipage}{0.9\linewidth}
        \centering
        \includegraphics[width=\linewidth]{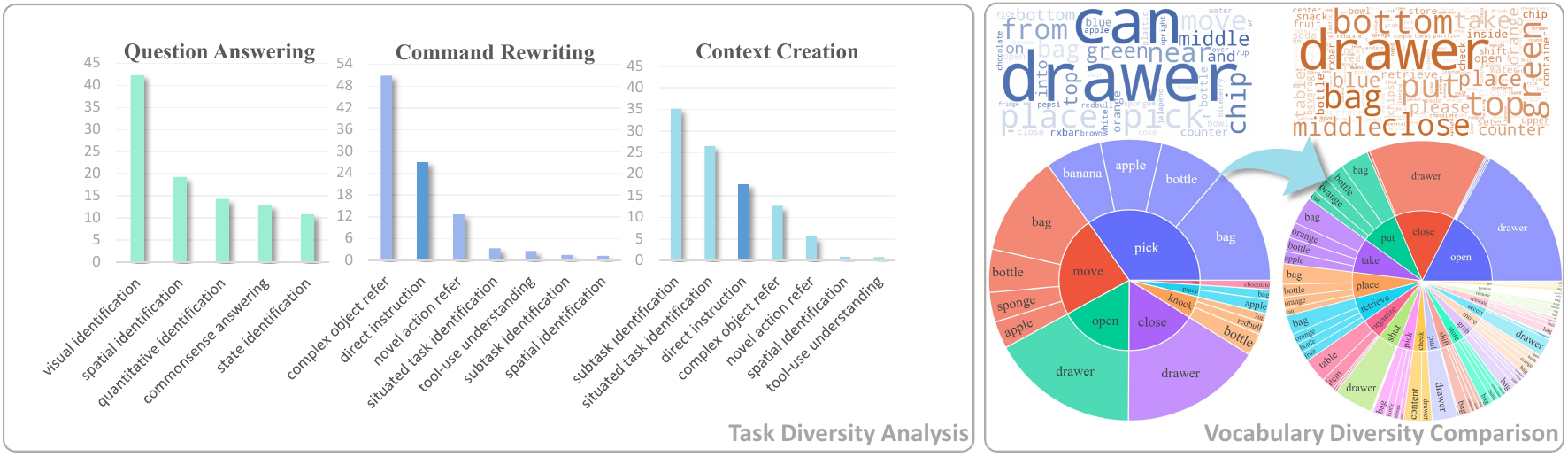}
        \caption{\textbf{Data analysis.} Left: We manually identify common task categories and calculate the distribution. The proportion of direct prompts is reduced in favor of more diverse, free-form expressions. Right: Word cloud and verb-noun analyses compare the original Fractal instructions with the VLA-IT corpus.}
        \label{fig:data_analysis}
    \end{minipage}
    
    \vspace{0.8cm}
    
    \begin{minipage}{0.9\linewidth}
        \centering
        \includegraphics[width=\linewidth]{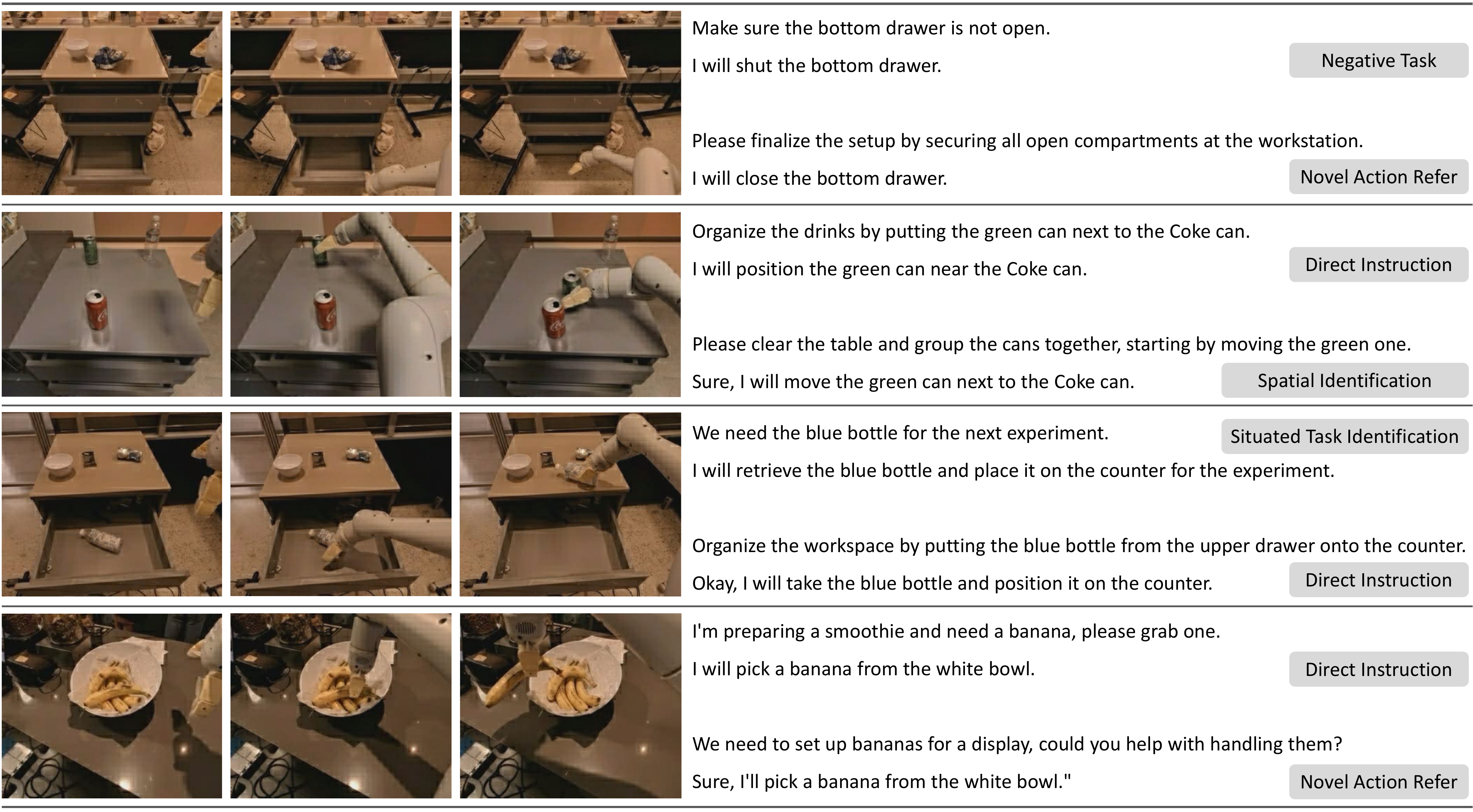}
        \caption{More VLA instructions on Fractal dataset.}
        \label{fig:examples_fractal}
    \end{minipage}
    
    \vspace{0.8cm}
    
    \begin{minipage}{0.9\linewidth}
        \centering
        \includegraphics[width=\linewidth]{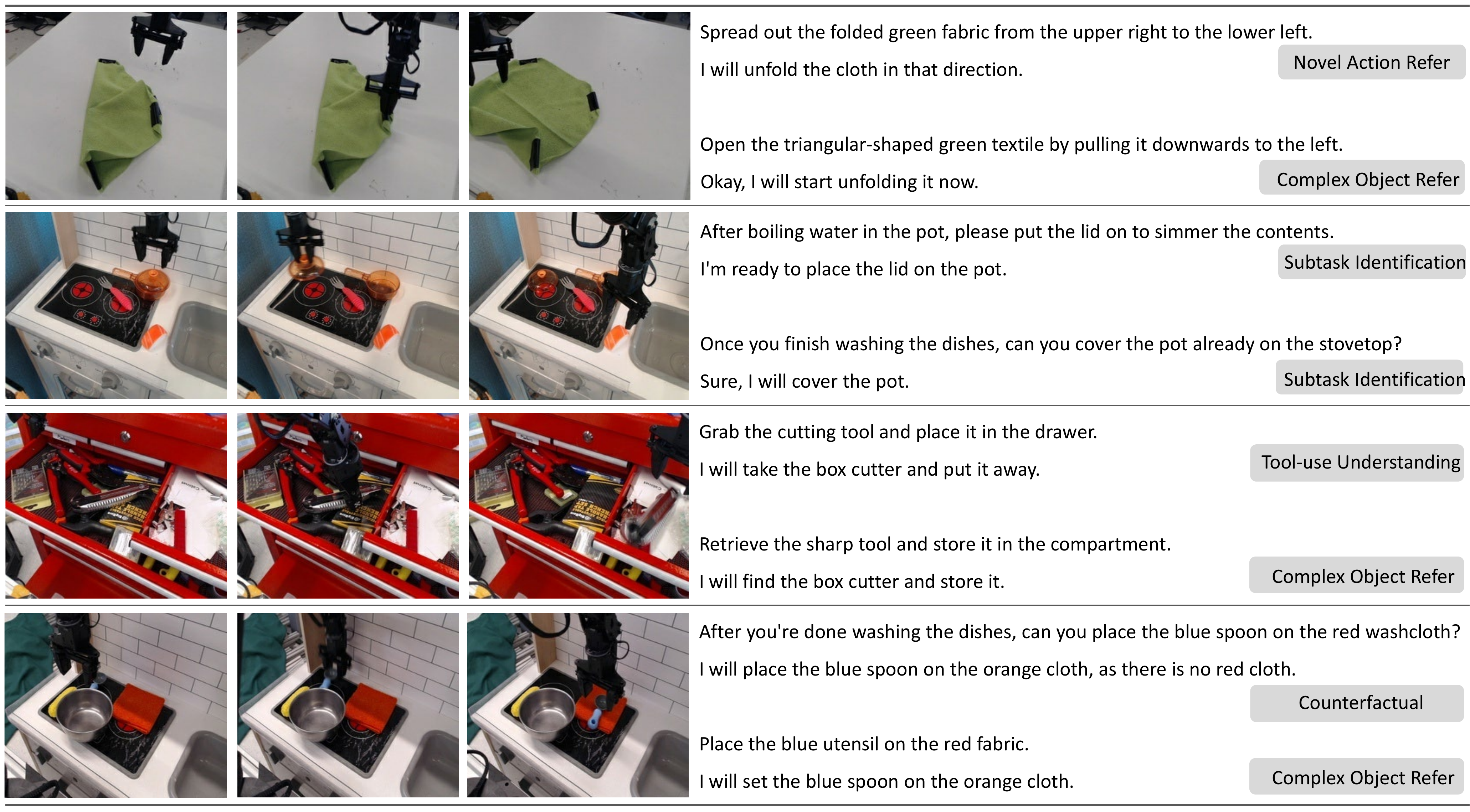}
        \caption{More VLA instructions on Bridge dataset.}
        \label{fig:examples_bridge}
    \end{minipage}
    
\end{figure}

The data examples for VIA-IT are provided in~\Cref{fig:examples_fractal,fig:examples_bridge}.

\subsection{Prompting}

The Prompt~\hyperref[prompt: data anno]{2}, along with three images captured at the beginning, middle, and end of each episode, is packaged and sent to GPT-4o. Episodes from the Bridge dataset~\citep{Bridge_data} that lack valid instructions are excluded from annotation.

\newpage
\phantomsection
\label{prompt: data anno}
{\scriptsize
\begin{tcolorbox}[colback=white, colframe=black!75!white, title=Data Annotation Prompt]

Imagine a robot assistant operating in a laboratory or household environment. The robot is expected to follow diverse commands based on realistic tasks and human interactions. Your task is to:

\begin{enumerate}[leftmargin=0.2in]
  \item Write a caption to describe the visual scene shown in the \textbf{first image}. You should \textbf{NOT} include the robot itself here.
  \item Based on the given robot task description and the images, generate new user instructions and corresponding robot responses with QA pairs.
\end{enumerate}

The new user instructions should align with the actions performed by the robot in the images and with the environment shown in the images. You are required to produce three categories of instructions:

\begin{enumerate}[leftmargin=0.2in]
  \item \textbf{Command Rewriting (CR)}: Rephrase the task description using diverse language styles and vocabulary. You may refer to objects by their utility, color, shape, or other attributes, but ensure the attribute you use is unique to each object.
  \item \textbf{Context Creation (CC)}: Generate detailed scenarios where the robot needs to perform the given instruction. The situation should involve realistic surroundings or tasks where this instruction would be necessary. You may also simulate a long-horizon task based on the context provided by the image. Your generated question should \textbf{NOT} include the answer itself.
  \item \textbf{Scene-related Commonsense QA (QA)}: Generate some other QA pairs that are related to the scene. The answer should be concise and consistent among the three images.
\end{enumerate}

For each instruction, provide a concise robot response that clearly (use simple words) communicates the next action the robot will take. \textbf{Do not chain multiple actions together using phrases like "and then."} If necessary, the response may include a brief explanation of the reasoning. Avoid repeating the instruction in the response.

\textbf{Response Format}: You MUST respond in JSON format. You should include \texttt{"Caption"}, \texttt{"CR"}, \texttt{"CC"}, and \texttt{"QA"} in your response. You should create 1-3 entries for each of CR, CC, and QA.

\textbf{Example 1}: For the instruction ``Close middle drawer": \\
\textit{(Corresponding three images omitted)} \\
\textbf{Caption}: ``A table with a Coke and chips on top, with its middle drawer open.''

\begin{lstlisting}
{
  "Caption": "A table with a Coke and chips on top, with its middle drawer open.",
  "CR": [ { "question": "Push the middle drawer closed.", 
            "answer": "Ok, I will close it." },
          { "question": "Ensure the center drawer is closed.", 
            "answer": "I will close the drawer." } ],
  "CC": [ { "question": "I want you to take out the Coke from the middle drawer and closing it.", 
            "answer": "The Coke is on the table, and the middle drawer is empty. So, I should close the middle drawer." },
          { "question": "Please push the middle drawer shut so we can clear the workspace.", 
            "answer": "Okay, I will close the middle drawer." } ],
  "QA": [ { "question": "What is in the middle drawer?", 
            "answer": "The middle drawer is empty." },
          { "question": "How many Coke cans are on the table?", 
            "answer": "One." } ]
}
\end{lstlisting}
\textbf{Example 2}: For the instruction ``move the apple near the Coke": \\
\textit{(Corresponding three images omitted)} \\
\textbf{Caption}: ``A table with Coke, apple, and soap on it.''

\begin{lstlisting}
{
  "Caption": "A table with Coke, apple, and soap on it.",
  "CR": [ { "question": "Move the healthy food near the Coke.", 
            "answer": "The healthy food refers to the apple, and I will move the apple to the Coke." },
          { "question": "Move the apple to the cylindrical-shaped object.", 
            "answer": "Of course!" } ],
  "CC": [ { "question": "Gather all objects near the Coke, except the soap.", 
            "answer": "I will move the apple to the Coke." } ],
  "QA": [ { "question": "I'm thirsty, what can I have?", 
            "answer": "The Coke is on the table." },
          { "question": "What is the healthy food on the table?", 
            "answer": "The apple." } ]
}
\end{lstlisting}

Your task description is ``\texttt{<placeholder>}''. \\
Now give your response in JSON format.

\end{tcolorbox}
}

% TODO: discuss the success rate
\subsection{Ground Truth Instruction for Data annotation}
\label{sec: Ground Truth Instruction For Data annotation}
During data generation, we observe that GPT-4o often struggles to accurately interpret robot behavior using only the three provided images, performing noticeably worse than humans. To quantify this, we randomly sample 100 examples and prompt GPT-4o to generate our four types of annotations using a similar prompt (excluding the ground truth instruction from a human expert). We then manually evaluate the correctness of the results: a sample is scored as 1 if no obvious errors are found, 0.5 if minor errors are present, and 0 if completely incorrect. 

\begin{figure}
    \centering
    \includegraphics[width=1\linewidth]{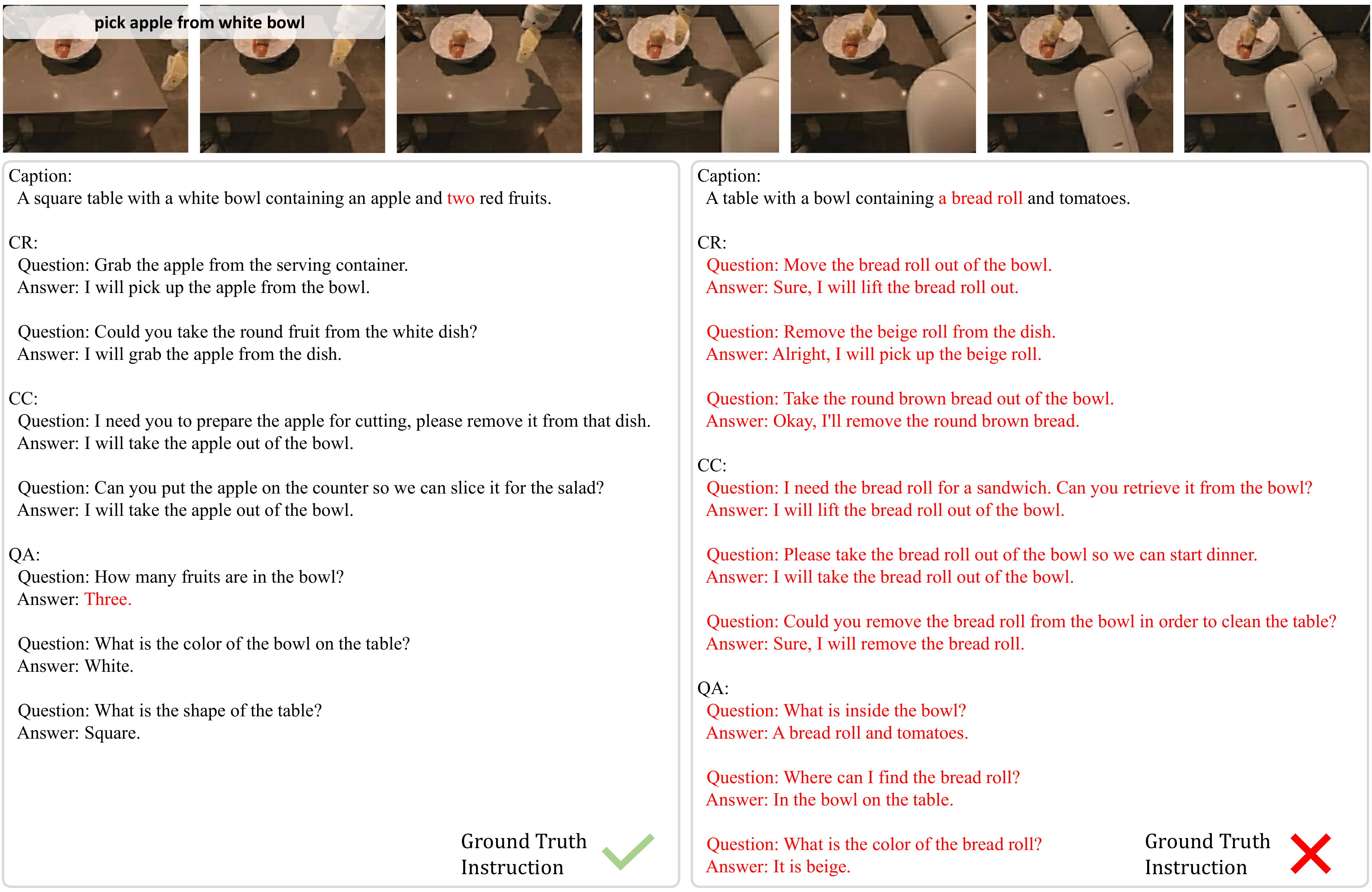}
    \caption{\textbf{Comparison of GPT annotations with and without ground truth instruction.} Errors are highlighted in red.}
    \label{fig:annotation_failures1}
\end{figure}

\begin{figure}
    \centering
    \includegraphics[width=1\linewidth]{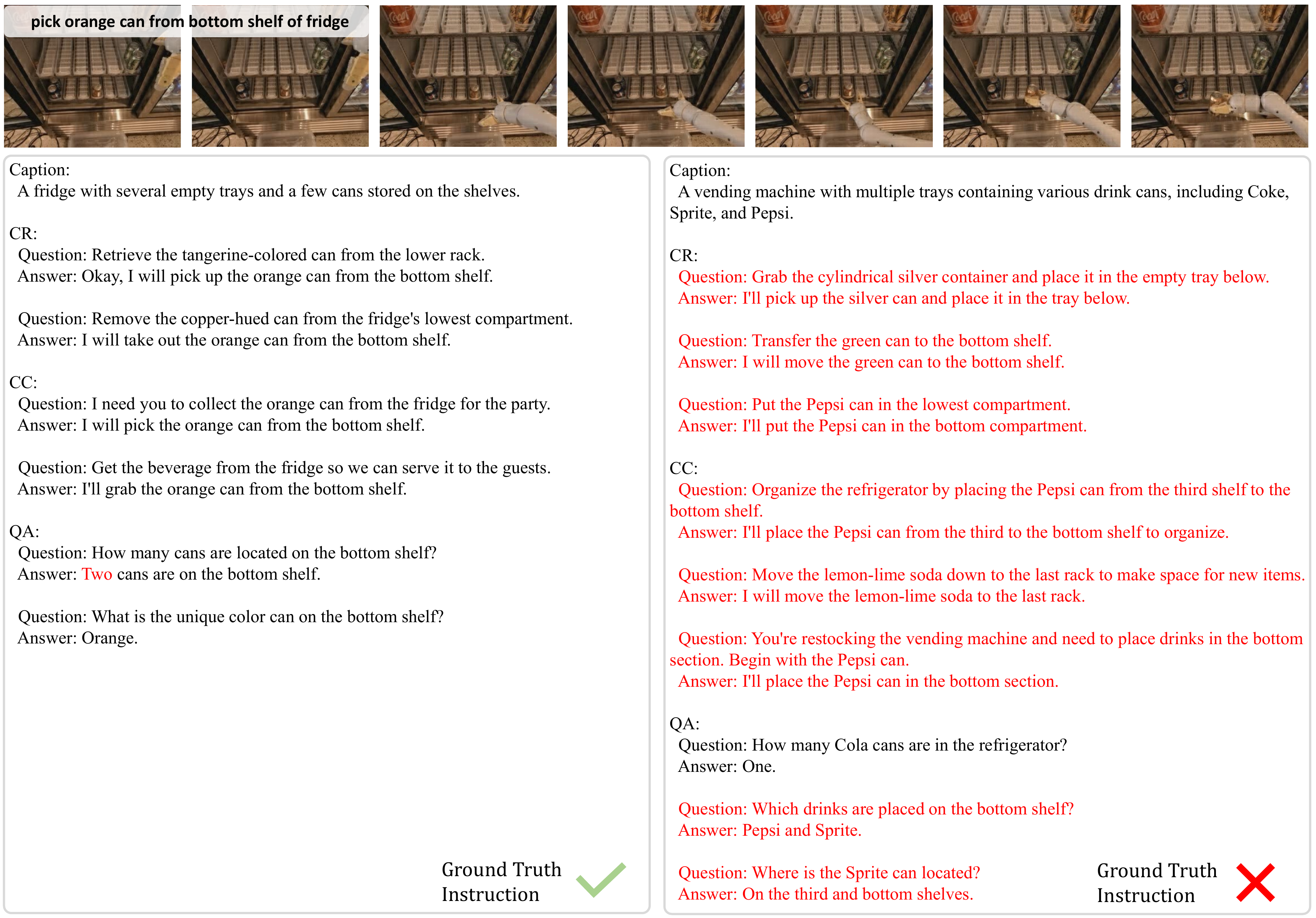}
    \caption{\textbf{Comparison of GPT annotations with and without ground truth instruction.} Errors are highlighted in red. In this case, GPT-4o incorrectly infers the temporal sequence of actions without access to the instruction.}
    \label{fig:annotation_failures2}
\end{figure}

The results are summarized in~\Cref{tab:anno,tab:error_types}, with two representative cases illustrated in~\Cref{fig:annotation_failures1,fig:annotation_failures2}. In the first case, GPT-4o hallucinates the robotic arm as a bread roll, leading to an incorrect caption and instruction. In the second, it reverses the temporal order of actions, resulting in an inaccurate annotation.

We attribute this performance gap to GPT-4o's lack of temporal grounding and the low visual quality of images in manipulation datasets. In contrast, human-provided instructions inherently encode temporal links across the image sequence by grounding the task in context, identifying target objects, and specifying corresponding robot actions. This finding underscores that, despite their impressive capabilities, even state-of-the-art VLMs lack embodied experience and temporal grounding, limiting their ability to infer fine-grained actions in robot manipulation tasks.

\begin{table}[h]
\centering
\begin{minipage}{0.48\textwidth}
\centering
\caption{\textbf{Data annotation success rate.} GPT-4o shows a significant performance drop without ground truth instructions during data annotation.}
\begin{tabular}{lc}
\toprule
Method                 & Success Rate \\
\midrule
With GT Instruction    & 95.4\%         \\
Without GT Instruction & 45.0\%         \\
\bottomrule
\end{tabular}
\label{tab:anno}
\end{minipage}%
\hfill
\begin{minipage}{0.48\textwidth}
\centering
\caption{\textbf{Distribution of common error types.} Error analysis of GPT-4o annotations generated without access to ground truth instructions, with long-tail errors omitted.}
\begin{tabular}{lc}
\toprule
\textbf{Error Type}        & \textbf{Percentage} \\
\midrule
Ignore Vision Context      & 32.5\% \\
Reverse Temporal Order     & 10.2\% \\
Minor Object Hallucination & 5.7\%  \\
\bottomrule
\end{tabular}
\label{tab:error_types}
\end{minipage}
\end{table}
% \subsection{Language Motion Examples}
% \label{sec: language motion examples}

% Language motion~\citep{rth} describes end-effector movements using natural language, enhancing the VLM's understanding of robotic manipulation. To generate such annotations, we leverage proprioceptive data that captures the end-effector’s position and orientation relative to the robot base. While the Bridge dataset~\citep{Bridge_data} adopts annotations from ECoT~\citep{ecot}, we additionally annotate the Fractal dataset~\citep{RT-1} using a similar approach. The examples on the Fractal dataset are presented in~\Cref{fig:language motion}.

% \begin{figure}[ht]
%     \centering
%     \includegraphics[width=1\linewidth]{figures/supp_language_motion.pdf}
%     \caption{Language motion examples}
%     \label{fig:language motion}
% \end{figure}

\clearpage
\newpage

\clearpage
\newpage

\section{Benchmark Details}
\label{sec: Benchmark Task Descriptions and Visualization}
\subsection{Multimodal}
\label{sec: multimodal benchmarks}
We use the automatic evaluation from VLMEvalKit~\citep{duan2024vlmevalkit} including MMMU(Val)~\citep{mmmu}, MMStar~\citep{MMstar}, MME~\citep{mme}, OCRBench~\citep{ocrbench}, HallB(Avg)~\citep{guan2024hallusionbench}, MMB(Dev En V1.1)~\citep{liu2024mmbench}, TextVQA~\citep{singh2019towards}, DoCVQA~\citep{mathew2021docvqa}, InfoVQA~\citep{mathew2022infographicvqa}, AI2D~\citep{ai2d}, ChartQA~\citep{masry2022chartqa} and RWQA~\citep{rwqa}. These benchmarks collectively evaluate diverse multimodal capabilities, including general visual question answering, document, infographic and chart understanding, OCR reasoning, and hallucination robustness.

\subsection{SimplerEnv-Instruct}
As shown in~\Cref{tab:dis_task}, although SimplerEnv-Instruct is primarily designed for instruction generalization, we incorporate diverse out-of-distribution objects, environments, and distractors to prevent VLA models from exploiting the benchmark by disregarding the instructions.

\begin{table}[ht]
\caption{Task distribution}
 \label{tab:dis_task}
\centering
\begin{tabular}{lcccc}
\toprule
Attr.      & with OOD Obj. & with OOD Env. & with Distract Obj. & Only Language OOD \\
\midrule
Percentage(\%) & 50.0          & 62.5          & 35.0    & 5.0  \\
\bottomrule
\end{tabular}
\end{table}

\begin{figure}[h]
    \centering
    \includegraphics[width=1\linewidth]{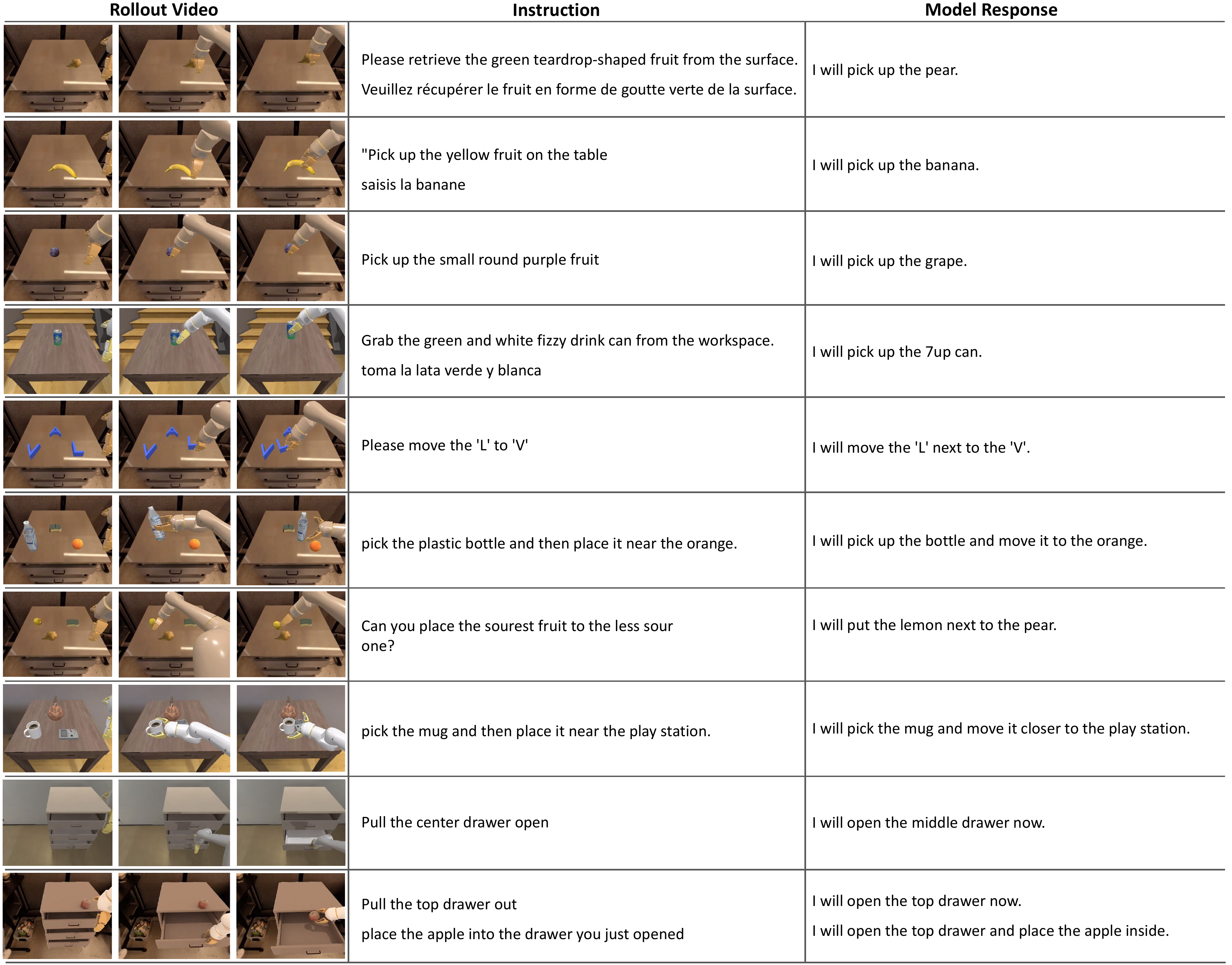}
    \caption{\textbf{Examples of task aggregation in SimplerEnv-Instruct.} We list ten examples with corresponding instructions and responses. Notably, InstructVLA shows the strong \textbf{zero-shot} ability to interpret multilingual instructions, recognize novel objects, and leverage OCR capabilities.}
    \label{fig:simpler_insturct1}
\end{figure}

\begin{figure}[h]
    \centering
    \includegraphics[width=1\linewidth]{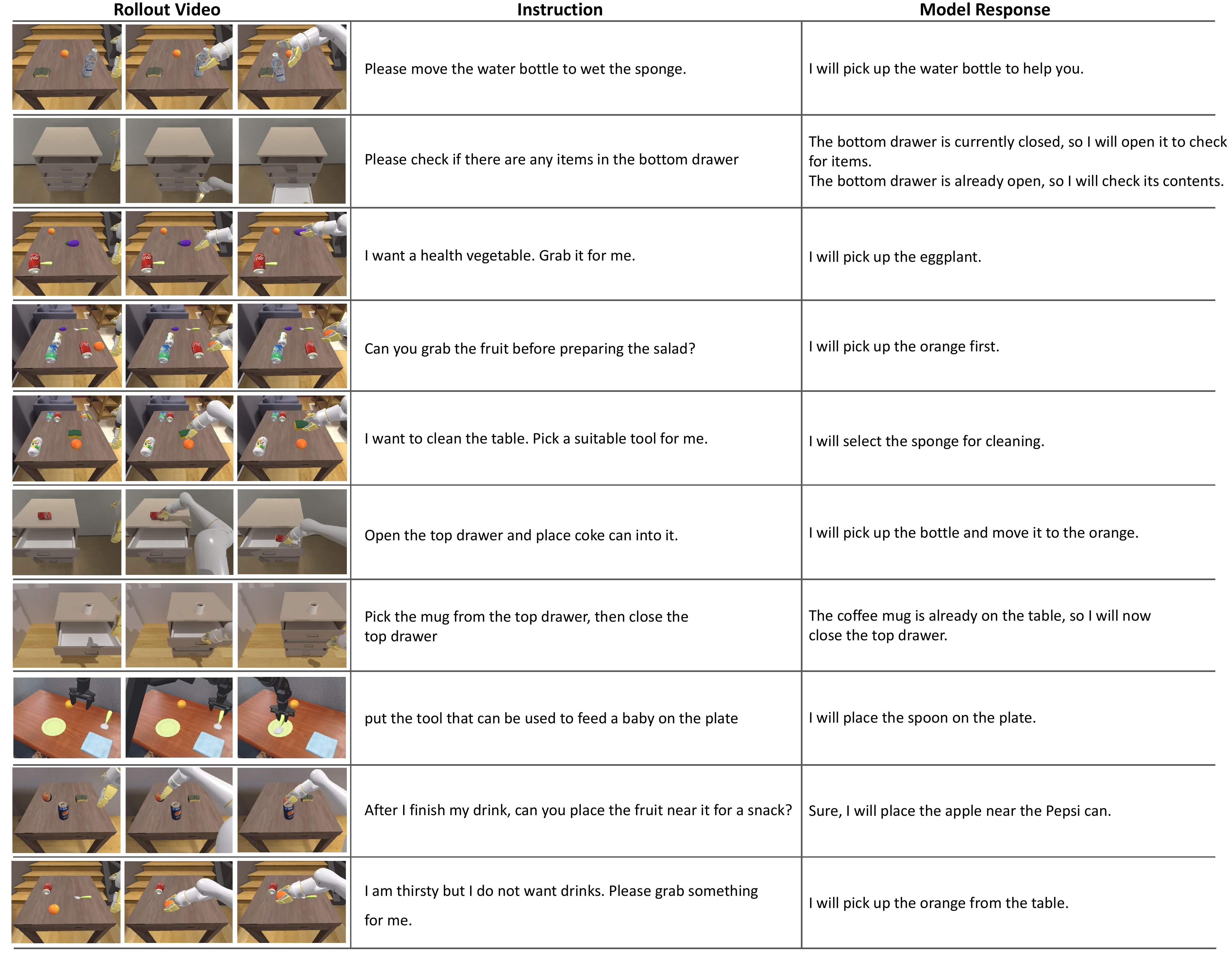}
    \caption{\textbf{Examples of Situated Reasoning in SimplerEnv-Instruct.} The second example's responses is recorded before and after the drawer is open.}
    \label{fig:simpler_insturct2}
\end{figure}

We select 10 task scenes with InstructVLA rollout actions and responses, as shown in~\Cref{fig:simpler_insturct1,fig:simpler_insturct2}, to illustrate its performance on both task aggregation and Situated Reasoning tasks.

\noindent\textbf{Acknowledgements of 3D assets.}
We gratefully acknowledge the creators of the following 3D assets used in \textit{SimplerEnv-Instruct}. All assets are licensed under the Creative Commons Attribution license:

\begin{itemize}[leftmargin=0.2in]
    \item \textit{Fruit Bowl Collection}\\
    \href{https://sketchfab.com/3d-models/fruit-bowl-collection-d89f6311cb0c4c37b0bf9cdd1e5abcb3}{https://sketchfab.com/3d-models/fruit-bowl-collection-d89f6311cb0c4c37b0bf9cdd1e5abcb3}
    \item \textit{Coffee Mug}\\
    \href{https://sketchfab.com/3d-models/coffee-mug-school-project-5f5ccee1514c440887c072fae8e0d699}{https://sketchfab.com/3d-models/coffee-mug-school-project-5f5ccee1514c440887c072fae8e0d699}
    \item \textit{Copper Tea Pot}\\
    \href{https://sketchfab.com/3d-models/copper-tea-pot-27f2ac58f7614f2796630bdc6f18ee2f}{https://sketchfab.com/3d-models/copper-tea-pot-27f2ac58f7614f2796630bdc6f18ee2f}
    \item \textit{Game Boy Classic}\\
    \href{https://sketchfab.com/3d-models/game-boy-classic-0ae80019e6f046168923286d7e628f6f}{https://sketchfab.com/3d-models/game-boy-classic-0ae80019e6f046168923286d7e628f6f}
\end{itemize}
All other assets are created using Blender or modified from SimplerEnv~\citep{simpleenv}.

\clearpage
\newpage

\section{Model Design and Training Details}

\label{sec: Model Design and Training Details}

\subsection{Instruction Format}

To train captioning, question answering, and instruction-following capabilities, we integrate all tasks into a unified dialogue format. For captioning and question answering, we adopt the template shown in Prompt~\hyperref[prompt:dialogue format]{3}, where the captioning instruction is sampled from Prompt~\hyperref[prompt: caption prompt]{4}. For free-form instructions, we append the postfix ``First answer my question.'' to elicit a direct response from the model, as illustrated in Prompt~\hyperref[prompt:instruction-format]{5}.

\phantomsection
\label{prompt:dialogue format}
{\scriptsize
\begin{tcolorbox}[colback=white, colframe=black!75!white, title=Dialogue Format]

\begin{lstlisting}[basicstyle=\ttfamily\scriptsize]
[
    {
        "role": "system", "content": DEFAULT_SYSTEM_MESSAGE
    },
    {
        "role": "user",
        "content": "[Question]",
        "image": image
    },
    {
        "role": "assistant", 
        "content": "[Answer]"
    },
    {
        "role": "user",
        "content": "What action should the robot take to [Instruction]?"
    },
    {
        "role": "assistant", 
        "content": "[Latent Action Queries]"
    }
]
\end{lstlisting}

\end{tcolorbox}
}

\phantomsection
\label{prompt: caption prompt}
{\scriptsize
\begin{tcolorbox}[colback=white, colframe=black!75!white, title=Caption Prompts]
\begin{itemize}[leftmargin=0.2in]
    \item Describe what’s on the table. Don’t mention the robot arm.
    
    \item What objects are in the scene? Ignore the robot arm.
    
    \item Tell me what you see on the table, not the robot.
    
    \item Describe the items and their positions, but skip the robot.
    
    \item Look at the table and describe it. Don’t include the arm.
    
    \item Only talk about the objects, not the machine.
    
    \item Give a short description of the scene, without the robot.
    
    \item Describe the setup on the table. Leave out the robotic arm.
    
    \item Focus on the objects and environment. Ignore the robot.
    
    \item Describe the environment and tabletop contents, excluding any robotic hardware.
\end{itemize}
\end{tcolorbox}
}

\phantomsection
\label{prompt:instruction-format}
{\scriptsize
\begin{tcolorbox}[colback=white, colframe=black!75!white, title=Instruction Format]

\begin{lstlisting}[basicstyle=\ttfamily\scriptsize]
[
    {
        "role": "system", "content": DEFAULT_SYSTEM_MESSAGE
    },
    {
        "role": "user",
        "content": "What action should the robot take to [Instruction]? First answer my question.",
        "image": image
    },
    {
        "role": "assistant", 
        "content": "[Response] [Latent Action Queries]"
    }
]
\end{lstlisting}

\end{tcolorbox}
}

{
\subsection{{MoE Adaptation}}
\label{sec: moe ada}
\begin{figure}
    \centering
    \includegraphics[width=1\linewidth]{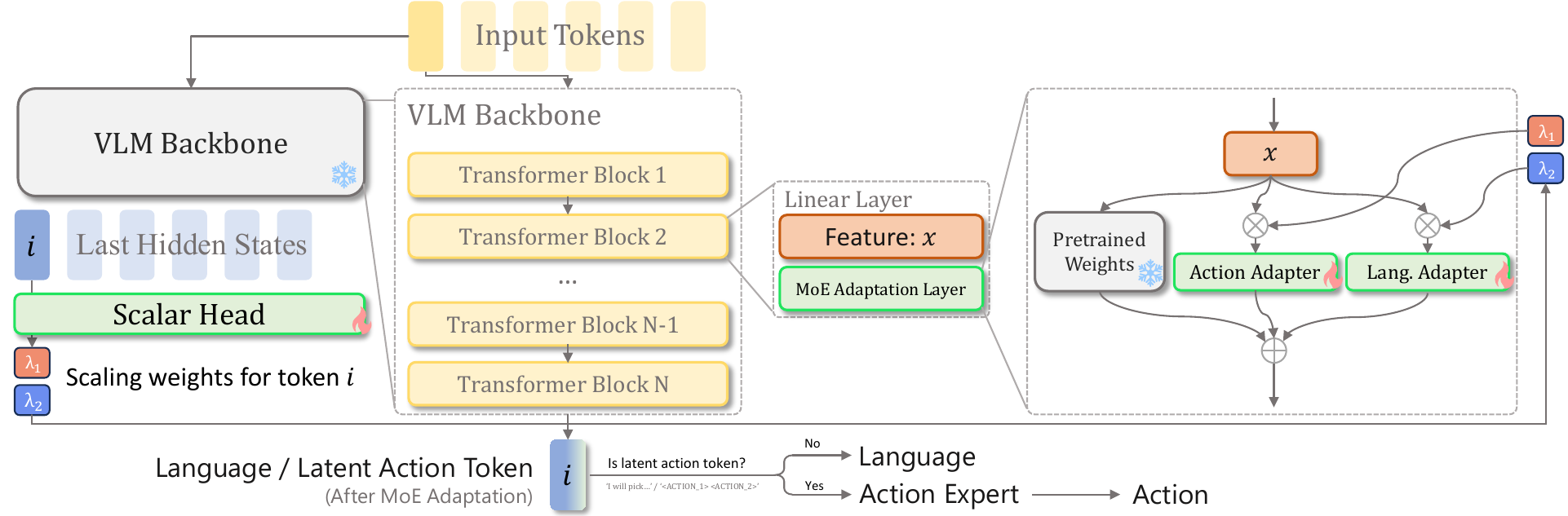}
    \caption{{\textbf{Detailed overview of the MoE adaptation architecture.} The frozen VLM backbone’s last hidden states are classified by a scalar head to produce gating weights $\lambda_1$ and $\lambda_2$, which control the weighted MoE adaptation. Similar to finetuning VLMs with multiple LoRA adapters, the MoE adaptation computes a weighted sum over the LoRA experts. The predicted tokens are then used differently based on their token type: language tokens are directly decoded as the model’s response, while features corresponding to action tokens are decoded by the action expert (see~\Cref{fig:model} (right)) to produce continuous actions.}}
    \label{fig:moe_ada}
\end{figure}
We briefly outline the LoRA~\citep{lora} mechanism, which serves as the basis for MoE adaptation design~\cite{xlora}. LoRA models weight updates as lying in a low-dimensional subspace by freezing the original weights $W_0\in\mathbb{R}^{d\times k}$ and parameterizing updates through a low-rank decomposition:
\begin{equation}
W_0 + \Delta W_0 = W_0 + BA,
\end{equation}
where $B\in\mathbb{R}^{d\times r}$ and $A\in\mathbb{R}^{r\times k}$ with $r\ll \min(d,k)$. The forward pass becomes
\begin{equation}
h = W_0 x + \Delta W_0 x = W_0 x + BAx.
\end{equation}

In practice, LoRA introduces a fixed scaling factor $\alpha$, yielding
\begin{equation}
h = W_0 x + \alpha BAx.
\end{equation}

As shown in~\Cref{fig:moe_ada}, the scalar head in MoE adaptation predicts a gating coefficient $\lambda$ to reweight two LoRA adapters, inspired by mixture-of-experts architectures~\citep{moe}. This is achieved by rescaling each adapter’s scaling factor :
\begin{equation}
\alpha^{*}_{i} = \alpha_{i}\cdot\lambda_{i},
\end{equation}
where $i$ indexes the LoRA adapters. The scalar head together with the LoRA adapters constitutes the MoE adaptation module.

Unlike X-LoRA~\citep{xlora}, which trains the scalar head and LoRA adapters separately. We first pretrain the action adapter in Stage-1 following the standard LoRA pipeline. In Stage-2, we introduce the language adapter for embodied reasoning and the scalar head, and train the complete MoE adaptation module jointly. We instantiate the scalar head as a 4-layer MLP with dimensions shown in~\Cref{tab:model} for simplicity, without any other auxiliary loss design.
}

\subsection{{Learning Objective and Inference Procedure}}
\label{sec: loss}
We adopt flow matching~\citep{pi_0,flowmatching} to learn the action chunk $\mathbf{A} \in \mathbb{R}^{H \times 7}$~\citep{act} over a horizon $H$. The training objective is defined as the flow matching loss:

\begin{equation}
\mathcal{L}_{FM} = \mathbb{E}\left[\left\| V{\theta}(\mathbf{A}^{\tau}, q_t) - (\epsilon - \mathbf{A}) \right\|^2\right],
\end{equation}

where $\tau \in [0,1)$ denotes the flow step, and $V_{\theta}(\mathbf{A}^{\tau}, q_t)$ is the network output conditioned on $q_t$, which encodes information from DINOv2~\citep{oquab2023dinov2} and a latent action $C$. The interpolated noisy action is given by $\mathbf{A}^{\tau} = \tau \mathbf{A} + (1 - \tau) \epsilon$, with $\epsilon \sim \mathcal{N}(\mathbf{0}, \mathbf{I})$.

During inference, we generate the action chunk using forward Euler integration:

\begin{equation}
\mathbf{A}^{\tau + 1/N} = \mathbf{A}^{\tau} + \frac{1}{N} V_{\theta}(\mathbf{A}^{\tau}, q_t),
\end{equation}

starting from $\mathbf{A}^{0} \sim \mathcal{N}(\mathbf{0}, \mathbf{I})$, with $N = 10$ denoising steps.
{
For language prediction, we use the standard cross-entropy loss. We simply sum the two losses with a 1:1 weighting. The data we used are detailed in~\Cref{tab: data}.

\begin{table}[]
\centering
\caption{{Overview of data used in Stage-1 action pretraining and Stage-2 VLA instruction tuning.}}
% \resizebox{0.8\textwidth}{!}{%
\begin{tabular}{lccc}
\toprule
\multicolumn{1}{c}{Supervision Type} & Self-Annotated & Stage-1 & Stage-2 \\
\midrule
Action    & \ding{55} & \ding{51}& \ding{51}\\
Language Motion  & \ding{51}& \ding{51}& \ding{51}\\
General Multimodal Datasets          & \ding{55} & \ding{55}& \ding{51}\\
Embodied Reasoning(VLA-IT dataset)   & \ding{51} & \ding{55}& \ding{51}\\ \bottomrule      
\end{tabular}
% }
\label{tab: data}
\end{table}
}
\subsection{Model Parameters}
\begin{table}
\centering
\caption{\textbf{Model parameters.} ``Adapter'' and ``Scalar Head'' are used for MoE adaptation. Specifically, two LoRA adapters are used to learn latent action generation and assistant response during VLA-IT.}
\begin{tabular}{lll}
\toprule
Component       & Parameter   & Value                          \\
\midrule
Adapter         & Rank        & 128                            \\
                & Alpha       & 256                            \\
                & Dropout     & 0.05                           \\
                & Target      & Attn. Q/K/V/O                  \\
                &             & MLP Up/Down                    \\
                \midrule
Scalar Head      & {Hidden Size} &  {$2048\to128\to128\to128\to2$}   \\
                & {Activation}  &  {ReLU} \\
                \midrule
Action Backbone & Depth       & 12                             \\
                & Head        & 12                             \\
                & Hidden Size & 768                            \\
                & RoPE Theta  & 1000                           \\
                \midrule
Proprioception Encoder(Optional) & Hidden Size & 8 $\to$ 768 $\to$ 768 \\
                & Activation & SiLU \\
                \midrule
Action Encoder with Time Embedding  & Hidden Size & 7+768 $\to$ 1536 $\to$ 768 \\
                & Activation & SiLU \\
\bottomrule
\end{tabular}
\label{tab:model}
\end{table}

\begin{table}
\centering
\caption{\textbf{Flow matching parameters.} The time steps is sampled from $p(\tau) = \beta(\frac{s-\tau}{s};1.5,1)$~\citep{pi_0}}
\begin{tabular}{lll}
\toprule
Component          & Parameter       & Value \\
\midrule
Flow Sampling      & s               & 0.999 \\
                   & Inference Steps & 10    \\
Sinusoidal Time Embed & Max Period      & 100   \\
\bottomrule
\end{tabular}
\label{tab:flowmatching}
\end{table}

Additional model parameters are provided in~\Cref{tab:model}, with flow-matching sampling settings detailed in~\Cref{tab:flowmatching}. All projectors—including those aligning latent actions and DINO-ViT visual features to the action expert’s dimension—use a simple two-layer MLP with SiLU activation. The action head, also a shallow MLP with SiLU, maps the action expert’s hidden states to $\mathbb{R}^{N \times 7}$, where $N=16$ is the prediction horizon and 7 denotes the action dimension, including the gripper.

\subsection{Inference Speed}

We evaluate the inference speed of InstructVLA on a single A100 GPU with BF16 precision, as shown in~\Cref{tab: speed}. To support language feedback during evaluation (i.e., CoT inference), in the ``Thinking'' setting, we enable VLM auto-regressive generation every 20 action expert steps. The ``Action Only'' setting bypasses language generation and directly decodes latent actions via a single VLM forward pass. In the ``Latent Action Caching'', latent actions are generated every two expert steps; this introduces minimal performance impact. All settings are tested without action chunking. Note that although the model predicts 16-step action sequences, only one step is executed.

\begin{table}[h]
\centering
\caption{\textbf{Inference speed.} Inference speed is evaluated under three settings \textbf{without using action chunking}. Each evaluation includes a 50-step warm-up followed by 200 steps for stable measurement.}

\begin{tabular}{cccc}
\toprule
                        & With Language & Action Only & Latent Action Caching \\
\midrule
Inference Frequency(Hz) & 2.51          & 3.50        & 4.96                    \\
\bottomrule
\end{tabular}
\label{tab: speed}
\end{table}

\subsection{Experiments Compute Resources}

The action pretraining phase requires approximately 27 hours on 64 A100 GPUs, with each node equipped with 1 TB of CPU memory. The VLA-IT phase takes about 12 hours under the same GPU configuration. Simulator-based evaluations are conducted with 8 A100 GPUs, while real-world experiments involve 4 hours of training on 32 A100 GPUs and deployment on a single A100 GPU. 

To assess minimal training resources, we further reproduce pretraining results using 8 A800 GPUs in 2.5 days as shown in~\Cref{tab:eval_results}.

\begin{table*}[ht]
\centering
\caption{\textbf{Evaluation results under different training settings.} 
We report mean success rates ($\% \pm$ standard error) across tasks, with Overall denoting the average over all tasks. ``Main'' corresponds to the results reported in the main table.}
\label{tab:eval_results}
\resizebox{\textwidth}{!}{
\begin{tabular}{lccccccccccccc}
\toprule
 & \multicolumn{4}{c}{Google Robot (VA)} & \multicolumn{4}{c}{Google Robot (VM)} & \multicolumn{4}{c}{WidowX Robot} & Overall \\
\cmidrule(lr){2-5} \cmidrule(lr){6-9} \cmidrule(lr){10-13}
Setting & Pick Coke & Move Near & Drawer & Apple In & Pick Coke & Move Near & Drawer & Apple In & Put Spoon & Put Carrot & Stack Cube & Put Eggplant & \\
\midrule
Main (8 epochs) & 92.3$\pm$0.7 & 71.9$\pm$1.3 & 61.7$\pm$0.8 & 33.1$\pm$2.5 & 79.6$\pm$1.9 & 68.3$\pm$3.1 & 52.3$\pm$3.8 & 50.3$\pm$3.8 & 43.1$\pm$6.4 & 40.3$\pm$14.6 & 9.7$\pm$9.6 & 94.4$\pm$2.4 & 56.2$\pm$2.9 \\
8 GPUs (4 epochs)       & 94.0$\pm$0.2 & 76.9$\pm$0.5 & 62.8$\pm$1.6 & 39.3$\pm$4.3 & 88.7$\pm$1.7 & 67.4$\pm$2.1 & 61.8$\pm$2.5 & 31.7$\pm$1.9 & 62.5$\pm$11.0 & 48.6$\pm$2.4 & 8.3$\pm$4.2 & 95.8$\pm$4.1 & 61.5$\pm$1.3 \\
\bottomrule
\end{tabular}}
\end{table*}

\clearpage
\newpage

\section{Multimodal Examples}
\label{sec: Multimodal Examples}
\begin{figure}[h]
    \centering
    \includegraphics[width=1\linewidth]{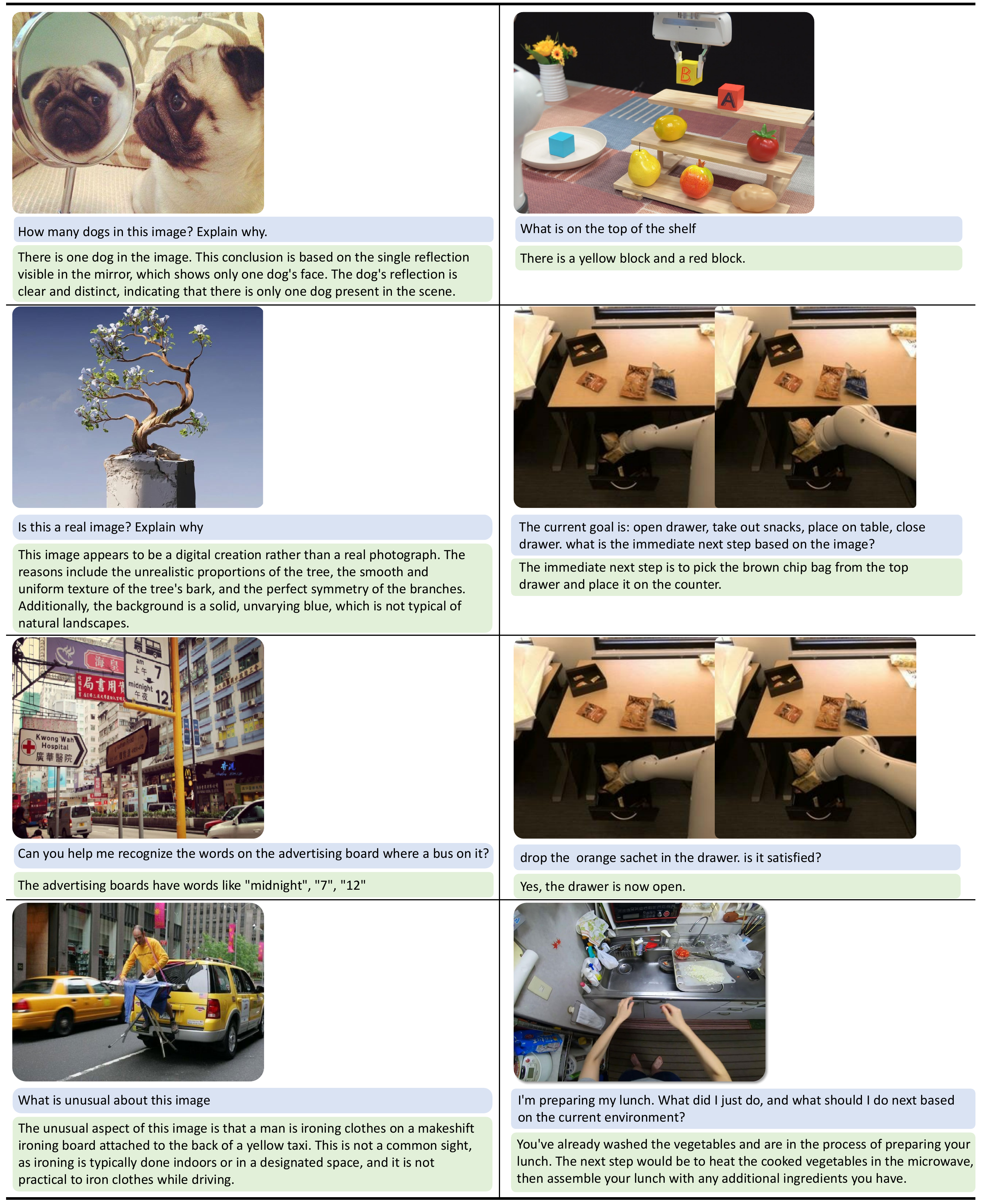}
    \caption{\textbf{Zero-shot multimodal question answering.} Four commonsense and four embodied examples are selected.}
    \label{fig:multi-modal}
\end{figure}

\Cref{fig:multi-modal} illustrates InstructVLA’s multimodal and embodied commonsense reasoning across diverse scenarios. The model demonstrates accurate visual inference (e.g., recognizing a dog via reflection, identifying synthetic images), basic scene text recognition, and reliable grounding of objects and colors. In manipulation tasks, it interprets high-level goals, predicts appropriate next actions, and verifies task completion. These capabilities showcase its integration of perception, language, and manipulation, enabling effective performance in complex daily-life scenarios.

\clearpage
\newpage

\section{Real-world Experiments Setup and Analysis}
\label{sec: Real-world Experiments Setup}

\begin{figure}[h]
    \centering
    \includegraphics[width=1\linewidth]{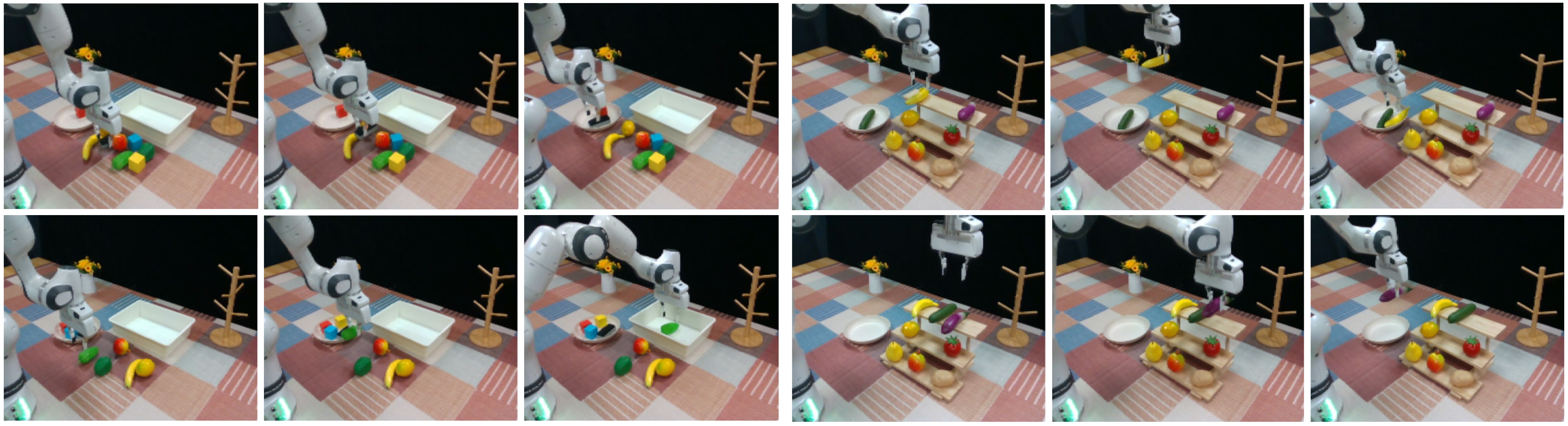}
    \caption{\textbf{Real-world dataset examples.} Four examples from the few-shot training set, illustrating cluster classification tasks (left) and rack pick-and-place tasks (right).}
    \label{fig:realworld_demo}
\end{figure}

\begin{figure}[h]
    \centering
    \includegraphics[width=1\linewidth]{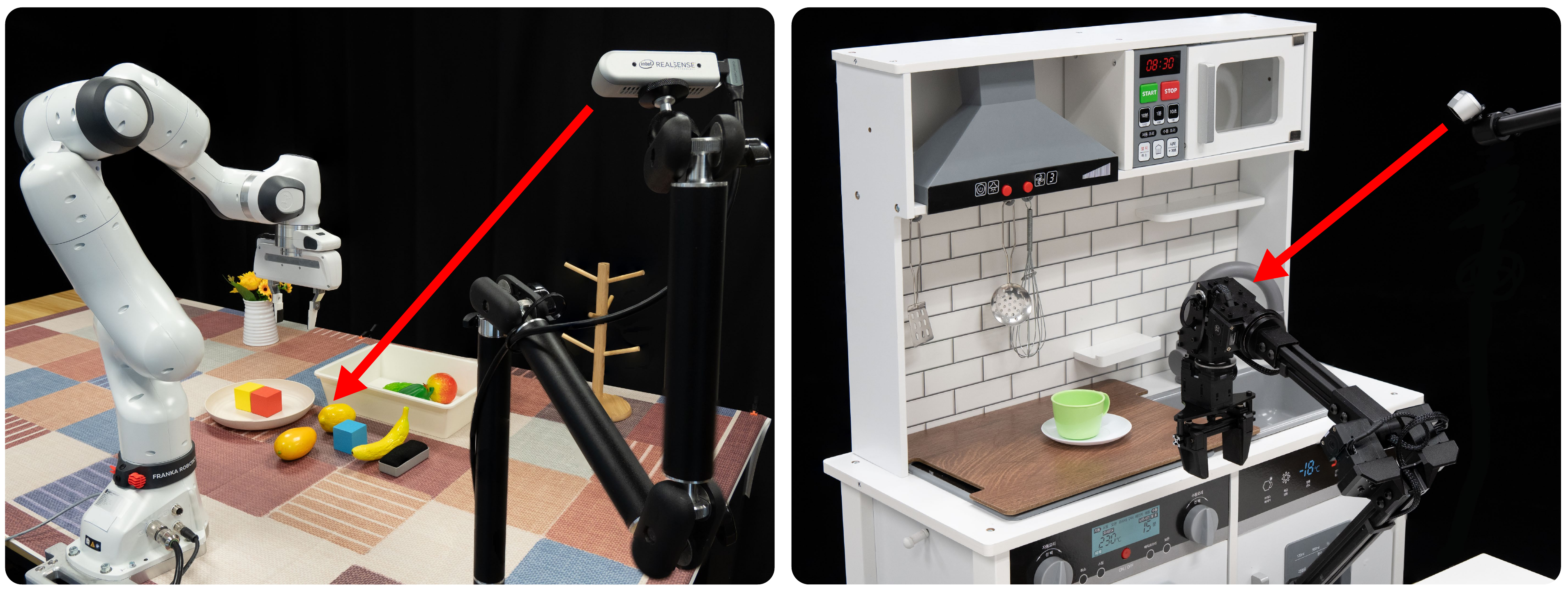}
    \caption{\textbf{Real-world settings.} A third-person view is captured using an Intel D435i camera for the Franka (few-shot) and WidowX (zero-shot) settings.}
    \label{fig:setting}
\end{figure}

We collect data exclusively for few-shot settings as shown in~\Cref{fig:realworld_demo}. In the first setting, which focuses on grasping objects in a clustered arrangement, the robot is instructed to classify objects within a $20 \times 40$ cm region on the table—placing all cubic objects into a plate and all others into a box. This setting includes 70 complete episodes, totaling 677 pick-and-place actions. In the second setting, which emphasizes spatial actions, the robot is instructed to randomly grasp three objects from the top of a rack and place them into a plate. We collect 60 complete episodes for this setting, comprising 180 pick-and-place actions. The experimental setups are depicted in~\Cref{fig:setting}.

\begin{figure}[h]
    \centering
    \includegraphics[width=1\linewidth]{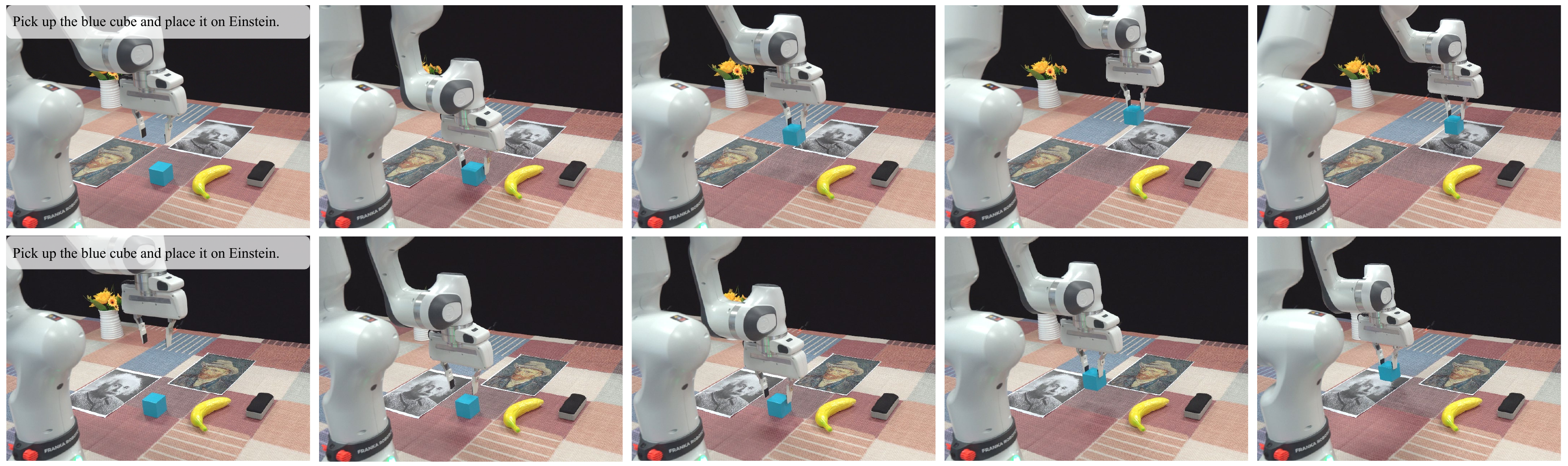}
    \caption{\textbf{Zero-shot grounding.} In a clustered pick-and-place setting, InstructVLA accurately places the blue cube by semantically grounding the reference to the celebrity.}
    \label{fig:real example 1}
\end{figure}

\begin{figure}[h]
    \centering
    \includegraphics[width=1\linewidth]{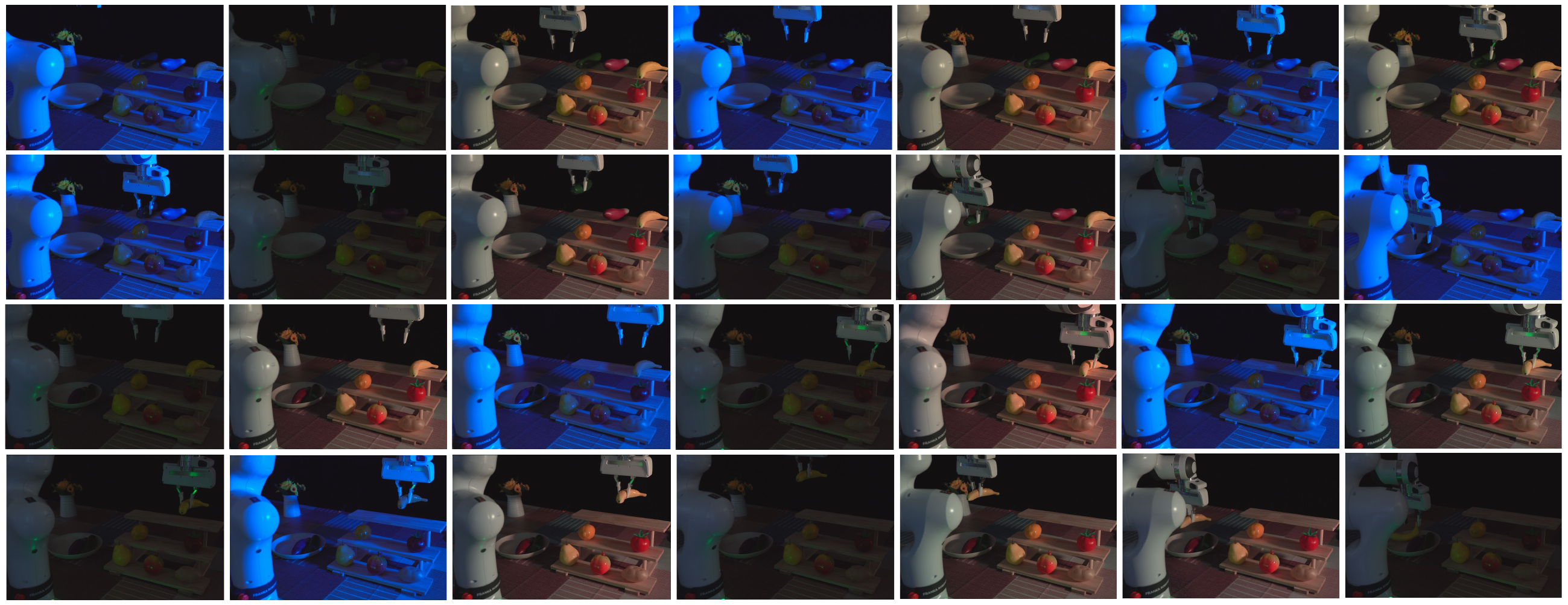}
    \caption{\textbf{Light distraction.} Stable visual features from DINO and SigLIP enable the model to operate robustly under extreme out-of-distribution lighting conditions.}
    \label{fig:light-distraction}
\end{figure}

\begin{figure}
    \centering
    \includegraphics[width=1\linewidth]{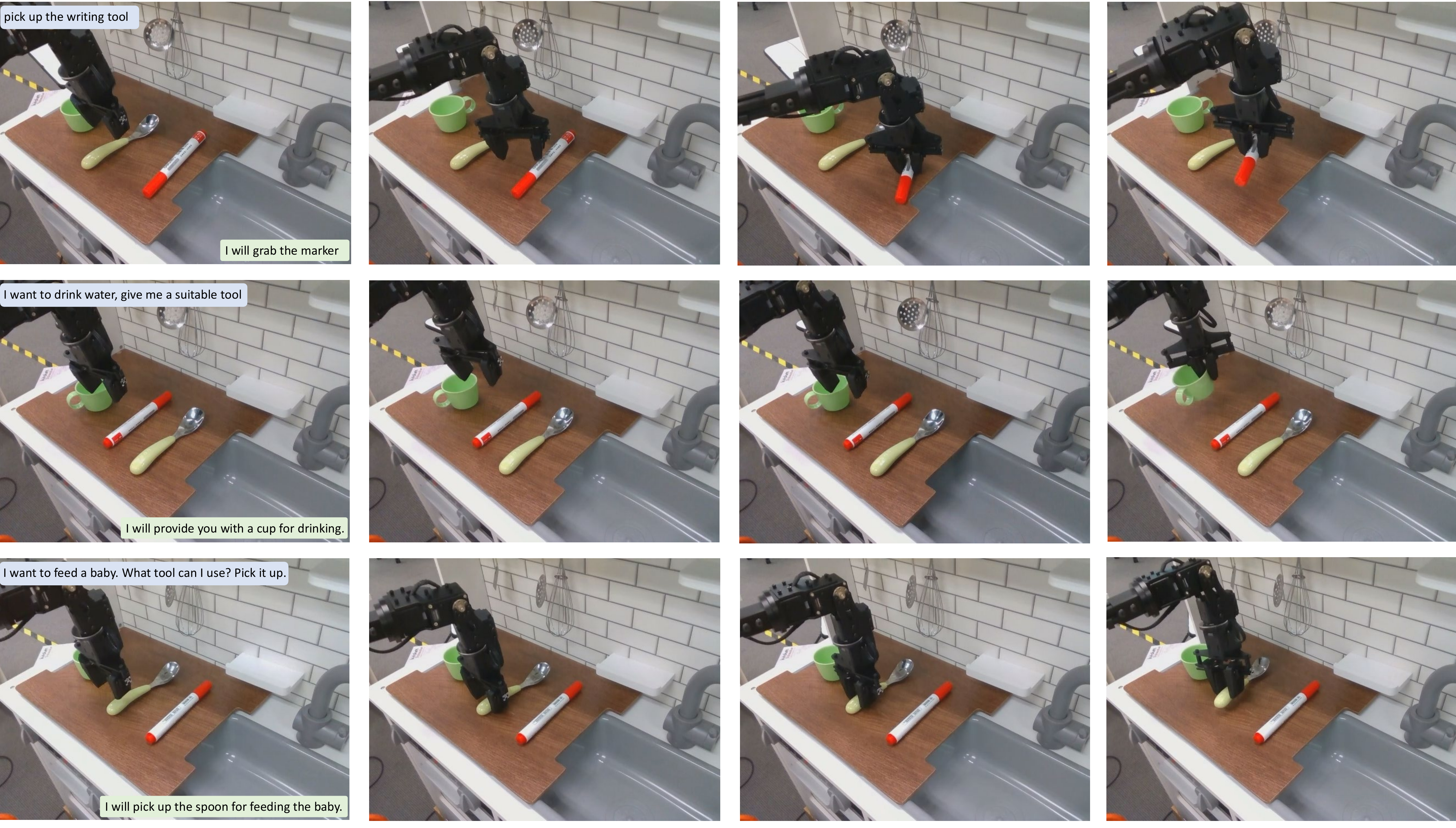}
    \caption{\textbf{Zero-shot evaluation.} We perform zero-shot evaluation in the Bridge kitchen environment with augmented background and novel objects. The instruction and model response are visualized in the first image.}
    \label{fig:zero-shot}
\end{figure}

To assess semantic grounding in novel contexts, we replace the plate and box in the cluster classification setting with images of celebrities. As illustrated in~\Cref{fig:real example 1}, the model accurately interprets instructions and places the blue cube correctly by leveraging object and celebrity recognition. 

\Cref{fig:light-distraction} shows that InstructVLA remains robust under extreme lighting conditions, supported by stable visual features from DINO and SigLIP. Finally, we evaluate zero-shot generalization in the Bridge kitchen environment with augmented backgrounds and unfamiliar objects. As shown in~\Cref{fig:zero-shot}, the model successfully follows novel instructions and completes the tasks.

\clearpage
\newpage

\section{Broader Impacts and Future Work}
\label{sec: Broader Impacts}

\subsection{Limitation}

InstructVLA integrates world knowledge into manipulation tasks by performing multimodal reasoning prior to action generation. Recent VLMs also excel at long-context processing and multi-turn dialogue. This motivates curating interleaved manipulation and reasoning with multi-turn interaction to support long-horizon tasks involving user intervention or reasoning-action alternation~\citep{yao2023react}. {Furthermore, the existing tasks are limited to basic primitives such as open/close and pick/place due to the constraints of the datasets we use~\citep{RT-1,Bridge_data} and the capabilities of the simulator. In contrast, standard VLM benchmarks typically contain thousands of tasks.} Extending InstructVLA and SimplerEnv-Instruct bench to more dexterous skills is essential for real-world deployment.

\subsection{LLM Usage Statement}

We employed large language models (LLMs) solely for grammar refinement and minor linguistic polishing. All LLM-assisted edits were carefully reviewed and verified by the authors to ensure that no fabricated content or unintended alterations to the original meaning were introduced. The research ideas, experimental design, data analysis, and conclusions presented in this work were entirely conceived and executed by the authors without LLM assistance.

\subsection{Broader Impacts}

InstructVLA contributes to the advancement of general-purpose embodied agents by integrating vision-language understanding with action generation. Its ability to follow free-form instructions and generalize to novel tasks supports applications in assistive robotics and human-robot collaboration. Nonetheless, as with other large pretrained models, careful attention must be given to potential limitations such as dataset bias and safety in real-world deployment. Ensuring responsible use and reliable performance across diverse environments is essential.

\subsection{Future Work}

We plan to incorporate additional sensory modalities, such as depth and tactile feedback, to enhance safety and reliability in physical interactions. Leveraging recent advances in digital twins and simulation technologies, we aim to reduce reliance on real-world data by utilizing large-scale synthetic datasets. Finally, we will extend the evaluation and deployment of InstructVLA to a broader range of environments to further assess its generalization capabilities.

\clearpage
\newpage

\end{document}